\documentclass{article} 
\usepackage{iclr2026_conference,times}

\usepackage{amsmath,amsfonts,bm}

\def\eqref#1{equation~\ref{#1}}

\def\1{\bm{1}}

\DeclareMathAlphabet{\mathsfit}{\encodingdefault}{\sfdefault}{m}{sl}
\SetMathAlphabet{\mathsfit}{bold}{\encodingdefault}{\sfdefault}{bx}{n}

\usepackage{url}
\usepackage{enumitem}
\usepackage{booktabs}       
\usepackage{amsfonts}       
\usepackage{nicefrac}       
\usepackage{microtype}      
\usepackage[dvipsnames,svgnames,x11names]{xcolor}
\usepackage{multirow}
\usepackage[hidelinks]{hyperref}
\usepackage{graphicx}
\usepackage{amsmath}
\usepackage{tabularx}
\usepackage{pifont}
\usepackage{subcaption}
\usepackage{pgfplots}
\usepackage{pgfplotstable}
\usepackage{enumitem}
\usepackage{pgfkeys}
\usepackage[ruled,vlined]{algorithm2e}
\usepackage[most]{tcolorbox}
\usepackage[table]{xcolor}
\usepackage{adjustbox}
\usepackage{caption}
\usepackage{amssymb}
\usepackage{wrapfig}

\definecolor{deepgreen}{rgb}{0.0, 0.5, 0.0}
\newcommand{\xmark}{\textcolor{red}{\textbf{\ding{55}}}}
\newcommand{\cmark}{\textcolor{deepgreen}{\textbf{\ding{51}}}}

\pgfplotsset{compat=1.18} 

\newtcolorbox{PlainBox}[1][]{
  enhanced,
  width=\textwidth,
  colback=yellow!20,
  colframe=black,
  boxrule=1pt,
  arc=4pt,
  left=6pt, right=6pt, top=6pt, bottom=6pt,
  #1
}

\newtcolorbox{SideBySideBox}[1][]{
  enhanced,
  width=\textwidth,
  colback=yellow!20,
  colframe=black,
  boxrule=1pt,
  arc=4pt,
  left=6pt, right=6pt, top=6pt, bottom=6pt,
  sidebyside,
  sidebyside gap=5mm,
  sidebyside align=top seam,
  lefthand width=0.2\textwidth,
  #1
}
\newtcolorbox{SideBySideBox2}[1][]{
  enhanced,
  width=\textwidth,
  colback=yellow!20,
  colframe=black,
  boxrule=1pt,
  arc=4pt,
  left=6pt, right=6pt, top=6pt, bottom=6pt,
  sidebyside,
  sidebyside gap=5mm,
  sidebyside align=top seam,
  lefthand width=0.5\textwidth,
  #1
}

\title{Meta-Adaptive Prompt Distillation for Few-Shot Visual Question Answering}

\author{Akash Gupta, Amos Storkey, Mirella Lapata \\
School of Informatics, University of Edinburgh\\
\texttt{\{akash.gupta,a.storkey\}@ed.ac.uk, mlap@inf.ed.ac.uk} \\
}

\iclrfinalcopy 
\begin{document}

\maketitle

\begin{abstract}
\label{sec:intro}
Large Multimodal Models (LMMs) often rely on in-context learning (ICL) to perform new visual question answering (VQA) tasks with minimal supervision. However, ICL performance, especially in smaller LMMs, does not always improve monotonically when increasing the number of examples. We hypothesize that this happens because the LMM is overwhelmed by extraneous information in the image embeddings that is irrelevant to the downstream task. To address this, we propose a meta-learning approach that induces few-shot capabilities in LMMs through a fixed set of soft prompts distilled from task-relevant visual features, which are adapted at test time using a small number of examples. We facilitate this distillation through an attention-mapper module that can be easily integrated with any LMM architecture and is jointly learned with soft prompts. Evaluation on the VL-ICL Bench shows that our method successfully achieves task adaptation in low-data regimes with just a few gradient steps, outperforming  ICL by 21.2\%.  Comparisons with parameter-efficient finetuning methods demonstrate that meta-learning further enhances this adaptation by 7.7\% for various VQA tasks.\footnote{We release our code here - \url{https://github.com/akashgupta97/MAPD}}
\end{abstract}

\section{Introduction}

Humans have the remarkable ability to quickly learn new tasks in
multimodal environments with just a few trial-and-error
attempts. Extensive research in cognitive science suggests that this
ability arises from learning hierarchical abstractions and maintaining
shared structural priors across related tasks based on past
experiences \citep{GRIFFITHS201924, finn2018learning,
  kirsch2022selfreferential}.  Drawing on this prior knowledge enables
rapid learning in new situations and reduces the need for large
amounts of task-specific demonstrations \citep{finn2017model}.

Large Multimodal Models (LMMs) are able to perform a multitude of tasks
ranging from reasoning to fine-grained image understanding and visual
question answering \citep{Liu_2024_CVPR,li2023ottermultimodalmodelincontext,laurencon2024mattersbuildingvisionlanguagemodels}. They
are typically built on top of a base Large Language Model (LLM) by
supplementing it with a vision encoder and a connecting module that
acts as a bridge for different modalities to interact.  When
(pre)trained at sufficient scale and finetuned on a wide range of
multimodal tasks (with natural language instructions), LMMs can learn
\emph{new} tasks by virtue of in-context learning (ICL), i.e., by
being prompted with a few input-output examples, without requiring any
updates to  model parameters \citep{zhao2024mmicl,
  zong2025vlicl,coda-forno2023metaincontext}. Although the
training-free nature of ICL has led to its rapid adoption across tasks
and domains, its underlying mechanism remains ill-understood
\citep{hendel-etal-2023-context, huang2024multimodal} and its
empirical behaviour can be inconsistent.   

\begin{figure}[t]
\includegraphics[width=\textwidth,keepaspectratio]{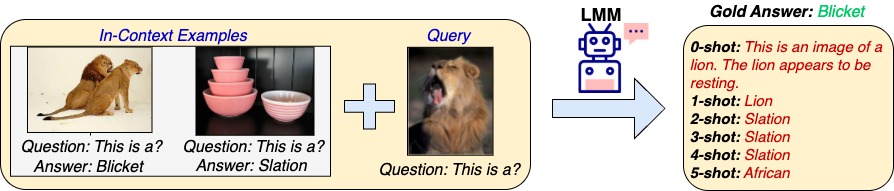}
\caption{Failure case of LLaVA-OneVision-7B  \citep{li2025llavaonevision} on an example from the Fast Open-Ended MiniImageNet classification task \citep{NEURIPS2021_01b7575c}. When no in-context examples are provided (0-shot), the model generates a generic description of the image. As more examples (shots) are added, it begins to learn the answer format (single word), but still fails to grasp the task, producing incorrect or irrelevant predictions. We only show the in-context examples (left) for 2-way 1-shot setting for the sake of brevity but provide model predictions (in red) for up to 5~shots.} 
\label{fig:motive}
\label{img:model}
\end{figure}

\citet{zong2025vlicl} demonstrate that ICL is most effective for large-scale LMMs  ({\raise.17ex\hbox{$\scriptstyle\mathtt{\sim}$}}72B parameters), while smaller models ($\leq$7B parameters)  struggle with increasing  in-context examples and their performance either plateaus or deteriorates, even when extending the context length or giving detailed instructions. They attribute this limitation to the fact that smaller models struggle with the large number of image tokens in long sequences. They become confused and perform the task haphazardly or default to their parametric knowledge, effectively ignoring the in-context examples.  Figure~\ref{fig:motive} shows a failure case from the Fast Open-Ended MiniImageNet dataset \citep{NEURIPS2021_01b7575c}, using LLaVA-OneVision-7B \citep{li2025llavaonevision}.  The task is framed in a 2-way N-shot format where a \emph{support} set with N labeled examples of two classes is provided.
The model uses ICL with the support set to classify new \emph{query} examples from the two classes.  Without any support set or in-context examples (0-shot), the model outputs a generic description about the image based on  parametric knowledge and ultimately fails to answer correctly, despite being prompted with a few examples. 

\begin{wrapfigure}[18]{R}{0.45\textwidth}
\hspace{-.2cm}
\vspace{-2.8ex}
\begin{tikzpicture}[scale=.6]
\begin{axis}[
    width=10cm,
    height=8cm,
    xlabel={Number of Shots},
    ylabel={Mean Accuracy (\%)},
    xmin=-0.5, xmax=8.5,
    ymin=10, ymax=75,
    xtick={0,1,2,4,8},
    ytick={15,25,35,45,55,65,75},
    grid=major,
    grid style={dashed, gray!30},
    legend style={at={(0.1,1.14)}, anchor=north west, font=\small},
    legend cell align={left},
    xlabel style={font=\large},
    ylabel style={font=\large},
    tick label style={font=\normalsize},
]

\addplot[
    color=red,
    mark=*,
    mark size=3pt,
    line width=1.5pt,
] coordinates {
    (0, 18.5)
    (1, 42.8)
    (2, 42.2)
    (4, 37.5)
    (8, 34.1)
};
\addlegendentry{Image-to-Text (I2T)}

\addplot[
    color=ForestGreen,
    mark=*,
    mark size=3pt,
    line width=1.5pt,
] coordinates {
    (0, 19.3)
    (1, 49.3)
    (2, 47.4)
    (4, 41.2)
    (8, 38.6)
};
\addlegendentry{Image-to-Text (I2T) + Detailed}

\addplot[
    color=blue,
    mark=square*,
    mark size=3pt,
    line width=1.5pt,
] coordinates {
    (0, 25.5)
    (1, 50.9)
    (2, 59.9)
    (4, 66.4)
    (8, 68.6)
};
\addlegendentry{Text-to-Text (T2T)}

\end{axis}
\end{tikzpicture}
\vspace{2ex}
\caption{I2T and T2T performance with LLaVA-OneVision-7B on Operator Induction and CLEVR Count Induction tasks.} \label{fig:prem_analysis} 
\end{wrapfigure}
Building on this observation,  we hypothesize that effective few-shot adaptation at  test time may be compromised by the  information added by the image embeddings. Figure \ref{fig:prem_analysis}  compares  Image-to-Text (I2T, red) and Text-to-Text (T2T, blue) performance of LLaVA-OneVision-7B  on the Operator Induction and CLEVR Count Induction tasks (see Appendix \ref{sec_app:eval_data}). The results reveal a significant performance gap: T2T ICL consistently outperforms I2T and improves monotonically with additional shots. Moreover, adding detailed task instructions to I2T (green; see Appendix \ref{sec_app:prem_analysis_ins}) actually degrades performance, suggesting that naively increasing the number of image embeddings in context impairs the model's inherent ICL ability. While a set of more precise image embeddings would be preferable, their continuous nature makes it challenging to distill task-specific information from them. As an alternative, we propose to \emph{learn} a fixed set of \emph{new} embeddings that can be easily finetuned at test time.

 This idea of task adaptation has gained significant traction in the literature through \emph{prompt tuning} \citep{lester-etal-2021-power} which finetunes a set of continuous \emph{soft} prompts  while keeping the underlying language model frozen; the prompts are prepended in the context at test time,  effectively steering the  model toward the desired task. 
Our approach learns new tasks using soft prompts that receive task information from the LLM in the form of loss gradients during finetuning. These gradients update the soft prompts which when fused with the image embeddings are able to  distill  relevant features from them. To facilitate this fusion, we propose an attention-mapper that uses a multi-head attention \citep{vaswani2017attention} architecture  for extracting relevant task-specific image information and can be substituted in the projection layer of any LMM architecture. 

Our approach relies on rapidly adapting to new tasks at test time using only a few examples, which is not addressed by traditional finetuning methods. 
Prior work \citep{finn2017model, ravi2017optimization}  addressed this challenge by training  a meta-learner that can infer an optimal learning strategy for a new task after being exposed to a distribution of tasks. 
We apply this procedure to  our multimodal prompt distillation setting by employing the widely known MAML algorithm \citep{finn2017model} and use its lightweight first-order approximation  to train the  attention-mapper and soft prompts.  We focus on visual
question answering (VQA; \citealt{antol2015vqa}; see example in Figure~\ref{fig:motive}), a general-purpose task often used to
evaluate the image understanding capabilities of LMMs, and
demonstrate the benefits of MAML training applied to LMM architectures. 
Our contributions are 
as follows:
\begin{itemize}[noitemsep]
\item We introduce MAPD (\textbf{M}eta-\textbf{A}daptive \textbf{P}rompt \textbf{D}istillation), an alternative to in-context learning that meta-learns a fixed set of soft prompts within large multimodal models (LMMs) via distillation. MAPD enables adaptation to new tasks with a few examples using a few gradient updates at test time, and consistently improves performance as the number of shots increases. To our knowledge, this is the first exploration of meta-learned prompt distillation for cross-task generalization in LMMs under  low-data settings.
    \item We propose a flexible attention-mapper module, inspired by \citet{najdenkoska2023meta}, that exploits all  patch features from the vision encoder and can be readily incorporated into the projection layer of any LMM architecture. It is trained jointly with \emph{soft} prompts and can be efficiently adapted at test-time to distill  task-specific visual information. 
    \item Extensive evaluation on VL-ICL Bench.\footnote{We only focus on
  single-image few-shot VQA tasks and leave the multi-image scenario for future work.} \citep{zong2025vlicl}, a diverse benchmark for image perception and mathematical reasoning,  demonstrates that our approach outperforms ICL and several other prompt distillation and parameter-efficient finetuning methods. 


\end{itemize}

\section{Related Work}

Our approach, Meta-Adaptive Prompt Distillation (MAPD), builds upon several existing research areas including few-shot learning, prompt tuning and test-time adaptation.

\textbf{Multimodal Few-shot Learning} Learning from a few examples has been a long-standing goal in machine learning. Early work by \cite{10.5555/3157382.3157504} introduced Matching Networks for one-shot image-to-text classification. This approach leverages a support set of labeled images to classify an unlabeled query image, laying the foundation for few-shot learning in vision tasks. With the advent of large language models (LLMs) and large multimodal models (LMMs; \citealt{alayrac2022flamingo,zhao2024mmicl}), in-context learning (ICL; \citealt{zhao2024mmicl,lester-etal-2021-power}) has emerged as a popular method for few-shot adaptation. ICL involves providing a few input-output examples directly in the model's prompt without updating its parameters. While this is a computationally inexpensive method, its performance for LMMs can be inconsistent \citep{zong2025vlicl} and may even degrade as more examples are added, particularly in smaller models. 

\textbf{Learning with Prompts} Another widely adopted approach for adapting  models to task-specific data is prompt optmization. \cite{wang-etal-2022-towards-unified} explored this idea with small language models ($\sim$ 0.1M) such as BERT for text classification. Subsequent work \citep{hou-etal-2022-metaprompting} introduced soft prompts to overcome the limitations of optimizing over discrete vocabulary tokens. \cite{Khattak_2023_ICCV} further proposed  PromptSRC for CLIP-based vision-language encoders, mitigating  overfitting of soft prompts. While these methods perform well on classification tasks, their extension to LLM-based architectures and problems such as  question-answering and mathematical reasoning remains limited.


\textbf{Test-Time Adaptation} These methods aim to dynamically adapt models during inference on test examples, that may have distributional differences from the training data. This adaptation can either involve training of model parameters \citep{hardt2024testtime} or can be entirely training free \citep{karmanov2024efficient}. Additionally, previous work \citep{hu2025testtime} has taken advantage of prompt tuning and other PEFT methods such as LoRA \citep{hu2022lora} to resolve catastrophic forgetting issues during test time training and achieve state-of-the-art performance.  \cite{shu2022testtime} propose  Test-time Prompt Tuning (TPT), a method that adapts vision-language models for zero-shot classification by tuning soft prompts on image augmentations. Previous work  \citep{najdenkoska2023meta,10376617} has also explored meta-learning of soft prompts for small models and a limited range of vision-language tasks such as fast-concept binding. 

We extend upon this idea to provide an alternative for few-shot adaptation in LMMs. Specifically, we design a meta-learning procedure, namely MAPD, to learn soft prompts that can distill task-relevant visual features from image embeddings and can be rapidly adapted at test time for a variety of new tasks using a few examples.
\cite{najdenkoska2023meta} rely on a single [CLS] token from CLIP's vision encoder, which limits the attention-mapper's capacity. We instead use the complete set of hidden patch features, enabling the attention-mapper to encode fine-grained visual information for distillation into soft prompts. We show that MAPD can be applied to any LMM architecture and achieves state-of-the-art performance on visual question answering tasks \citep{antol2015vqa}. 


\section{Problem Formulation}
\label{sec:prompt_tuning}

\subsection{Few-shot Visual Question Answering}
\label{sec:fsvqa}
Visual Question Answering (VQA; \citealt{antol2015vqa}) is a key task for  evaluating the ability of vision-language models to understand images by accurately responding to questions about various aspects of visual content. These questions can vary widely, ranging from descriptions of objects inside bounding boxes \citep{Krishna2017} to solving high-school geometry problems \citep{gao2025gllava},  but are mostly grounded in the visual information present in the image.

In VQA,  we typically have a dataset $\mathcal{D} = \{(X_v^i, X_q^i, X_a^i)\}_{i=1}^{|\mathcal{D}|}$ where $X_v \in \mathcal{I}$, $X_q \in \mathcal{Q}$ and $X_a \in \mathcal{A}$, and $\mathcal{I}$ is the set of all images, $\mathcal{Q}$  the set of all questions, and $\mathcal{A}$  the set of all answers. Our goal is to learn a function~$f_\theta$ parametrized by~$\theta$, that maximizes the  likelihood of the answer given the image and the question, $\prod_{i=1}^{|\mathcal{D}|} p_\theta(X_a^i | X_v^i, X_q^i)$. Following the standard train-test paradigm in deep learning, we evaluate whether $f_\theta$ generalizes well by dividing dataset~$D$ into $(D^{\text{train}}, D^{\text{test}})$ such that maximizing the above likelihood on $D^{\text{train}}$ also maximizes the likelihood of answer on $D^{\text{test}}$. A common assumption  is that the size of $D^{\text{train}}$ is large enough so that  function~$f_\theta$ does not overfit on~$D^{\text{train}}$. In the context of \emph{few-shot} VQA, we  treat the in-context examples (or shots) given to an LMM during ICL as~$D^{\text{train}}$. Since the examples in $D^{\text{train}}$ are limited (as few as 1-shot), it becomes harder to avoid overfitting while training  and still perform well on $D^{\text{test}}$. We conceptualize this problem as one of learning about an underlying task represented by $D^{\text{train}}$ and adopt meta-learning \citep{finn2017model} which exploits the shared structure across a  distribution of  tasks to  learn a prior over  model parameters, thereby enabling  stable transfer to new tasks with limited data.  In the following, we describe how we enforce this prior over parameters through the  curation of \textit{meta-tasks} containing few-shots. A sketch of  our  model architecture and training procedure  is shown in Figure~\ref{img:model}.

\begin{figure}[t]
\includegraphics[width=\textwidth,keepaspectratio]{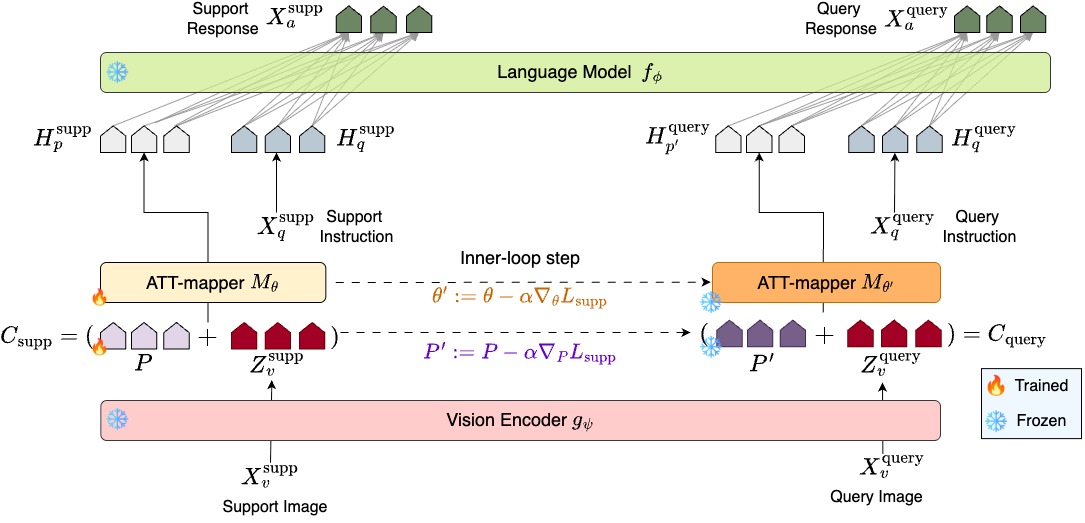}
\caption{Our proposed {MAPD} framework based on LLaVA v1.5-7B \citep{Liu_2024_CVPR}:  image embeddings are distilled into soft prompts~$P$ during instruction finetuning. The support set $(X_v^{\text{supp}}, X_q^{\text{supp}}, X_a^{\text{supp}})$ is processed initially to the obtain loss value $L_{\text{supp}}$ which is used in the inner-loop to obtain task-specific parameters $\{\theta', P'\}$. Next, the query set $(X_v^\text{{query}}, X_q^{\text{query}}, X_a^{\text{query}})$ is used to calculate the query loss for the outer-loop meta-parameter optimization $\{\theta, P\}$.}
\label{img:model}
\end{figure}

\subsection{Improving Task Understanding with Meta-tasks}
\label{sec:meta_tasks}
The core idea of optimization-based meta-learning is to learn a good initialization of parameters, which when finetuned on a specific task, enables  stable transfer for that task with a few gradient steps \citep{finn2017model}. To promote this capability,  training involves processing batches of few-shot datasets  that represent an underlying task. We refer to these few-shot datasets as \textit{meta-tasks} and propose to create them from our finetuning data mixture based on the original LLaVA datasets. We provide details of our  specific data mixture in Appendix~\ref{sec_app:fin_data}. 

More formally, let $p(\mathcal{D})$ denote our data mixture. We create  meta-task~$T^j$ by randomly sampling a fixed subset of VQA examples (image, question, answer triplets) from  dataset $D^i \sim p(\mathcal{D})$ and partitioning the examples further into support and query sets $T^j = \{D^{\text{supp}}, D^{\text{query}}\}$. To be consistent with the notation introduced in Section \ref{sec:fsvqa}, we treat the support set as $D^{\text{supp}} \equiv D^{\text{train}}$ and the query set as $D^{\text{query}} \equiv D^{\text{test}}$. We continue  this process until all samples from~$D^i$ have been assigned to at least one meta-task. This meta-task construction is performed for \emph{each dataset} in $p(\mathcal{D})$, resulting in meta-task distribution $p(\mathcal{T}^{{\text{meta}}})$. We now describe our model architecture designed to process these meta-tasks. Further details on the number and composition of meta-tasks for training and evaluation are provided in Appendix \ref{sec_app:meta_task}.

\subsection{Model Architecture}
\label{sec:arch}

We design our LMM architecture (Figure \ref{img:model}) based on the visual instruction tuning framework  of LLaVA v1.5\footnote{We adopt LLaVA v1.5 due to its simplicity and publicly available training code and datasets (Section \ref{sec_app:fin_data}). This prevents mixing between training and test datasets and enables evaluation over unseen tasks. We demonstrate in Section \ref{sec:abls} that our method can be easily applied to other LMM architectures.} \citep{Liu_2024_CVPR} and further describe our modifications for incorporating the attention-mapper. For clarity, we omit the distinction between support and query sets in this section as both are processed in the same manner. As shown in Figure~\ref{img:model}, the  model consists of a  pretrained CLIP ViT-L/14 visual encoder ($g_\psi$) with an aspect ratio of 336px; for an input image~$X_v$, the encoder gives us hidden visual features~$Z_{v}$ which are then  passed to the projection layer that consists of an  attention-mapper~$M_{\theta}$ responsible for extracting useful features from~$Z_v$.

\textbf{Attention Mapper}\hspace{.7em} We re-design the projection layer of LLaVA v1.5  to include soft prompts~$P$  by introducing an attention-mapper $M_\theta$ for  improved \emph{task-specific} feature extraction. Specifically, we prepend~$Z_v$ with a set of $m$~learnable prompt tokens~$P$ to obtain a sequence $C = (P, Z_v)$ which is then passed to the attention-mapper (see Figure~\ref{img:model}). Both  prompt tokens $P$ and  weights~$\theta$ are  initialized with  Xavier Uniform initialization \citep{glorot2010understanding}. We define the mapper as:
\begin{align}
H_{p+v} &= M_\theta(Q, K, V) = \sigma(QK^T)*V
\end{align}
where the query is $Q = M_\theta^q \cdot C\:;\:$, the key is $K = M_\theta^k \cdot C\:;\:$,  the value is $V = M_\theta^v \cdot C$,  and their corresponding matrices are~$\{M_\theta^q, M_\theta^k, M_\theta^v\}$. The mapper computes the dot product of the query and key vectors which are then passed to a softmax function to compute activation scores for every feature in vector $V$. Finally, we extract the first~$m$ embeddings corresponding to the learnable prompt tokens from the  set $H_{p+v}$ that correspond to the task-specific image embeddings~$H_p$. These are now passed to the LLM ($f_\phi$) as prompts for further processing. We denote the  trainable parameters for the attention-mapper with $\theta_p=\{\theta, P\}$.

\textbf{Language Model}\hspace{.7em}  The quality of the learned prompts highly depends on the underlying language model. We update the LLM of LLaVA v1.5 with the state-of-the-art Qwen2.5-7B-Instruct LLM,  which has demonstrated strong performance on complex tasks such as  mathematical reasoning and coding and supports the generation of up to 8K~tokens. The LLM ($f_\phi$) receives the concatenated sequence of image and text tokens to generate the answer $X_a = f_\phi([H_p, H_q])$.
In this pipeline, only the attention mapper parameters $\theta_p$ are trained, making our approach parameter-efficient for cross-task generalization. The number of trainable parameters is  approximately~24M (see Appendix~\ref{sec_app:model} for hyperparameters). The training objective maximizes the likelihood function, $p_{\theta_p}(X_a | X_v, X_q)$, parametrized by $\theta_p$,  where $X_a$ is the answer, $X_v$ is the image, and $X_q$ is the question. For clarity, we refer to this model, namely LLaVA-ATT-Qwen2.5 7B, as our base LMM in the following sections.

\subsection{Model Training}
\label{sec:training}
We train the attention mapper parameters to learn image-conditioned soft prompts in two stages following a curriculum learning procedure similar to  LLaVA v1.5 \citep{liu2023visual}. In the first-stage, which is  aimed at feature alignment, the attention-mapper is pretrained on the LCS-558K subset of the LAION/CC/SBU dataset filtered with a more balanced concept coverage \citep{liu2023visual}. Further details on pretraining are mentioned in Appendix~\ref{sec_app:train_det}. In the second stage, which aims to  distill task-specific image features into prompts~$H_p$, the attention-mapper parameters ~${\theta_p}$ are finetuned on  diverse task-specific instructions. We describe our MAML-based finetuning procedure below and also introduce alternative  methods which we compare against in our experiments. 




\subsubsection{Learning to Distill Prompts with First-order Meta Learning}

Our prompt distillation procedure, {MAPD}, uses the model-agnostic first-order approximation of MAML \citep{finn2017model} which aims to learn a robust initialization of meta-parameters that enable efficient adaptation to new tasks with just a few gradient updates.  We borrow the implementation of \cite{antoniou2018how} and use their first-order version and (learnable) per-step learning rates~($\alpha$) to further optimize the training process. We sample a batch $B$ of meta-tasks from $p(\mathcal{T}^{\text{meta}})$ and  use the support set of each task to convert $\theta_p$ into task specific parameters $\theta_p'$ with a few gradient steps. Equations~(\ref{eq:inner}) and (\ref{eq:inner_grad}) show a \emph{single} step of this inner loop:

\noindent\begin{minipage}{.5\linewidth}
\begin{align}
\label{eq:inner}
L^{\text{supp}}_{\theta_p} &= \dfrac{-1}{|D^{\text{supp}}|} \sum_{i=1}^{|D^{\text{supp}}|} \log(p_{\theta_p}(X_a^i | X_v^i, X_q^i))
\end{align}
\end{minipage}
\noindent\begin{minipage}{.5\linewidth}
\begin{align}
\label{eq:inner_grad}
\theta_p' &= \theta_p - \alpha \nabla_{\theta_p} L^{\text{supp}}_{\theta_p} 
\end{align}
\end{minipage}

The \emph{outer}  loop  involves optimizing the meta-parameters which in our case are the original attention-mapper parameters~$\theta_p$ on the query set using the task-specific parameters $\theta_p'$:

\noindent\begin{minipage}{.55\linewidth}
\begin{align}
L_{\theta_p'}^{\text{query}} &= \dfrac{-1}{|D^{\text{query}}|} \sum_{i=1}^{|D^{\text{query}}|}\log(p_{\theta_p'}(X_a^i | X_v^i, X_q^i))
\end{align}
\end{minipage}
\noindent\begin{minipage}{.45\linewidth}
\begin{align}
\label{eq:outer_grad}
\theta_p &:= \theta_p - \beta \sum_{j=1}^{|B|} \nabla_{\theta'_{p,j}} L^{\text{query}}_{\theta'_{p,j}} 
\end{align}
\end{minipage}

Equation~(\ref{eq:outer_grad}) is the first-order approximation of the meta-update in MAML \citep{finn2017model} that treats the gradient of $\theta'_{p,j}$ w.r.t. $\theta_p$ for a meta task as a constant. This approximation avoids backpropagating through the entire computation graph of the inner loop and avoids the Hessian-vector product estimation of the query loss. This saves huge GPU memory while still approximating a gradient in the same direction as the true MAML gradient \citep{lillog}. We provide a sketch of  MAPD training in Figure~\ref{img:model} and a more detailed algorithm in Appendix \ref{sec_app:psuedo_aglo} as Algorithm ~\ref{algo:fomaml}

\subsubsection{Alternative Methods for Prompt Distillation} 
\label{sec:alternative-methods}
We also implement other prompt distillation methods based on our model architecture to compare their performance with {MAPD} on few-shot VQA tasks. We provide a more formal description of these methods below, highlighting important differences from our framework.

\textbf{Multi-Task Prompt Distillation}\hspace{.7em}  We define a multi-task baseline where we eliminate the bi-level optimization of MAPD. Specifically, at each iteration, we sample a batch of meta-tasks from~$p(\mathcal{T}^{\text{meta}})$ and optimize the following loss per task:
\begin{align}
    L_{\theta_p} = \dfrac{-1}{N} \sum_{i=1}^{N} \log(p_{\theta_p}(X_a^i | X_v^i, X_q^i))
\end{align}
such that $N = |D^{\text{supp}}| + |D^{\text{query}}|$. This loss is accumulated across the entire batch of meta-tasks used to update~$\theta_p$. We refer to this baseline as {Multi-Task\textsuperscript{PD}}.

\textbf{In-Context Prompt Distillation}\hspace{.7em} Previous work \citep{chen-etal-2022-meta, min-etal-2022-metaicl} suggests it is possible to meta-learn task information by reducing the bi-level optimization of MAML to a sequence prediction problem over in-context examples with the help of pretrained LLMs. We develop a method called In-Context\textsuperscript{PD}, where we concatenate the support set with each query example in a meta-task, and optimize the following loss function to distill this task information from LLMs into soft prompts:
\begin{align}
    L_{\theta_p} = \dfrac{-1}{|D^{\text{query}}|} \sum_{i=1}^{|D^{\text{query}}|} \log(p_{\theta_p}(X_a^i | X_v^i, X_q^i, D^{\text{supp}}))
\end{align}

\textbf{Methods without Meta-tasks}\hspace{.7em} To further understand the benefit of curating meta-tasks (see Section~\ref{sec:meta_tasks}), we compare with the original finetuning procedure of LLaVA-v1.5 7B but only train $\theta_p$ without any meta-tasks for fair comparison.  We refer to this method as NoMeta-task$^{\text{PD}}$ in subsequent sections. We also compare with model averaging, which is computationally efficient and has been shown to increase performance on out-of-distribution datasets \citep{choshen2022fusingfinetunedmodelsbetter, wortsman2022model}.We separately finetune the attention-mapper on each dataset $D^i \sim p(\mathcal{D})$, and take an average of all  dataset-specific parameters $\theta_p^i$ weighted by their corresponding dataset size ratios:
\begin{align}
    \theta_p^{\text{avg}} = \sum_{i=1}^{|\mathcal{D}|} \theta_p^i \cdot w^i 
\end{align}
where $w^i = |D^i|\:/\:|\mathcal{D}|$. We refer to this baseline as {Model-Avg\textsuperscript{PD} in subsequent sections.

\subsection{Test-Time Adaptation} After learning optimal parameters with MAPD (and alternative distillation strategies), we adapt the attention-mapper to a new test task by fine-tuning for $K$~gradient steps. We empirically find that $K \leq 30$ is sufficient for all prompt distillation methods to converge, which we attribute to our adaptation procedure training only 24M parameters over a few examples. We further explain how this value is selected in Appendix~\ref{sec_app:test_time_adapt}. Concretely, given K $\leq$ 30 steps, we perform task-specific finetuning of the parameters~$\theta_p$ on the support set~$D^{\text{supp}}_{\text{test}}$ of test task $T^j_{\text{test}}$, using the inner-loop optimizer mentioned in equation \ref{eq:inner_grad}. We then evaluate model performance on the query set $D^{\text{query}}_{\text{test}}$ for that task.

\section{Experimental Results}
\label{exps}

\subsection{Evaluation Datasets}

For evaluation purposes, our test datasets follow the same structure as the meta-tasks introduced in Section~\ref{sec:meta_tasks}, with support and query examples. 
We use the recently introduced VL-ICL benchmark \citep{zong2025vlicl}, designed to test the ICL capabilities of LMMs on various tasks like fast concept binding, multimodal reasoning, and fine-grained perception.
Meta-tasks for testing are created by randomly sampling a support set from the training split of the VL-ICL datasets and a test/query set from their respective testing splits.\footnote{We also keep a separate validation set for each VL-ICL dataset (sampled from the training split) to select the best  model  which we then evaluate on the test (query) set. 
More details can be found in Section~\ref{sec_app:test_time_adapt}} In line with our training pipeline, which exclusively utilizes datasets containing a single image per example (see Section~\ref{sec_app:fin_data}), we focus solely on single image-to-text scenarios, leaving multi-image cases for future work.  We report results on four tasks from VL-ICL: a) \emph{Fast Open MiniImageNet (Open-MI)}, where the model must name new objects based on a few examples; b) \emph{Operator Induction}, where the model must solve image tasks of the type $2\:?\:7 = ?$ given training examples like $1\:?\:3 = 4$; c) \emph{CLEVR Count Induction}, where the model must count objects that satisfy given attributes like "\textit{shape: sphere}"; and d) \emph{TextOCR}, where the model
must transcribe highlighted text contained in an image. We provide more details on these tasks in Appendix~\ref{sec_app:eval_data}. 
The final model performance is calculated as the average across all meta-tasks.

\begin{table}[t]
\centering
\begin{adjustbox}{width=\linewidth}
  \begin{tabular}{l|c|c|c|c|c}
    \toprule
   {Methods} & {MT} & {Open-MI} & {OP\_IND} & CLEVR & TextOCR \\
    \rowcolor{gray!20} 
    \midrule
    \textbf{TTA with ICL} & & & & & \\
    NoMeta-task$^{\text{PD}}$ & \xmark & ${43.8 \pm 0.9}$ & ${12.1 \pm 0.6}$ & ${18.0 \pm 0.2}$ & $6.8 \pm 0.4$ \\
    Model-Avg$^{\text{PD}}$ & \xmark & ${26.6 \pm 0.7}$ & ${9.2 \pm 0.5}$ & ${7.6 \pm 0.1}$ & $2.8 \pm 0.3$ \\
    In-Context$^{\text{PD}}$ & \cmark & ${51.1 \pm 0.9}$ & ${20.6 \pm 0.8}$ & ${24.1 \pm 0.2}$ & $23.8 \pm 0.3$ \\
    Multi-Task$^{\text{PD}}$ & \cmark & ${48.6 \pm 0.9}$ & ${10.0 \pm 0.6}$ & ${12.5 \pm 0.1}$ & $6.9 \pm 0.4$ \\
    MAPD & \cmark & ${53.3 \pm 0.9}$ & ${9.60 \pm 0.5}$ & ${12.3 \pm 0.1}$ & $7.30 \pm 0.4$ \\
    \midrule
    \rowcolor{gray!20}
    \textbf{TTA with FT $\leq$ 30} & {} & {} & {} & {} & {}\\
    NoMeta-task$^{\text{PD}}$ & \xmark & ${68.0 \pm 0.8}$ & ${38.8 \pm 0.6}$ & ${25.8 \pm 0.2}$ & $22.5 \pm 0.3$ \\
    Model-Avg$^{\text{PD}}$ & \xmark & ${63.1 \pm 0.8}$ & ${40.0 \pm 0.6}$ & ${29.1 \pm 0.2}$ & $21.5 \pm 0.3$ \\
    In-Context$^{\text{PD}}$ & \cmark & ${64.5 \pm 0.8}$ & ${30.9 \pm 0.5}$ & ${30.9 \pm 0.2}$ & $18.9 \pm 0.3$ \\
    Multi-Task$^{\text{PD}}$ & \cmark & ${74.6 \pm 0.7}$ & ${45.1 \pm 0.5}$ & ${29.9 \pm 0.2}$ & $22.9 \pm 0.4$ \\
    \textbf{MAPD} & \cmark & $\textbf{77.9} \pm \textbf{0.7}$ & $\textbf{47.7} \pm \textbf{0.5}$ & $\textbf{31.4} \pm \textbf{0.2}$ & $\textbf{26.4} \pm \textbf{ 0.5}$ \\
    \bottomrule
  \end{tabular}
  \end{adjustbox}
  \caption{\label{tab:main_results}Evaluation on tasks from the VL-ICL Bench \citep{zong2025vlicl} with LLaVA-ATT-Qwen2.5 7B as the base LMM. Each method trains 24M attention-mapper parameters. We report the mean accuracy across shots $\{1,2,4,5,8\}$ with 95\% binomial confidence intervals and compare different prompt distillation approaches. TTA:Test-Time Adaptation, FT: Finetuning with $K\leq30$ gradient steps, ICL: In-Context Learning, MT: Meta-Tasks used (\cmark) or not (\xmark) during training. Qualitative results are in Appendix \ref{sec_app:quality} and \ref{sec_app:op_abls}.}
\end{table}

 

\subsection{Model Comparisons}
\label{result}


Our results are summarized in Table~\ref{tab:main_results}, which compares MAPD against alternative prompt distillation methods (see Section~\ref{sec:alternative-methods}) and reports the mean accuracy of up to eight shots. We compare two types of test-time adaptation methods, namely in-context learning (ICL) which prompts the underlying LLM with no distillation of image embeddings and finetuning (FT) with K $\leq$ 30 gradient steps, which are further distinguished based on whether they use meta-tasks during training. Results for individual shots are in Appendix~\ref{sec_app:det_result};  additional results for more shots are in Appendix~\ref{sec_app:more_shots}\footnote{We also provide ICL performance of publicly available models in Appendix \ref{sec_app:llava_compare} for reference.}.

\textbf{Prompt distillation improves task induction in LMMs at test-time.}\hspace{.7em} Our results in Table~\ref{tab:main_results} show that FT adaptation with few-shots (support  examples) largely outperforms ICL at test time evaluated over query examples, with an average increase~of~21.2\%  over all datasets. These results highly support our hypothesis that  distilling task-specific information from image embeddings to create targeted prompts  improves the few-shot capabilities of the underlying LLM (in our case  Qwen-2.5-7B-Instruct). 
\begin{wraptable}[11]{r}{0.7\textwidth}
  \resizebox{\linewidth}{!}{
  \begin{tabular}{l|c|c|c|c}
    \toprule
    LoRA & {Open-MI} & {OP\_IND} & CLEVR & TextOCR \\
    \midrule
    \rowcolor{gray!20}
    \textbf{TTA with FT $\leq$ 30} & {} & {} & {} & {}\\
    All \hspace*{1.2em}LLM layers & ${55.1 \pm 0.7}$ & ${13.3 \pm 0.6}$ & ${15.1 \pm 0.2}$ & ${10.4 \pm 0.4}$ \\
    \text{[0-15]} LLM layers & ${67.3 \pm 0.8}$ & ${25.5 \pm 0.6}$ & $\textbf{30.0} \pm \textbf{0.2}$ & ${23.8 \pm 0.3}$ \\
     \text{[0-15]} LLM layers + ATT & $\textbf{69.1} \pm \textbf{0.8}$ & $\textbf{30.5} \pm \textbf{0.5}$ & ${28.7 \pm 0.2}$ & $\textbf{24.5} \pm \textbf{0.3}$ \\
    \bottomrule
  \end{tabular}}
  \vspace{-1ex}
  \caption{\label{tab:lora_results}LoRA configurations for base LMM and evaluated on the VL-ICL bench. We report the mean accuracy across shots $\{1,2,4,5,8\}$ with 95\% binomial confidence intervals. ATT: Attention-Mapper; TTA: Test-Time Adaptation; FT: Finetuning with $K \leq 30$ gradient steps.}
\end{wraptable}
Additionally, our results show that finetuning just the attention-mapper parameters only requires a few gradient steps ($K\leq30$) at test-time to generalize to unseen tasks and does not lead to overfitting over the support examples (Appendix~\ref{sec_app:test_time_adapt}). 
For a one-to-one comparison, we look into In-context\textsuperscript{PD}, which performs better with FT on 3 out of 4 tasks compared to its ICL adaptation and enables prompting the underlying LLM with a fixed set of learned task-specific embeddings. 


\begin{figure}[t]
\centering
\begin{subfigure}[t]{0.35\textwidth}
\vspace{0pt}\centering
\begin{tikzpicture}[scale=.6]
\begin{axis}[
    width=10cm,
    height=7cm,
    xlabel={Number of Soft Prompts~$P$ (shown in $\log_2(P)$)},
    ylabel={Mean Accuracy},
    ylabel style={font=\normalsize,yshift=-.2cm},
    xmin=2, xmax=8,
    ymin=15.0, ymax=60.0,
    xtick={2,4,6,8},
    ytick={20.0, 25.0, 30.0, 35.0, 40.0, 45.0, 50.0, 55.0, 60.0},
    grid=both,
    axis lines=box,
    legend style={
        font=\small,
        at={(0.5,1.03)},
        anchor=south,
        legend columns=4
        minimum width=7cm
    },
    tick label style={font=\small},
    title style={font=\small},
    label style={font=\small},
    every node near coord/.append style={font=\footnotesize, yshift=5pt},
    enlargelimits=0.05,
    line width=1.5pt,
    mark size=2pt,
    mark options={solid},
    axis lines=left,
]
\addplot[
  color=blue,
    mark=*,
    thick
] coordinates {
    (2, 24.45)
    (4, 27.50)
    (6, 31.13)
    (8, 31.88)
};

\addplot[
  color=orange,
    mark=*,
    thick
] coordinates {
    (2, 30.42)
    (4, 33.63)
    (6, 35.92)
    (8, 42.58)
};

\addplot[
  color=green!50!black,
    mark=*,
    thick
] coordinates {
    (2, 35.75)
    (4, 40.80)
    (6, 45.17)
    (8, 52.70)
};

\addplot[
  color=red,
    mark=*,
    thick
] coordinates {
    (2, 36.92)
    (4, 45.83)
    (6, 50.10)
    (8, 56.63)
};

\addplot[
  color=blue,
    mark=*,
    mark size = 1.7pt,
    dashed
] coordinates {
    (2, 16.54)
    (4, 19.16)
    (6, 20.36)
    (8, 23.5)
};
\addplot[
  color=orange,
    mark=*,
    mark size = 1.7pt,
    dashed
] coordinates {
    (2, 26.0)
    (4, 30.65)
    (6, 27.42)
    (8, 26.63)
};
\addplot[
  color=green!50!black,
    mark=*,
    dashed,
    mark size = 1.7pt
] coordinates {
    (2, 33.38)
    (4, 35.63)
    (6, 33.07)
    (8, 31.25)
};
\addplot[
  color=red,
    mark=*,
    mark size = 1.7pt,
    dashed
] coordinates {
    (2, 32.83)
    (4, 37.83)
    (6, 35.79)
    (8, 36.1)
};

\legend{1-Shot(M), 2-Shot(M), 4-Shot(M), 5-Shot(M), 1-shot(I), 2-shot(I), 4-shot(I), 5-shot(I)}
\end{axis}
\end{tikzpicture}

\end{subfigure}\hfill
\begin{subfigure}[t]{0.55\textwidth}
\vspace{0pt}\centering
 \begin{tikzpicture}[scale=.6]
\begin{axis}[
    ybar,
    bar width=11pt,
    width=13cm,
    height=7cm,
    enlarge x limits=0.15,
    ymin=0,
    ymax=100,
    ylabel={Mean Score (\%)},
    ylabel style={font=\normalsize,yshift=-.2cm},
    xtick=data,
    xticklabel style={align=center, font=\small},
    yticklabel style={font=\small},
    xticklabels={
        {Original\\(EM)},
        {Task Induction\\(EM)},
        {Perception\\(EM)},
        {Math Reasoning\\(Qwen-VL Score)}
    },
    legend style={
        font=\small,
        at={(0.5,1.15)},
        anchor=north,
        legend columns=5
    },
    grid=major,
    grid style={dashed,gray!20},
    nodes near coords,
    every node near coord/.append style={font=\fontsize{6}{5}\selectfont,
      yshift=0pt,text=black}
]

\addplot+[style={fill=blue!50, draw=black}] plot coordinates {(0,36.1) (1,75.5) (2,63.3) (3,83.9)};
\addplot+[style={fill=red!60, draw=black}]  plot coordinates {(0,27.2) (1,55.5) (2,76.9) (3,65.1)};
\addplot+[style={fill=cyan!60, draw=black}]  plot coordinates {(0,37.3) (1,64.4) (2,59.2) (3,85.6)};
\addplot+[style={fill=green!60, draw=black}] plot coordinates {(0,39.8) (1,77.8) (2,77.8) (3,86.5)};
\addplot+[style={fill=orange!70, draw=black}] plot coordinates {(0,44.1) (1,89.5) (2,82.5) (3,91.5)};

\legend{NoMeta-task\textsuperscript{PD}, In-context\textsuperscript{PD}, Model-Avg\textsuperscript{PD}, Multi-Task\textsuperscript{PD}, MAPD}
\end{axis}
\end{tikzpicture}
  \end{subfigure}
  \caption{(a) \textbf{Left}: Performance comparison between MAPD+FT (M) and In-Context\textsuperscript{PD}+ICL (I). Mean Accuracy is computed across all  VL-ICL datasets. We consider different prompt token lengths   $P=\{4,16,64,256\}$ which are shown in $\log_2(P)$ scale for different shots. (b) \textbf{Right}: Performance of different prompt distillation methods on three Operator Induction subtasks: Task Induction, Perception, and Math Reasoning. We report mean exact-match (EM; $\%$) for 1,2 and 8-shots as defined in the VL-ICL Bench \citep{zong2025vlicl} except for  Mathematical Reasoning, which uses mean ratings generated by Qwen-2.5-VL-32B-Instruct. More details  can be found in Appendix \ref{sec_app:op_abls}}
\label{fig:other_analysis}
\end{figure}


\textbf{Meta-learning and meta-tasks improve few-shot learning.}\hspace{.7em} Table~\ref{tab:main_results} shows that methods using meta-tasks are indeed superior. For ICL-based adaptation, In-Context\textsuperscript{PD} performs best, while for FT-based adaptation, our proposed approach, MAPD, achieves the best overall performance across all four datasets at test time. 
This further suggests that first-order MAML learns the best initialization of attention-mapper parameters~$\theta_p$. These parameters are subsequently adapted for a test task with a few gradient steps and few-shot examples to produce a precise set of soft prompts that improves LMM predictions on that task. Our detailed results in Table \ref{tab:baseline} in Appendix~\ref{sec_app:det_result} further show that for FT-based adaptation, MAPD is most effective in the 2-shot case for Operator Induction, surpassing Multi-Task\textsuperscript{PD} by~10\%. Finally, MAPD with FT is the only approach that exhibits strictly monotonic improvements as the number of shots increases, showing better scaling behavior.

\textbf{MAPD surpasses other efficient finetuning approaches for few-shot adaptation.} We compare MAPD with  LoRA \citep{hu2022lora}, a state-of-the-art parameter-efficient finetuning (PEFT) approach. In Table~\ref{tab:lora_results}, we integrate LoRA in the base LMM  in three configurations and evaluate on VL-ICL: (1)~naively applying LoRA to all underlying LLM layers (as done in LLaVA v1.5; \citealt{liu2023visual}) increases the number of trainable parameters ($\sim$ 300M) and the model finds it difficult to converge within 30~gradient steps at test-time;  (2)~restricting LoRA to the first 16~LLM layers  ($\sim$ 24M parameters)  offers better test-time performance; and (3)~adding LoRA to the attention-mapper layers further boosts performance as it provides some distillation over the image embeddings before prompting the underlying LLM. Ultimately, MAPD still outperforms the best LoRA configuration by an average of 7.7\% across all VL-ICL datasets. This demonstrates that MAPD is the  best choice for achieving fast test-time adaptation in low-data scenarios.  We provide additional LoRA training details in Appendix \ref{sec_app:train_det} and further detailed results can be found in Appendix \ref{sec_app:det_result}.

\subsection{Ablation Studies and Analysis}
\label{sec:abls}
\begin{wrapfigure}[13]{R}{0.52\textwidth}
\hspace{-.2cm}\begin{tikzpicture}[scale=.7]
\begin{axis}[
    ybar,
    bar width=12pt,
    width=11cm,
    height=5cm,
    enlarge x limits=0.15,
    ymin=0,
    ymax=100,
    ylabel={Mean Score (\%)},
    ylabel style={font=\normalsize,yshift=-.2cm},
    xtick=data,
    xticklabel style={align=center, font=\small},
    yticklabel style={font=\small},
    xticklabels={
        {TextOCR},
        {CLEVR},
        {Operator\\Induction},
        {OPEN\_MI}
    },
    legend style={
        font=\small,
        at={(0.5,1.1)},
        anchor=north,
        legend columns=4
    },
    grid=major,
    grid style={dashed,gray!20},
    nodes near coords,
    every node near coord/.append style={font=\fontsize{7}{5}\selectfont,
      yshift=0pt,text=black}
]
\addplot+[style={fill=Goldenrod!60, draw=black}] plot coordinates {(0,18.3) (1,20.3) (2,18.4) (3,53.5)};
\addplot+[style={fill=RedViolet!60, draw=black}] plot coordinates {(0,14.1) (1,29.2) (2,25.5) (3,62.3)};
\addplot+[style={fill=BlueGreen!50, draw=black}] plot coordinates {(0,19.6) (1,15.1) (2,27.3) (3,56.5)};
\addplot+[style={fill=YellowGreen!80, draw=black}]  plot coordinates {(0,26.4) (1,31.4) (2,47.7) (3,77.9)};
\legend{MLP, SP+MLP, ATT, SP+ATT}
\end{axis}
\end{tikzpicture}
\vspace{-2.5ex}
\captionof{figure}{Projection layer architectures in the  base LMM. SP: Soft Prompts, ATT: Attention-Mapper, MLP: 2-layer MLP (originally used in LLaVA v1.5).} \label{fig:mapper} 
\end{wrapfigure}

In this section, we present ablation studies across various model architectures and sizes, along with a more in-depth analysis of the benefits of test-time fine-tuning using MAPD. Please refer to appendix for further ablations on testing 1) robustness to image perturbations (Appendix \ref{sec_app:img_perturb}) and 2) different few-shot selection strategies (Appendix \ref{sec_app:sel_criteria}).

\textbf{What are the benefits of the attention mapper and soft prompts?} In Figure~\ref{fig:mapper}, we compare different designs for the projection layer in the base LMM for rapid few-shot learning.  We clearly see that MAPD benefits most by incorporating  the attention-mapper and soft prompts (SP+ATT). We draw two key conclusions from this experiment: (1)~distilling task-relevant information from CLIP embeddings with soft prompts yields substantial improvements, with an average gain of 16.3\% across architectures. (2)~replacing the 2-layer MLP used in LLaVA v1.5 with an attention mapper leads to an additional average gain of 13.1\%, thanks to its inherent weighting mechanism of pairwise similarities over CLIP embeddings.


\textbf{How does the number of soft prompts affect performance?} We examine how MAPD’s performance changes with the number of soft prompts across varying shot settings for VL-ICL datasets in Figure \ref{fig:other_analysis}(a). Additionally, we show results of our best ICL approach, In-Context\textsuperscript{PD},  as a baseline for this comparison. We see that MAPD scales favorably and learns more consistent task information from gradient updates at test time as the number of soft prompts is increased. Furthermore, its marginal improvement per added prompt token is substantially greater when more shots are provided. In contrast, the performance of In-Context\textsuperscript{PD} generally deteriorates with more prompts and struggles to jointly attend to more examples and longer prompts. We also investigate this further by analysing the softmax attention values of the underlying LLM for soft prompt embeddings containing image information in Appendix \ref{sec_app:att_entropy}. We note that the attention entropy decreases as the context length grows for In-Context\textsuperscript{PD}. This highlights its inherent limitation in being unable to attend to all the soft prompts when answering a query, while MAPD consistently attends to all the soft prompts.

use of a fixed number of soft prompts is superior to In-Context\textsuperscript{PD}, where the number of soft prompts vary according to the number of shots, .

\textbf{To what extent does MAPD facilitate task understanding at test time?} We take a closer look at how effectively MAPD captures task understanding at test time, using the Operator Induction task (See Figure~\ref{img:op_ind}) as a case study. To solve this task, the model should correctly (a) identify the operands in the query example (\emph{Perception}); (b) identify the operation from few-shot examples (\emph{Task Induction}); and (c) use its own mathematical knowledge over the identified elements to reason towards the answer (\emph{Mathematical Reasoning}). To test whether the model understands these subtasks, we design specific prompts and modify query examples as listed in Appendix~\ref{sec_app:op_abls}. In Figure~\ref{fig:other_analysis}(b), we observe that MAPD outperforms other prompt distillation approaches on all three subtasks, leading to better performance on the original task. MAPD shows a major improvement for task induction with an increase of 11.7\% compared to MultiTask\textsuperscript{PD}. We also observe that solving each subtask individually is easier than tackling the original task, as the latter requires integrating knowledge from all subtasks, which is challenging when only a few shots are available at test time. Finally, MAPD excels at mathematical reasoning, effectively utilizing the underlying LLM's reasoning capabilities. 


\begin{table}[t]
  
  \centering
  \begin{adjustbox}{width=\linewidth}
  \begin{tabular}{cccccccc}
    \toprule
     Vision Encoder & LLM & TTA & NoMeta-task$^{\text{PD}}$ & Model-Avg$^{\text{PD}}$ & In-Context$^{\text{PD}}$ & Multi-Task$^{\text{PD}}$ & MAPD \\
    \midrule
      \multirow{-1}{*}{\cellcolor[gray]{.9} CLIP ViT-L/14} & \multirow{-1}{*}{\cellcolor[gray]{.9}\shortstack{Qwen2.5-7B}} & \cellcolor[gray]{0.9}ICL & \cellcolor[gray]{0.9}$43.8 \pm 0.9$ & \cellcolor[gray]{0.9}$26.6 \pm 0.7$ & \cellcolor[gray]{0.9}$51.1 \pm 0.9$& \cellcolor[gray]{0.9}$48.6 \pm 0.9$ & \cellcolor[gray]{0.9}$\textbf{53.3} \pm \textbf{0.9}$ \\
     \cellcolor[gray]{0.9} &  \cellcolor[gray]{0.9}{Instruct} & \cellcolor[gray]{0.9}FT$\leq$30 & \cellcolor[gray]{0.9}$68.0 \pm 0.8$ & \cellcolor[gray]{0.9}$63.1 \pm 0.8$ & \cellcolor[gray]{0.9}$64.5 \pm 0.8$ & \cellcolor[gray]{0.9}$74.6 \pm 0.7$ & \cellcolor[gray]{0.9}$\textbf{77.9} \pm \textbf{0.7}$ \\
    \midrule
     \multirow{2}{*}{CLIP ViT-L/14} & \multirow{2}{*}{\shortstack{Qwen2.5-3B\\Instruct}} & ICL & $24.3 \pm 0.7$ & $30.5 \pm 0.7$ & $\textbf{48.3} \pm \textbf{0.9}$ & $39.1 \pm 0.7$ & $32.9 \pm 0.7$ \\
    {} & {} &  FT$\leq$30 & $56.5 \pm 0.9$ & $66.0 \pm 0.5$ & $47.5 \pm 0.9$ & $61.1 \pm 0.8$ & $\textbf{67.3} \pm \textbf{0.6}$ \\
    \midrule
    \multirow{2}{*}{CLIP ViT-L/14} & \multirow{2}{*}{\shortstack{Vicuna v1.5-7B}} & ICL & $20.0 \pm 0.7$ & $26.2 \pm 0.7$ & $46.3 \pm 0.9$ & $29.1 \pm 0.7$ & $\textbf{49.9} \pm \textbf{0.9}$ \\
    {} & {} & FT$\leq$30 & $69.1 \pm 0.8$ & $74.9 \pm 0.4$ & $66.7 \pm 0.8$ & $70.3 \pm 0.7$ & $\textbf{75.8} \pm \textbf{0.4}$  \\
    \midrule
    \multirow{2}{*}{SigLIP-SO400M} & \multirow{2}{*}{\shortstack{Qwen2.5-7B\\Instruct}} & ICL & $42.6 \pm 0.9$ & $40.7 \pm 0.9$ & $47.3 \pm 0.9$ & $\textbf{50.0} \pm \textbf{0.9}$ & $43.6 \pm 0.9$ \\
    {} & {} &  FT$\leq$30 & $52.0 \pm 0.9$ & $56.5 \pm 0.8$ & $56.0 \pm 0.8$ & $59.3 \pm 0.5$ & $\textbf{60.5} \pm \textbf{0.5}$ \\
    \midrule
    \multirow{2}{*}{CLIP ViT-L/14} & \multirow{2}{*}{\shortstack{Qwen3-8B}} & ICL & $55.0 \pm 0.9$ & $48.5 \pm 0.9$ & $\textbf{63.5} \pm \textbf{0.7}$ & $57.6 \pm 0.5$ & $60.3 \pm 0.5$ \\
    {} & {} &  FT$\leq$30 & $72.3 \pm 0.9$ & $69.1 \pm 0.7$ & $71.4 \pm 0.9$ & $80.4 \pm 0.6$ & $\textbf{83.5} \pm \textbf{0.6}$ \\
    \bottomrule
  \end{tabular}
  \end{adjustbox}
  \caption{Comparison of prompt distillation approaches under different LMM settings while keeping the attention-mapper and soft prompts fixed. We report mean accuracy across 1 to 5 shots with 95\% binomial confidence intervals for the OPEN\_MI benchmark. The original  LLaVA-ATT-Qwen2.5 7B  architecture is highlighted in gray. FT: Finetuning with K $\leq$ 30 gradient steps, ICL: In-Context Learning, TTA:  Test-Time Adaptation.  NoMeta-task$^{\text{PD}}$ and Model-Avg$^{\text{PD}}$ do not use meta-tasks.}
  \label{tab:baseline_avg}
\end{table}

\textbf{Does MAPD generalize across different LMM architectures?} We next examine different LMM architectures that affect MAPD's performance. Specifically, we report results in four settings that vary the underlying LLM and vision encoder: a)~using a smaller LLM (Qwen2.5-3B-Instruct); b)~using a different and relatively weaker LLM (Vicuna v1.5-7B); c) using a different vision encoder, SigLIP \citep{zhai2023sigmoidlosslanguageimage}; and d) and using a relatively powerful LLM (Qwen3-8B) \citep{yang2025qwen3technicalreport}. In Table~\ref{tab:baseline_avg},  MAPD outperforms other baselines with FT adaptation across different model configurations on the Open\_MI task, demonstrating its robustness and generalizability. Fine-tuning based test-time adaptation for prompt distillation substantially outperforms ICL, with average improvements of +24.6, +37.06, +12.02, and +18.04 across the four settings, respectively. This highlights the significant benefits of test-time prompt distillation.

\section{Conclusion}
This work introduced Meta-Adaptive Prompt Distillation (MAPD), 
a novel meta-learning approach that endows LMMs with few-shot capabilities. MAPD employs a fixed set of soft prompts, distilled from task-relevant image features, which can be adapted at test time using only a few examples. A key component of our method is an attention-mapper module, which can be flexibly integrated with any LMM architectures and  is jointly learned with soft prompts to facilitate distillation. Extensive evaluation on the VL-ICL benchmark shows that MAPD consistently outperforms traditional ICL and other efficient finetuning approaches across a diverse range of VQA tasks. 
Additional analysis (Appendix~\ref{sec_app:tta_compute}) shows that MAPD incurs higher per-query latency than ICL due to gradient-based adaptation, but scales more favorably with increased test-time compute budget and is more data-efficient. Future work could focus on improving MAPD's computational efficiency for resource-constrained scenarios and extending it to multi-image tasks and complex reasoning problems.
 


\section*{Acknowledegments}
We thank the anonymous reviewers for their feedback. We gratefully acknowledge the support of the UK Engineering and Physical Sciences Research Council (grant EP/W002876/1).

\bibliography{main}

@article{GRIFFITHS201924,
title = {Doing more with less: meta-reasoning and meta-learning in humans and machines},
journal = {Current Opinion in Behavioral Sciences},
volume = {29},
pages = {24-30},
year = {2019},
note = {Artificial Intelligence},
issn = {2352-1546},
doi = {https://doi.org/10.1016/j.cobeha.2019.01.005},
url = {https://www.sciencedirect.com/science/article/pii/S2352154618302122},
author = {Thomas L Griffiths and Frederick Callaway and Michael B Chang and Erin Grant and Paul M Krueger and Falk Lieder}
}

@book{finn2018learning,
  title={Learning to learn with gradients},
  author={Finn, Chelsea B},
  year={2018},
  publisher={University of California, Berkeley}
}

@inproceedings{
kirsch2022selfreferential,
title={Self-Referential Meta Learning},
author={Louis Kirsch and J{\"u}rgen Schmidhuber},
booktitle={First Conference on Automated Machine Learning (Late-Breaking Workshop)},
year={2022},
url={https://openreview.net/forum?id=WAcLlCixQP7}
}

@inproceedings{finn2017model,
  title={Model-agnostic meta-learning for fast adaptation of deep networks},
  author={Finn, Chelsea and Abbeel, Pieter and Levine, Sergey},
  booktitle={International conference on machine learning},
  pages={1126--1135},
  year={2017},
  organization={PMLR}
}

@inproceedings{
liu2023visual,
title={Visual Instruction Tuning},
author={Haotian Liu and Chunyuan Li and Qingyang Wu and Yong Jae Lee},
booktitle={Thirty-seventh Conference on Neural Information Processing Systems},
year={2023},
url={https://openreview.net/forum?id=w0H2xGHlkw}
}

@misc{laurencon2024mattersbuildingvisionlanguagemodels,
      title={What matters when building vision-language models?}, 
      author={Hugo Lauren\c{c}on and L\'{e}o Tronchon and Matthieu Cord and Victor Sanh},
      year={2024},
      eprint={2405.02246},
      archivePrefix={arXiv},
      primaryClass={cs.CV},
      url={https://arxiv.org/abs/2405.02246}, 
}

@inproceedings{NEURIPS2021_01b7575c,
 author = {Tsimpoukelli, Maria and Menick, Jacob L and Cabi, Serkan and Eslami, S. M. Ali and Vinyals, Oriol and Hill, Felix},
 booktitle = {Advances in Neural Information Processing Systems},
 editor = {M. Ranzato and A. Beygelzimer and Y. Dauphin and P.S. Liang and J. Wortman Vaughan},
 pages = {200--212},
 publisher = {Curran Associates, Inc.},
 title = {Multimodal Few-Shot Learning with Frozen Language Models},
 url = {https://proceedings.neurips.cc/paper_files/paper/2021/file/01b7575c38dac42f3cfb7d500438b875-Paper.pdf},
 volume = {34},
 year = {2021}
}

@InProceedings{Liu_2024_CVPR,
    author    = {Liu, Haotian and Li, Chunyuan and Li, Yuheng and Lee, Yong Jae},
    title     = {Improved Baselines with Visual Instruction Tuning},
    booktitle = {Proceedings of the IEEE/CVF Conference on Computer Vision and Pattern Recognition (CVPR)},
    month     = {June},
    year      = {2024},
    pages     = {26296-26306}
}

@misc{li2023ottermultimodalmodelincontext,
      title={Otter: A Multi-Modal Model with In-Context Instruction Tuning}, 
      author={Bo Li and Yuanhan Zhang and Liangyu Chen and Jinghao Wang and Jingkang Yang and Ziwei Liu},
      year={2023},
      eprint={2305.03726},
      archivePrefix={arXiv},
      primaryClass={cs.CV},
      url={https://arxiv.org/abs/2305.03726}, 
}

@inproceedings{
zhao2024mmicl,
title={{MMICL}: Empowering Vision-language Model with Multi-Modal In-Context Learning},
author={Haozhe Zhao and Zefan Cai and Shuzheng Si and Xiaojian Ma and Kaikai An and Liang Chen and Zixuan Liu and Sheng Wang and Wenjuan Han and Baobao Chang},
booktitle={The Twelfth International Conference on Learning Representations},
year={2024},
url={https://openreview.net/forum?id=5KojubHBr8}
}

@inproceedings{
zong2025vlicl,
title={{VL}-{ICL} Bench: The Devil in the Details of Multimodal In-Context Learning},
author={Yongshuo Zong and Ondrej Bohdal and Timothy Hospedales},
booktitle={The Thirteenth International Conference on Learning Representations},
year={2025},
url={https://openreview.net/forum?id=cpGPPLLYYx}
}

@inproceedings{
coda-forno2023metaincontext,
title={Meta-in-context learning in large language models},
author={Julian Coda-Forno and Marcel Binz and Zeynep Akata and Matthew Botvinick and Jane X Wang and Eric Schulz},
booktitle={Thirty-seventh Conference on Neural Information Processing Systems},
year={2023},
url={https://openreview.net/forum?id=sx0xpaO0za}
}

@inproceedings{hendel-etal-2023-context,
    title = "In-Context Learning Creates Task Vectors",
    author = "Hendel, Roee  and
      Geva, Mor  and
      Globerson, Amir",
    editor = "Bouamor, Houda  and
      Pino, Juan  and
      Bali, Kalika",
    booktitle = "Findings of the Association for Computational Linguistics: EMNLP 2023",
    month = dec,
    year = "2023",
    address = "Singapore",
    publisher = "Association for Computational Linguistics",
    url = "https://aclanthology.org/2023.findings-emnlp.624/",
    doi = "10.18653/v1/2023.findings-emnlp.624",
    pages = "9318--9333"
}

@inproceedings{
huang2024multimodal,
title={Multimodal Task Vectors Enable Many-Shot Multimodal In-Context Learning},
author={Brandon Huang and Chancharik Mitra and Leonid Karlinsky and Assaf Arbelle and Trevor Darrell and Roei Herzig},
booktitle={The Thirty-eighth Annual Conference on Neural Information Processing Systems},
year={2024},
url={https://openreview.net/forum?id=W0okTgsPvM}
}

@inproceedings{lester-etal-2021-power,
    title = "The Power of Scale for Parameter-Efficient Prompt Tuning",
    author = "Lester, Brian  and
      Al-Rfou, Rami  and
      Constant, Noah",
    editor = "Moens, Marie-Francine  and
      Huang, Xuanjing  and
      Specia, Lucia  and
      Yih, Scott Wen-tau",
    booktitle = "Proceedings of the 2021 Conference on Empirical Methods in Natural Language Processing",
    month = nov,
    year = "2021",
    address = "Online and Punta Cana, Dominican Republic",
    publisher = "Association for Computational Linguistics",
    url = "https://aclanthology.org/2021.emnlp-main.243/",
    doi = "10.18653/v1/2021.emnlp-main.243",
    pages = "3045--3059"
}

@article{
li2025llavaonevision,
title={{LL}a{VA}-OneVision: Easy Visual Task Transfer},
author={Bo Li and Yuanhan Zhang and Dong Guo and Renrui Zhang and Feng Li and Hao Zhang and Kaichen Zhang and Peiyuan Zhang and Yanwei Li and Ziwei Liu and Chunyuan Li},
journal={Transactions on Machine Learning Research},
issn={2835-8856},
year={2025},
url={https://openreview.net/forum?id=zKv8qULV6n},
note={}
}

@inproceedings{
hu2022lora,
title={Lo{RA}: Low-Rank Adaptation of Large Language Models},
author={Edward J Hu and yelong shen and Phillip Wallis and Zeyuan Allen-Zhu and Yuanzhi Li and Shean Wang and Lu Wang and Weizhu Chen},
booktitle={International Conference on Learning Representations},
year={2022},
url={https://openreview.net/forum?id=nZeVKeeFYf9}
}

@inproceedings{qin-etal-2023-learning,
    title = "Learning to Initialize: Can Meta Learning Improve Cross-task Generalization in Prompt Tuning?",
    author = "Qin, Chengwei  and
      Joty, Shafiq  and
      Li, Qian  and
      Zhao, Ruochen",
    editor = "Rogers, Anna  and
      Boyd-Graber, Jordan  and
      Okazaki, Naoaki",
    booktitle = "Proceedings of the 61st Annual Meeting of the Association for Computational Linguistics (Volume 1: Long Papers)",
    month = jul,
    year = "2023",
    address = "Toronto, Canada",
    publisher = "Association for Computational Linguistics",
    url = "https://aclanthology.org/2023.acl-long.659/",
    doi = "10.18653/v1/2023.acl-long.659",
    pages = "11802--11832"
}

@inproceedings{antol2015vqa,
  title={Vqa: Visual question answering},
  author={Antol, Stanislaw and Agrawal, Aishwarya and Lu, Jiasen and Mitchell, Margaret and Batra, Dhruv and Zitnick, C Lawrence and Parikh, Devi},
  booktitle={Proceedings of the IEEE international conference on computer vision},
  pages={2425--2433},
  year={2015}
}

@inproceedings{
najdenkoska2023meta,
title={Meta Learning to Bridge Vision and Language Models for Multimodal Few-Shot Learning},
author={Ivona Najdenkoska and Xiantong Zhen and Marcel Worring},
booktitle={The Eleventh International Conference on Learning Representations },
year={2023},
url={https://openreview.net/forum?id=3oWo92cQyxL}
}

@INPROCEEDINGS{10376617,
  author={Li, Juncheng and Gao, Minghe and Wei, Longhui and Tang, Siliang and Zhang, Wenqiao and Li, Mengze and Ji, Wei and Tian, Qi and Chua, Tat-Seng and Zhuang, Yueting},
  booktitle={2023 IEEE/CVF International Conference on Computer Vision (ICCV)}, 
  title={Gradient-Regulated Meta-Prompt Learning for Generalizable Vision-Language Models}, 
  year={2023},
  volume={},
  number={},
  pages={2551-2562},
  doi={10.1109/ICCV51070.2023.00241}}

@Article{Krishna2017,
author={Krishna, Ranjay
and Zhu, Yuke
and Groth, Oliver
and Johnson, Justin
and Hata, Kenji
and Kravitz, Joshua
and Chen, Stephanie
and Kalantidis, Yannis
and Li, Li-Jia
and Shamma, David A.
and Bernstein, Michael S.
and Fei-Fei, Li},
title={Visual Genome: Connecting Language and Vision Using Crowdsourced Dense Image Annotations},
journal={International Journal of Computer Vision},
year={2017},
month={May},
day={01},
volume={123},
number={1},
pages={32-73},
issn={1573-1405},
doi={10.1007/s11263-016-0981-7},
url={https://doi.org/10.1007/s11263-016-0981-7}
}

@inproceedings{
gao2025gllava,
title={G-{LL}a{VA}: Solving Geometric Problem with Multi-Modal Large Language Model},
author={Jiahui Gao and Renjie Pi and Jipeng Zhang and Jiacheng Ye and Wanjun Zhong and Yufei Wang and Lanqing HONG and Jianhua Han and Hang Xu and Zhenguo Li and Lingpeng Kong},
booktitle={The Thirteenth International Conference on Learning Representations},
year={2025},
url={https://openreview.net/forum?id=px1674Wp3C}
}

@misc{sharegpt,
  author       = {ShareGPT},
  title        = {ShareGPT},
  howpublished = {\url{https://sharegpt.com/}},
  year         = {2023}
}

@misc{lillog,
    author     = {Lilian Weng},
  title        = {Meta-Learning: Learning to Learn Fast},
  howpublished = {\url{https://lilianweng.github.io/posts/2018-11-30-meta-learning/}},
  year         = {2018}
}

@misc{qwen2025qwen25technicalreport,
      title={Qwen2.5 Technical Report}, 
      author={Qwen and : and An Yang and Baosong Yang and Beichen Zhang and Binyuan Hui and Bo Zheng and Bowen Yu and Chengyuan Li and Dayiheng Liu and Fei Huang and Haoran Wei and Huan Lin and Jian Yang and Jianhong Tu and Jianwei Zhang and Jianxin Yang and Jiaxi Yang and Jingren Zhou and Junyang Lin and Kai Dang and Keming Lu and Keqin Bao and Kexin Yang and Le Yu and Mei Li and Mingfeng Xue and Pei Zhang and Qin Zhu and Rui Men and Runji Lin and Tianhao Li and Tianyi Tang and Tingyu Xia and Xingzhang Ren and Xuancheng Ren and Yang Fan and Yang Su and Yichang Zhang and Yu Wan and Yuqiong Liu and Zeyu Cui and Zhenru Zhang and Zihan Qiu},
      year={2025},
      eprint={2412.15115},
      archivePrefix={arXiv},
      primaryClass={cs.CL},
      url={https://arxiv.org/abs/2412.15115}, 
}

@inproceedings{glorot2010understanding,
  title={Understanding the difficulty of training deep feedforward neural networks},
  author={Glorot, Xavier and Bengio, Yoshua},
  booktitle={Proceedings of the thirteenth international conference on artificial intelligence and statistics},
  pages={249--256},
  year={2010},
  organization={JMLR Workshop and Conference Proceedings}
}

@inproceedings{lu2024mathvista,
title={MathVista: Evaluating Mathematical Reasoning of Foundation Models in Visual Contexts},
author={Pan Lu and Hritik Bansal and Tony Xia and Jiacheng Liu and Chunyuan Li and Hannaneh Hajishirzi and Hao Cheng and Kai-Wei Chang and Michel Galley and Jianfeng Gao},
booktitle={The Twelfth International Conference on Learning Representations},
year={2024},
url={https://openreview.net/forum?id=KUNzEQMWU7}
}

@inproceedings{
antoniou2018how,
title={How to train your {MAML}},
author={Antreas Antoniou and Harrison Edwards and Amos Storkey},
booktitle={International Conference on Learning Representations},
year={2019},
url={https://openreview.net/forum?id=HJGven05Y7},
}

@INPROCEEDINGS{9008296,
  author={Yun, Sangdoo and Han, Dongyoon and Chun, Sanghyuk and Oh, Seong Joon and Yoo, Youngjoon and Choe, Junsuk},
  booktitle={2019 IEEE/CVF International Conference on Computer Vision (ICCV)}, 
  title={CutMix: Regularization Strategy to Train Strong Classifiers With Localizable Features}, 
  year={2019},
  volume={},
  number={},
  pages={6022-6031},
  doi={10.1109/ICCV.2019.00612}}

@inproceedings{
zhang2018mixup,
title={mixup: Beyond Empirical Risk Minimization},
author={Hongyi Zhang and Moustapha Cisse and Yann N. Dauphin and David Lopez-Paz},
booktitle={International Conference on Learning Representations},
year={2018},
url={https://openreview.net/forum?id=r1Ddp1-Rb},
}

@inproceedings{min-etal-2022-metaicl,
    title = "{M}eta{ICL}: Learning to Learn In Context",
    author = "Min, Sewon  and
      Lewis, Mike  and
      Zettlemoyer, Luke  and
      Hajishirzi, Hannaneh",
    editor = "Carpuat, Marine  and
      de Marneffe, Marie-Catherine  and
      Meza Ruiz, Ivan Vladimir",
    booktitle = "Proceedings of the 2022 Conference of the North American Chapter of the Association for Computational Linguistics: Human Language Technologies",
    month = jul,
    year = "2022",
    address = "Seattle, United States",
    publisher = "Association for Computational Linguistics",
    url = "https://aclanthology.org/2022.naacl-main.201/",
    doi = "10.18653/v1/2022.naacl-main.201",
    pages = "2791--2809"
}

@inproceedings{chen-etal-2022-meta,
    title = "Meta-learning via Language Model In-context Tuning",
    author = "Chen, Yanda  and
      Zhong, Ruiqi  and
      Zha, Sheng  and
      Karypis, George  and
      He, He",
    editor = "Muresan, Smaranda  and
      Nakov, Preslav  and
      Villavicencio, Aline",
    booktitle = "Proceedings of the 60th Annual Meeting of the Association for Computational Linguistics (Volume 1: Long Papers)",
    month = may,
    year = "2022",
    address = "Dublin, Ireland",
    publisher = "Association for Computational Linguistics",
    url = "https://aclanthology.org/2022.acl-long.53/",
    doi = "10.18653/v1/2022.acl-long.53",
    pages = "719--730"
}

@misc{choshen2022fusingfinetunedmodelsbetter,
      title={Fusing finetuned models for better pretraining}, 
      author={Leshem Choshen and Elad Venezian and Noam Slonim and Yoav Katz},
      year={2022},
      eprint={2204.03044},
      archivePrefix={arXiv},
      primaryClass={cs.CL},
      url={https://arxiv.org/abs/2204.03044}, 
}

@misc{wortsman2022model,
      title={Model soups: averaging weights of multiple fine-tuned models improves accuracy without increasing inference time}, 
      author={Mitchell Wortsman and Gabriel Ilharco and Samir Yitzhak Gadre and Rebecca Roelofs and Raphael Gontijo-Lopes and Ari S. Morcos and Hongseok Namkoong and Ali Farhadi and Yair Carmon and Simon Kornblith and Ludwig Schmidt},
      year={2022},
      eprint={2203.05482},
      archivePrefix={arXiv},
      primaryClass={cs.LG},
      url={https://arxiv.org/abs/2203.05482}, 
}

@inproceedings{
alayrac2022flamingo,
title={Flamingo: a Visual Language Model for Few-Shot Learning},
author={Jean-Baptiste Alayrac and Jeff Donahue and Pauline Luc and Antoine Miech and Iain Barr and Yana Hasson and Karel Lenc and Arthur Mensch and Katherine Millican and Malcolm Reynolds and Roman Ring and Eliza Rutherford and Serkan Cabi and Tengda Han and Zhitao Gong and Sina Samangooei and Marianne Monteiro and Jacob Menick and Sebastian Borgeaud and Andrew Brock and Aida Nematzadeh and Sahand Sharifzadeh and Mikolaj Binkowski and Ricardo Barreira and Oriol Vinyals and Andrew Zisserman and Karen Simonyan},
booktitle={Advances in Neural Information Processing Systems},
editor={Alice H. Oh and Alekh Agarwal and Danielle Belgrave and Kyunghyun Cho},
year={2022},
url={https://openreview.net/forum?id=EbMuimAbPbs}
}

@inproceedings{10.5555/3157382.3157504,
author = {Vinyals, Oriol and Blundell, Charles and Lillicrap, Timothy and Kavukcuoglu, Koray and Wierstra, Daan},
title = {Matching networks for one shot learning},
year = {2016},
isbn = {9781510838819},
publisher = {Curran Associates Inc.},
address = {Red Hook, NY, USA},
booktitle = {Proceedings of the 30th International Conference on Neural Information Processing Systems},
pages = {3637–3645},
numpages = {9},
location = {Barcelona, Spain},
series = {NIPS'16}
}

@inproceedings{johnson2017clevr,
  title={Clevr: A diagnostic dataset for compositional language and elementary visual reasoning},
  author={Johnson, Justin and Hariharan, Bharath and Van Der Maaten, Laurens and Fei-Fei, Li and Lawrence Zitnick, C and Girshick, Ross},
  booktitle={Proceedings of the IEEE conference on computer vision and pattern recognition},
  pages={2901--2910},
  year={2017}
}

@inproceedings{singh2021textocr,
  title={Textocr: Towards large-scale end-to-end reasoning for arbitrary-shaped scene text},
  author={Singh, Amanpreet and Pang, Guan and Toh, Mandy and Huang, Jing and Galuba, Wojciech and Hassner, Tal},
  booktitle={Proceedings of the IEEE/CVF conference on computer vision and pattern recognition},
  pages={8802--8812},
  year={2021}
}

@article{vaswani2017attention,
  title={Attention is all you need},
  author={Vaswani, Ashish and Shazeer, Noam and Parmar, Niki and Uszkoreit, Jakob and Jones, Llion and Gomez, Aidan N and Kaiser, {\L}ukasz and Polosukhin, Illia},
  journal={Advances in neural information processing systems},
  volume={30},
  year={2017}
}

@inproceedings{
ravi2017optimization,
title={Optimization as a Model for Few-Shot Learning},
author={Sachin Ravi and Hugo Larochelle},
booktitle={International Conference on Learning Representations},
year={2017},
url={https://openreview.net/forum?id=rJY0-Kcll}
}

@misc{zhai2023sigmoidlosslanguageimage,
      title={Sigmoid Loss for Language Image Pre-Training}, 
      author={Xiaohua Zhai and Basil Mustafa and Alexander Kolesnikov and Lucas Beyer},
      year={2023},
      eprint={2303.15343},
      archivePrefix={arXiv},
      primaryClass={cs.CV},
      url={https://arxiv.org/abs/2303.15343}, 
}

@inproceedings{
shu2022testtime,
title={Test-Time Prompt Tuning for Zero-Shot Generalization in Vision-Language Models},
author={Manli Shu and Weili Nie and De-An Huang and Zhiding Yu and Tom Goldstein and Anima Anandkumar and Chaowei Xiao},
booktitle={Advances in Neural Information Processing Systems},
editor={Alice H. Oh and Alekh Agarwal and Danielle Belgrave and Kyunghyun Cho},
year={2022},
url={https://openreview.net/forum?id=e8PVEkSa4Fq}
}

@inproceedings{
hardt2024testtime,
title={Test-Time Training on Nearest Neighbors for Large Language Models},
author={Moritz Hardt and Yu Sun},
booktitle={The Twelfth International Conference on Learning Representations},
year={2024},
url={https://openreview.net/forum?id=CNL2bku4ra}
}

@article{karmanov2024efficient,
          title={Efficient Test-Time Adaptation of Vision-Language Models},
          author={Karmanov, Adilbek and Guan, Dayan and Lu, Shijian and El Saddik, Abdulmotaleb and Xing, Eric},
          journal={The IEEE/CVF Conference on Computer Vision and Pattern Recognition},
          year={2024}
  }

@inproceedings{
hu2025testtime,
title={Test-Time Learning for Large Language Models},
author={Jinwu Hu and Zitian Zhang and Guohao Chen and Xutao Wen and Chao Shuai and Wei Luo and Bin Xiao and Yuanqing Li and Mingkui Tan},
booktitle={Forty-second International Conference on Machine Learning},
year={2025},
url={https://openreview.net/forum?id=iCYbIaGKSR}
}

@inproceedings{hou-etal-2022-metaprompting,
    title = "{M}eta{P}rompting: Learning to Learn Better Prompts",
    author = "Hou, Yutai  and
      Dong, Hongyuan  and
      Wang, Xinghao  and
      Li, Bohan  and
      Che, Wanxiang",
    editor = "Calzolari, Nicoletta  and
      Huang, Chu-Ren  and
      Kim, Hansaem  and
      Pustejovsky, James  and
      Wanner, Leo  and
      Choi, Key-Sun  and
      Ryu, Pum-Mo  and
      Chen, Hsin-Hsi  and
      Donatelli, Lucia  and
      Ji, Heng  and
      Kurohashi, Sadao  and
      Paggio, Patrizia  and
      Xue, Nianwen  and
      Kim, Seokhwan  and
      Hahm, Younggyun  and
      He, Zhong  and
      Lee, Tony Kyungil  and
      Santus, Enrico  and
      Bond, Francis  and
      Na, Seung-Hoon",
    booktitle = "Proceedings of the 29th International Conference on Computational Linguistics",
    month = oct,
    year = "2022",
    address = "Gyeongju, Republic of Korea",
    publisher = "International Committee on Computational Linguistics",
    url = "https://aclanthology.org/2022.coling-1.287/",
    pages = "3251--3262"
}

@inproceedings{wang-etal-2022-towards-unified,
    title = "Towards Unified Prompt Tuning for Few-shot Text Classification",
    author = "Wang, Jianing  and
      Wang, Chengyu  and
      Luo, Fuli  and
      Tan, Chuanqi  and
      Qiu, Minghui  and
      Yang, Fei  and
      Shi, Qiuhui  and
      Huang, Songfang  and
      Gao, Ming",
    editor = "Goldberg, Yoav  and
      Kozareva, Zornitsa  and
      Zhang, Yue",
    booktitle = "Findings of the Association for Computational Linguistics: EMNLP 2022",
    month = dec,
    year = "2022",
    address = "Abu Dhabi, United Arab Emirates",
    publisher = "Association for Computational Linguistics",
    url = "https://aclanthology.org/2022.findings-emnlp.37/",
    doi = "10.18653/v1/2022.findings-emnlp.37",
    pages = "524--536"
}

@InProceedings{Khattak_2023_ICCV,
    author    = {Khattak, Muhammad Uzair and Wasim, Syed Talal and Naseer, Muzammal and Khan, Salman and Yang, Ming-Hsuan and Khan, Fahad Shahbaz},
    title     = {Self-regulating Prompts: Foundational Model Adaptation without Forgetting},
    booktitle = {Proceedings of the IEEE/CVF International Conference on Computer Vision (ICCV)},
    month     = {October},
    year      = {2023},
    pages     = {15190-15200}
}

@misc{yang2025qwen3technicalreport,
      title={Qwen3 Technical Report}, 
      author={An Yang and Anfeng Li and Baosong Yang and Beichen Zhang and Binyuan Hui and Bo Zheng and Bowen Yu and Chang Gao and Chengen Huang and Chenxu Lv and Chujie Zheng and Dayiheng Liu and Fan Zhou and Fei Huang and Feng Hu and Hao Ge and Haoran Wei and Huan Lin and Jialong Tang and Jian Yang and Jianhong Tu and Jianwei Zhang and Jianxin Yang and Jiaxi Yang and Jing Zhou and Jingren Zhou and Junyang Lin and Kai Dang and Keqin Bao and Kexin Yang and Le Yu and Lianghao Deng and Mei Li and Mingfeng Xue and Mingze Li and Pei Zhang and Peng Wang and Qin Zhu and Rui Men and Ruize Gao and Shixuan Liu and Shuang Luo and Tianhao Li and Tianyi Tang and Wenbiao Yin and Xingzhang Ren and Xinyu Wang and Xinyu Zhang and Xuancheng Ren and Yang Fan and Yang Su and Yichang Zhang and Yinger Zhang and Yu Wan and Yuqiong Liu and Zekun Wang and Zeyu Cui and Zhenru Zhang and Zhipeng Zhou and Zihan Qiu},
      year={2025},
      eprint={2505.09388},
      archivePrefix={arXiv},
      primaryClass={cs.CL},
      url={https://arxiv.org/abs/2505.09388}, 
}
\bibliographystyle{iclr2026_conference}

\newpage
\appendix
\section{Appendix}

\begin{enumerate}
    \item Section \ref{sec_app:implement}: Implementation Details
    \begin{enumerate}
        \item Section \ref{sec_app:fin_data} Finetuning Data Mixture
        \item {Section \ref{sec_app:meta_task} Details on Meta-Task Creation}
        \item Section \ref{sec_app:model} Model Configurations
        \item Section \ref{sec_app:train_det} Training Details
        \item Section \ref{sec_app:psuedo_aglo} Psuedo Algorithm of MAPD
        \end{enumerate}
    \item Section \ref{sec_app:eval_det} Evaluation
        \begin{enumerate}
            \item {Section \ref{sec_app:prem_analysis_ins} Detailed Task Instructions for LMM Evaluation.}
            \item Section \ref{sec_app:eval_data} Evaluation Datasets from VL-ICL Bench
            \item Section \ref{sec_app:test_time_adapt} Test-Time Adaptation Details
            \item Section \ref{sec_app:det_result} Detailed Results on VL-ICL Bench
            \item Section \ref{sec_app:llava_compare} Performance of Publicly Available LMMs on VL-ICL Bench
            \item Section \ref{sec_app:quality} Qualitative Results on VL-ICL Bench
            \item Section \ref{sec_app:img_perturb} Robustness Against Image Perturbations
            \item Section \ref{sec_app:sel_criteria} How to Select Few-Shot Examples for Better Performance?
            \item Section \ref{sec_app:op_abls} Details on Ablation Study for Operator Induction
            \item Section \ref{sec_app:att_entropy} Attention Entropy Analysis
            \item Section \ref{sec_app:more_shots} Scaling to More Shots
        \end{enumerate}
    \item Section \ref{sec_app:tta_compute} Test-time Compute Analysis for ICL vs FT
\end{enumerate}

\newpage

\subsection{Implementation Details}
\label{sec_app:implement}

\subsubsection{Finetuning Data Mixture}
\label{sec_app:fin_data}

For model finetuning, we create our multi-task data mixture for single image per example using the visual instruction tuning data of LLaVA v1.5 \citep{liu2023visual} which contains a mixture of 12 different datasets\footnote{We use this dataset only for academic research purposes as mentioned by the original authors and follow the Open AI Usage Policy for GPT-4 generated datasets. Additionally, we conform to the license (CC-BY-4.0) for Cauldron datasets.} ranging from long conversations to academic multiple-choice questions. Since we are only training image-based prompts, we remove the language-only ShareGPT-40K dataset \citep{sharegpt}. Additionally, we include 3 different math reasoning/QA datasets from the LLaVA OneVision data mixture \citep{li2025llavaonevision} which are known to improve LMM performance on difficult reasoning and logical QA tasks \citep{lu2024mathvista}. We further get rid of the extra answer formatting instructions to test the true few-shot transfer learning ability of our approach without the need of  external task induction. Table \ref{tab:datamix} shows the list of all the datasets along with their size and question types.

\begin{table}[t]
  \caption{Finetuning Data Mixture Statistics}
  \label{tab:datamix}
  \centering
  \begin{tabular}{@{}l|c|l@{}}
    \toprule
   Dataset & No. of examples & Question Types \\
    \midrule
    \multirow{3}{*}{LLaVA-Instruct} & \multirow{3}{*}{157,712} & Conversations (57,669) \\ 
    {} & {} & Detailed Image Description (23,240) \\ 
    {} & {} & Complex Reasoning (76,803) \\
    \midrule
    GQA & 72,140 & Visual Reasoning \\
    \midrule
    \multirow{2}{*}{OCR-VQA} & \multirow{2}{*}{80,000} & Image Question Answering \\
    {} & {} & with Reading Comprehension \\
    \midrule
    \multirow{2}{*}{TextVQA} & \multirow{2}{*}{21,953} & Image Question Answering \\
    {} & {} & with Reading Comprehension \\
    \midrule
    \multirow{2}{*}{Visual Genome} & \multirow{2}{*}{86,417} & Image Question Answering  \\
    {} & {} & and Bounding Box Prediction \\
    \midrule
    \multirow{2}{*}{MAVIS-Math-Metagen} & \multirow{2}{*}{87,348} & Visual Math \\
    {} & {} & Question Answering \\
    \midrule
    TabMWP-Cauldron & 22,717 & Tabular Math Reasoning \\
    \midrule
    \multirow{2}{*}{RefCOCO} & \multirow{2}{*}{48,447} & Image Question Answering  \\
    {} & {} & and Bounding Box Prediction \\
    \midrule
    \multirow{2}{*}{OKVQA} & \multirow{2}{*}{8,998} & Knowledge Grounded \\
    {} & {} & Image Question Answering \\
    \midrule
    VQAv2 & 82,783 & Image Question Answering \\
    \midrule
    \multirow{2}{*}{A-OKVQA} & \multirow{2}{*}{66,160} & Multiple-Choice Question \\
    {} & {} & Answering \\
    \midrule
    \multirow{2}{*}{Geo-170k (QA)} & \multirow{2}{*}{67,823} & Math Question Answering \\
    {} & {} & and Reasoning \\
    \midrule
    Total & 802,498 & {} \\
   \bottomrule
   \end{tabular}
\end{table}

\subsubsection{Details on Meta-Task Creation}
\label{sec_app:meta_task}
As mentioned in Section \ref{sec:meta_tasks}, meta-tasks are small subsets of examples randomly sampled from a single VQA dataset ($D^i$) within the training data mixture ($p(\mathcal{D})$). Each meta-task consists of support and query sets, each containing a fixed number of VQA examples (image, question, answer triplets). The support set provides few-shot demonstrations to the model, either as in-context examples or for gradient-based adaptation, depending on the prompt distillation method. The query set is used to optimize the LMM (specifically, the attention-mapper parameters ($\theta_p$) in our case) during fine-tuning, and to evaluate performance during inference. This meta-task construction protocol remains consistent across both the fine-tuning stage and test-time fine-tuning, following the framework established by \citep{zong2025vlicl}.

During test-time adaptation, we use the publicly available VL-ICL benchmark code\footnote{VL-ICL: \url{https://github.com/ys-zong/VL-ICL}.} to construct meta-tasks of fixed sizes. VQA examples are randomly sampled from the predefined training and test splits of each dataset. Table \ref{tab:mt_tests} specifies the number of meta-tasks per test set, which remains constant throughout our evaluation. All results reported in the paper represent average accuracy computed over the query examples of these meta-tasks, ensuring fair comparison across all prompt distillation methods and shot configurations.

\begin{table}[t]
  \caption{Meta-Task composition during test-time adaptation}
  \label{tab:mt_tests}
  \centering
  \begin{tabular}{@{}l|c|c|c|c@{}}
    \toprule
    {} & Open\_MI & Operator Induction & CLEVR & TextOCR \\
    \midrule
    No. of Meta-tasks & 5000 & 4000 & 6000 & 5000 \\
    \midrule
    Support examples & \multirow{2}{*}{[1,2,4,5]} & \multirow{2}{*}{[1,2,4,8]} & \multirow{2}{*}{[1,2,4,8]} & \multirow{2}{*}{[1,2,4,8]} \\ 
    per meta-task & & & & \\
    \midrule
    Query examples & \multirow{2}{*}{1} & \multirow{2}{*}{1} & \multirow{2}{*}{1} & \multirow{2}{*}{1} \\ 
    per meta-task & & & & \\
   \bottomrule
   \end{tabular}
\end{table}

During the attention-mapper fine-tuning stage, in order to keep a balanced ratio of train-validation splits across multiple datasets in Section \ref{sec_app:fin_data} used in this stage, we divide each dataset into $98\%$ for training and $2\%$ for validation and then combine them separately to create the final train and validation splits. We then construct meta-tasks by randomly sampling VQA examples. We treat the support-query composition as a tunable hyperparameter alongside those listed in Table, performing a grid search to identify the configuration that minimizes validation loss for each prompt distillation method. Table \ref{tab:mt_finetune} details the optimal support-query compositions, number of meta-tasks, and total number of training and validation examples used for each method. 

\begin{table}[t]
  \caption{Meta-Task composition during the finetuning stage}
  \label{tab:mt_finetune}
  \centering
  \begin{tabular}{@{}l|c|c|c@{}}
    \toprule
    {} & MAPD & Multi-task\textsuperscript{PD} & In-Context\textsuperscript{PD} \\
    \midrule
    No. of Meta-tasks & 39,650 / & 79,300 / & 72,100 / \\
    (train/val) & {8000} & {8000} & {8000} \\
    \midrule
    Support examples & \multirow{2}{*}{10 /} & \multirow{2}{*}{5 /} & \multirow{2}{*}{10 /} \\
    per meta-task & \multirow{2}{*}{[1,2,4,5,8]} & \multirow{2}{*}{[1,2,4,5,8]} & \multirow{2}{*}{[1,2,4,5,8]} \\
    (train/val) & {} & {} & {} \\
    \midrule
    Query examples & \multirow{3}{*}{10 / 1} & \multirow{3}{*}{5 / 1} & \multirow{3}{*}{1 / 1} \\ 
    per meta-task & & & \\
    (train/val) & {} & {} & {} \\
    \midrule
    Total no. of examples & 793,000 / & 793,000 / & 793,000 / \\
    (train/val) & 16,000 & 16,000 & 16,000 \\
   \bottomrule
   \end{tabular}
\end{table}

Additionally, for In-Context\textsuperscript{PD}, we follow the in-context tuning algorithm of \citep{chen-etal-2022-meta}, which uses only 1 query example per meta-task during training and yields optimal performance for this prompt distillation method. Note that the validation is done across a different number of support examples for robustness and the total number of training and validation examples remains constant across all methods to ensure fair comparison, regardless of meta-task composition.

\subsubsection{Model Configurations}
\label{sec_app:model}

\textbf{Models} We use the publicly available implementation of LLaVA v1.5\footnote{LLaVA v1.5: \url{https://github.com/haotian-liu/LLaVA/tree/main/llava}} and first-order MAML\footnote{MAML: \url{https://github.com/AntreasAntoniou/HowToTrainYourMAMLPytorch}} to implement our baselines. Additionally, we use the pretrained model weights from Huggingface for Qwen2.5-7B-Instruct LLM\footnote{Qwen2.5-7B-Instruct: \url{https://huggingface.co/Qwen/Qwen2.5-7B-Instruct}} and the CLIP ViT-L/14-336px visual encoder\footnote{CLIP-ViT-L/14-336px: \url{https://huggingface.co/openai/clip-vit-large-patch14-336}}. The output embedding dimension size of CLIP is 1,024 and the input word embedding size of the Qwen LLM is 3,584. We set the training context length as 4096 for all baselines except for in-context baseline where it is 8,192 as it requires training with longer sequences. The attention-mapper is a single multi-head attention block with 8 heads. The token length of the soft prompt $P$ as described in Section \ref{sec:arch} for the attention mapper is set to $m=256$. The total number of trainable parameters for our model is approximately 24M making our approach significantly parameter-efficient for finetuning.

\subsubsection{Training Details}
\label{sec_app:train_det}
\textbf{Pretraining Stage} During the pretraining stage, we only train the attention-mapper and soft prompts for 4 epochs and use a learning rate of 2e-3 with a batch size of 64 per GPU. We perform a train-validation split on the LCS-558K dataset \citep{liu2023visual} by keeping $98\%$ of the examples for training and $2\%$ for validation and take the checkpoint with the lowest validation loss. We use this checkpoint as our base for further task-specific finetuning.

\textbf{Finetuning Stage} For finetuning, we perform a grid search across a fixed set of values as we are constrained by our GPU resources (4 H200 GPUs). For each prompt distillation method, we select the configuration that achieves the lowest validation loss following standard train-val-test procedures. Table \ref{tab:mt_hyper} (for meta-task methods) and Table \ref{tab:nmt_hyper} (for non-meta-task methods) provide details of all hyperparameters for which we performed grid search.  We also provide additional training details below, separately for each method, along with their corresponding best sets of hyperparameters after grid search. All approaches were finetuned for 1 epoch to ensure a complete pass over the entire finetuning data mixture. 

\begin{table}[t]
  \caption{Grid search values for meta-task methods}
  \label{tab:mt_hyper}
  \centering
  \begin{tabular}{@{}l|c|c|c|c@{}}
    \toprule
    {} & {No. of support/query} & \multirow{2}{*}{Learning Rate} & {Inner-loop} & Batch Size\\
    {} & {per meta task} & {} & learning rate & (\# of meta-tasks) \\
    \midrule
    Search Values & [1, 5, 10, 15] & [1e-3, 5e-4, 2e-5] & [1e-1, 5e-2, 5e-1] & [1, 5] \\
    \midrule
   \bottomrule
   \end{tabular}
\end{table}

\begin{table}[t]
  \caption{Grid search values for non meta-task methods}
  \label{tab:nmt_hyper}
  \centering
  \begin{tabular}{@{}l|c|c@{}}
    \toprule
    {} & {Learning Rate} & Batch Size \\
    \midrule
    Search Values & [1e-3, 5e-4, 2e-5] & [16, 32, 64, 80] \\
    \midrule
   \bottomrule
   \end{tabular}
\end{table}

\begin{enumerate}
    \item \textbf{MAPD:}  We use 5 inner-loop steps and initialize the inner-loop learning rate $\alpha$=1e-1. The outer-loop learning rate is set as 1e-3 with a per GPU batch size of 1 meta-task with a gradient accumulation of 2 steps. Each meta-task for training contains 10 support and 10 query examples. Training time $\sim$ 10 hours. 
    \item \textbf{Multi-Task\textsuperscript{PD}:}  Similar to MAPD, we use a learning rate of 1e-3 with a per GPU batch size of 1 meta-task with a gradient accumulation of 4 steps. Each meta-task for training contains 5 support and 5 query examples. Training time $\sim$ 4.5 hours
    \item \textbf{In-Context\textsuperscript{PD}:}  We use a learning rate of 1e-3 with a gradient accumulation of 4 steps and 5 meta tasks per GPU. Each meta task for training contains 10 support examples and 1 query example. The support examples were concatenated with the strategy that ensured all image tokens of a meta-task are present in the sequence and we truncate the text tokens if the sequence exceeded the context length of 8192. Further, the few-shot question and answers were concatenated by inserting "Question:" and "Answer:" strings in between them, inspired from \citep{alayrac2022flamingo}. Training time $\sim$ 4.5 hours
    \item \textbf{ModelAvg\textsuperscript{PD}:}  We first finetune individual models on each dataset in the finetuning data mixture (Section \ref{sec_app:fin_data}) with a learning rate of 5e-4. For all the datasets, we choose a per GPU batch size of 8 with gradient accumulation of 2 steps. Average time per dataset $\sim$ 3 hours
    \item \textbf{NoMeta-task\textsuperscript{PD}:}   Here,  we finetune on the complete data mixture in one training run by sampling batches randomly and again use a per GPU batch size of 8 with a gradient accumulation of 2 steps. We also use a learning rate of 5e-4. Training time $\sim$ 4 hours.
    \item \textbf{LoRA}: We only apply LoRA to the attention matrices ($Q,K,V$) of each layer. For training, we use a learning rate of 5e-4 and a per GPU batch size of 8 with gradient accumulation of 2 steps. Further, we performed hyperparameter search for choosing LoRA parameters - rank (r) and scaling factor ($\alpha$) for the three settings shown in Table \ref{tab:lora_results}. Training time $\sim$ 4 hours.
        \begin{enumerate}
            \item \textit{All LLM layers} ($r=128, \alpha = 256$)
            \item \textit{[0-15] LLM layers} ($r=16, \alpha=64$)
            \item   \textit{[0-15] LLM layers + ATT}: ($r=16, \alpha=64$)
            
        \end{enumerate}
\end{enumerate}
\textbf{Computational Requirements}\hspace{.7em}  We find that the GPU requirement for training the attention-mapper mostly depends on the size of the underlying LLM used. For the 7B model training, we use 4 H200 GPUs with a VRAM of 143GB per GPU and for 3B models only 2 H200 GPUs were needed.  For both the stages, the hyperparameters were tuned using their corresponding validation sets and we choose the checkpoints at the end of first epoch to report our results.

\subsubsection {Pseudo algorithm for MAPD}
\label{sec_app:psuedo_aglo}
We highlight our full MAPD algorithm based on FoMAML in detail with inner and outer loop that is used to train the attention-mapper parameters $\theta_p$ in Algorithm \ref{algo:fomaml}.

\begin{algorithm}[h]
\caption{Meta-Adaptive Prompt Distillation (MAPD)}
\label{algo:fomaml}
\KwIn{Meta-Task distribution $p(\mathcal{T}^{\text{meta}})$, inner‐loop learning rate $\alpha$, meta learning rate $\beta$}
\KwOut{Meta-parameters $\theta_p= \{\theta, P\}$}
Initialize $\theta_p$ with Xavier Uniform Initialization\;
\While{not converged}{
  Sample batch of meta-tasks $\{T_j\}_{j=1}^N \sim p(\mathcal{T}^{\text{meta}})$\;
  \ForEach{task $T_j = \{D^{\text{supp}}_j, D^{\text{query}}_j\}$ in batch}{
    Evaluate $L^{\text{supp}}_{\theta_{p,j}} = \dfrac{-1}{|D^{\text{supp}}_j|} \displaystyle \sum_{i=1}^{|D^{\text{supp}}_j|} \log(p_{\theta_{p,j}}(X_a^i | X_v^i, X_q^i))$\;
    Adapt parameters with $K$ gradient steps:\\  
     \For{$k=1,\dots,K$}{
    \[
      \theta^{k}_{p,j} \leftarrow  \theta^{k-1}_{p,j} - \alpha \nabla_{\theta^{k-1}_{p,j}} L^{\text{supp}}_{\theta^{k-1}_{p,j}}
    \]
    }
  }
  Evaluate $L_{\theta^{K}_{p,j}}^{\text{query}} = \dfrac{-1}{|D^{\text{query}}_j|} \displaystyle \sum_{i=1}^{|D^{\text{query}}_j|}\log(p_{\theta^{K}_{p,j}}(X_a^i | X_v^i, X_q^i))$\;
  First-Order Meta‐Update:  
  \[
    \theta_p 
    \leftarrow 
    \theta_p 
    - 
    \beta\,\sum_{j=1}^N \nabla_{\theta^K_{p,j}}
      L_{\theta^{K}_{p,j}}^{{\text{query}}}
  \]
}
\end{algorithm}

\subsection{Evaluation Details}

\label{sec_app:eval_det}

\subsubsection{Detailed Task Instructions for LMM Evaluation.}
\label{sec_app:prem_analysis_ins}

Figure~\ref{fig:prem_prompts} shows the detailed task instructions used to evaluate LLaVA-OneVision-7B in the Image-to-Text (I2T) ICL setting.

\begin{figure}[t]
  \centering
  \begin{PlainBox}
    \begin{itemize}[leftmargin=.2cm]
        \item \textbf{Operator Induction} - \textit{"The image contains two digit numbers and a ? representing the mathematical operator. Induce the mathematical operator (addition, multiplication, minus) according to the results of the in-context examples and calculate the result."}
        \item \textbf{CLEVR Count Induction} - \textit{"The image contains objects of different shapes, colors, sizes and materials. The question describes the attribute and its value. You need to find all objects within the image that satisfy the condition. You should induce what operation to use according to the results of the in-context examples and then calculate the result."}
    \end{itemize}
  \end{PlainBox}
\caption{Detailed task instruction for LLaVA-OneVision-7B LMM evaluation on VL-ICL tasks.}
  \label{fig:prem_prompts}
\end{figure}

\subsubsection{Evaluation Datasets from VL-ICL Bench}
\label{sec_app:eval_data}

\begin{table}[t]
  \caption{Evaluation Dataset Statistics}
  \label{tab:evaldata}
  \centering
  \begin{tabular}{@{}l|c|c|c|c@{}}
    \toprule
   \multirow{2}{*}{Dataset} & \multirow{2}{*}{Task Category} & Train Set & Test Set & \multirow{2}{*}{Size (GB)} \\
   {} & {} & (Support) & (Query) & {} \\ 
    \midrule
    Fast Open-MiniImageNet & \multirow{2}{*}{Fast-Concept Binding} & \multirow{2}{*}{5,000} & \multirow{2}{*}{200} & \multirow{2}{*}{0.18} \\
    (OPEN\_MI) & {} & {} &{} & {} \\
    \midrule
    \multirow{2}{*}{CLEVR Count Induction} & Fine-Grained Perception, & \multirow{2}{*}{800} & \multirow{2}{*}{200} & \multirow{2}{*}{0.18} \\
    {} & Task Induction & {} & {} \\ 
    \midrule
    \multirow{2}{*}{Operator Induction} & Perception, Task Induction & \multirow{2}{*}{80} & \multirow{2}{*}{60} & \multirow{2}{*}{0.01} \\
    {} & Mathematical Reasoning & {} & {} \\
    \midrule
    TextOCR & Perception, Task Induction & 800 & 200 & 0.01 \\
    \bottomrule
    \end{tabular}
\end{table}

The VL-ICL Bench \cite{zong2025vlicl} includes a diverse variety of tasks to test different capabilities of models like Fast-Concept binding, Mathematical Induction, and Fine-grained perception. Given the nature of our model architecture and training (Section \ref{sec:prompt_tuning}), we only focus on the single-image Image-to-text (I2T) tasks. Table \ref{tab:evaldata} shows the  dataset statistics. We also give brief descriptions of these tasks below along with some examples for better understanding.

\begin{enumerate}
    \item \textbf{Fast Open-Ended MiniImageNet (OPEN\_MI)} - This is a variant of the MiniImageNet few-shot object recognition task \citep{10.5555/3157382.3157504}, which was repurposed for few-shot prompting \citep{NEURIPS2021_01b7575c}. It is essentially an open-ended image classfication problem, but contains nonsense categorical names like \textit{dax} or \textit{blicket} making the test performance not influenced by the prior knowledge of an LMM but only dependent on the support examples. This design ensures to test the few-shot abilities of LMMs and how quickly they can learn about new concepts.
    For the results shown in Table \ref{tab:baseline}, we use the 2-way version of this task involving classification between two nonsense categories. An example of a 2-way 1-shot task is shown in Figure \ref{img:open_mi}.
    \begin{figure}[t]
    \includegraphics[width=\textwidth,keepaspectratio]{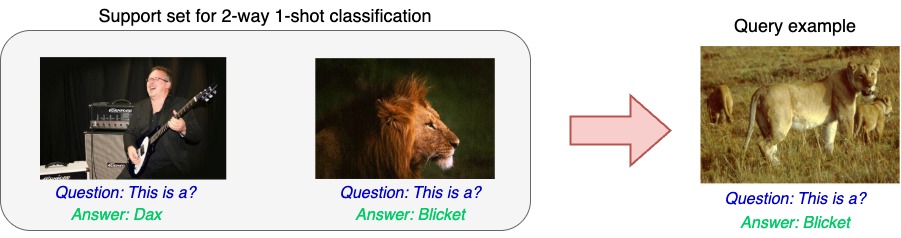}
    \caption{2-way Fast Open-Ended MiniImageNet}
    \label{img:open_mi}
    \end{figure}
    \begin{figure}[t]
    \includegraphics[width=\textwidth,keepaspectratio]{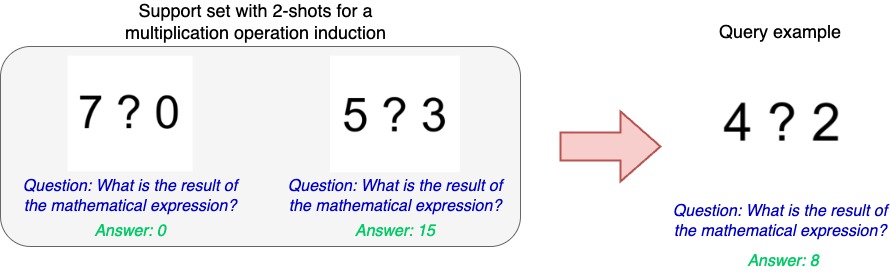}
    \caption{Operator Induction}
    \label{img:op_ind}
    \end{figure}
    \item \textbf{Operator Induction} - Initially proposed by \citep{zong2025vlicl}, this dataset tests various capabilties of LMMs like Task Induction, Perception and Mathematical Reasoning. The support examples involve two operands with a missing mathematical operation and an answer. When testing, the task is to identify the hidden operation from the support example and use it to calculate the result over the operands in the query. An example of a 2-shot task is shown in Figure \ref{img:op_ind}.
    \item \textbf{CLEVR Count Induction} - This dataset contains images from the widely used CLEVR dataset \citep{johnson2017clevr} where each image contains a set of objects that have certain characteristics based on attributes like shape, size, color and material. The task is to learn to count the objects of the given attribute in the support example and transfer that knowledge to count the objects of any attribute in the query example. An example of a 2-shot task is shown in Figure \ref{img:clevr}.
    \begin{figure}[t]
    \includegraphics[width=\textwidth,keepaspectratio]{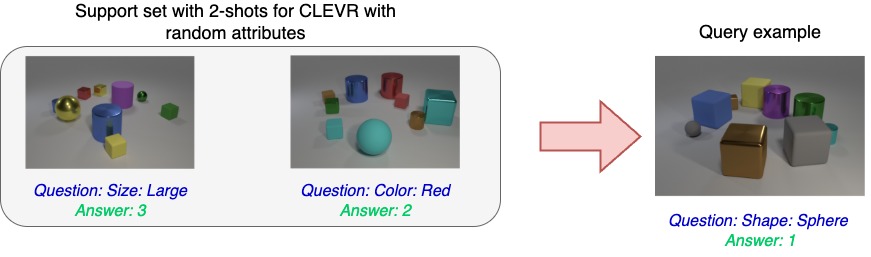}
    \caption{CLEVR Count Induction}
    \label{img:clevr}
    \end{figure}
    \item \textbf{TextOCR} - This dataset has been repurposed by \citep{zong2025vlicl} from the TextOCR dataset \citep{singh2021textocr} to create a task where the LMM should learn to output the text within a red bounding box from the support examples. Even though this task could be solved in a zero-shot setting as we see in the 0-shot case with a detailed prompt, we still only focus on inducing task knowledge from the few-shot examples. An example of a 2-shot task is shown in Figure \ref{img:textocr}.
    \begin{figure}[h]
    \includegraphics[width=\textwidth,keepaspectratio]{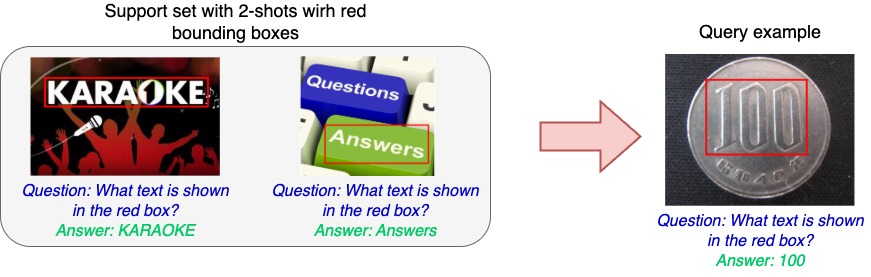}
    \caption{TextOCR}
    \label{img:textocr}
    \end{figure}
\end{enumerate}

\newpage

\subsubsection{Test-Time Adaptation Details}
\label{sec_app:test_time_adapt}
We follow a similar test-time adaptation procedure to \citep{qin-etal-2023-learning} to find the best hyperparameter settings for each prompt distillation method, ensuring fair comparison. We first sample 10\% of the examples from the training split of each test task and combine them to form a validation set. After meta-task creation over the VL-ICL datasets \citep{zong2025vlicl} using the remaining training and test splits, we perform a maximum of $K=30$ inner-loop steps over each support set and select the $K$th-step model that achieves the lowest validation loss. This model is then used to compute results on the query set. To further validate whether $K=30$ is a sufficient number of fine-tuning steps, we plot average test accuracy curves (up to 40 gradient steps) across VL-ICL datasets, methods, and shot settings in Figure~\ref{fig:test_compare}. Accuracies converge within 30 steps, confirming our choice of $K$. We also provide examples of how predictions evolve during test-time adaptation in Figures~\ref{img:open_mi_img_grad_a}, \ref{img:clevr_img_grad_a}, \ref{img:textocr_img_grad_a}, and~\ref{img:opind_img_grad_a}. For reproducibility, Table~\ref{tab:lr} reports the best learning rates for each method, selected via grid search over $[0.1, 1.0]$ on the validation set with a batch size of one meta-task.
\begin{figure}[t]
  \centering
  \begin{subfigure}[b]{0.48\textwidth}
    \centering
    \includegraphics[width=\textwidth,keepaspectratio]{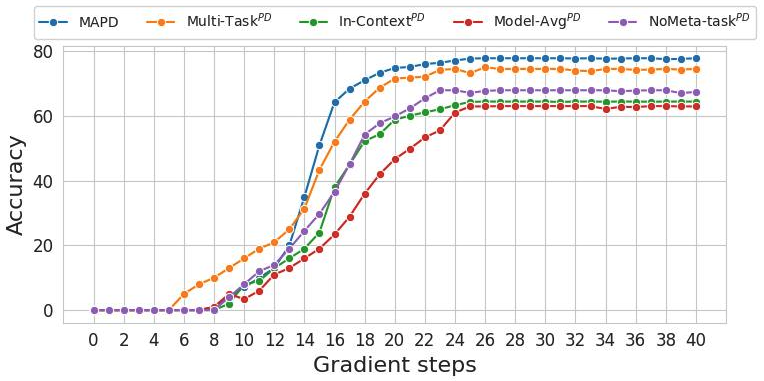}
    \caption{OPEN\_MI test performance}
    \label{img:open_mi_a}
  \end{subfigure}
  \hfill                           
  \begin{subfigure}[b]{0.48\textwidth}
    \centering
    \includegraphics[width=\textwidth,keepaspectratio]{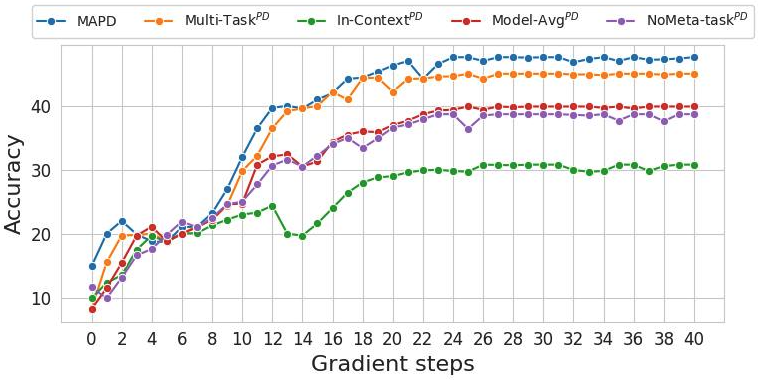}
    \caption{Operator Induction test performance}
    \label{img:op_ind_a}
  \end{subfigure}
  \begin{subfigure}[b]{0.48\textwidth}
    \centering
    \includegraphics[width=\textwidth,keepaspectratio]{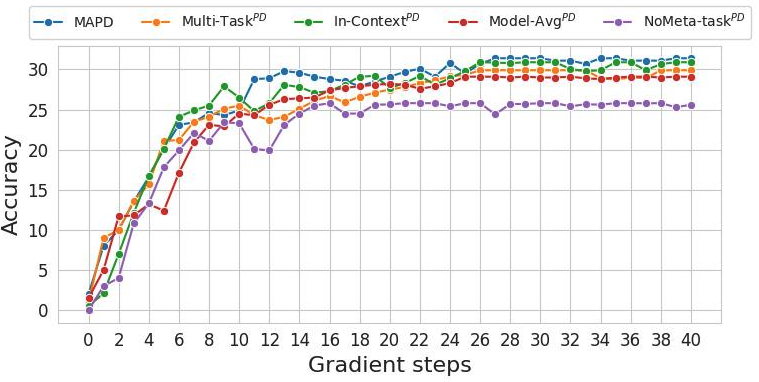}
    \caption{CLEVR test performance}
    \label{img:clevr_a}
  \end{subfigure}
  \begin{subfigure}[b]{0.48\textwidth}
    \centering
    \includegraphics[width=\textwidth,keepaspectratio]{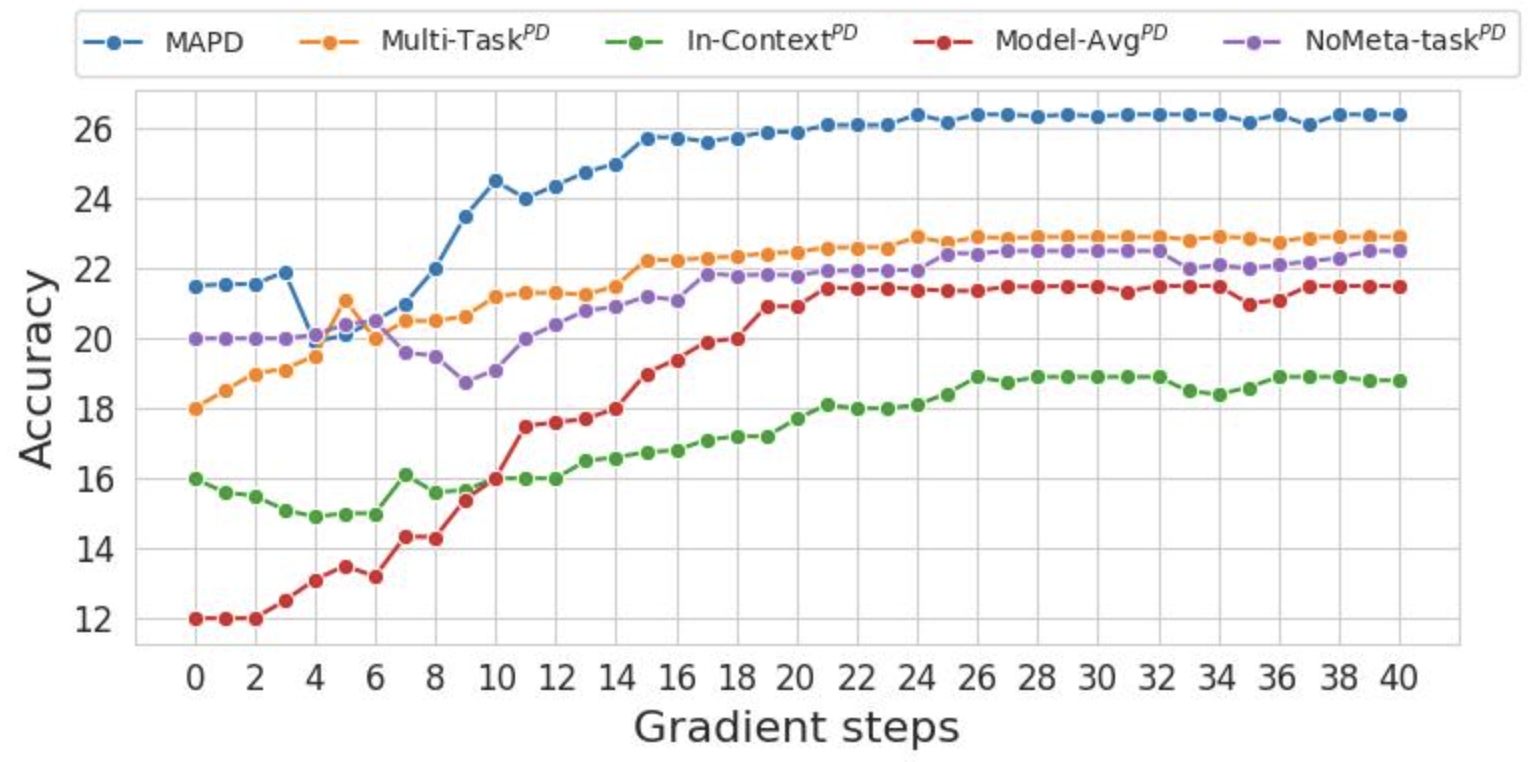}
    \caption{TEXTOCR test performance}
    \label{img:textocr_a}
  \end{subfigure}
  \caption{Average test performances of MAPD with finetuning on different datasets}
  \label{fig:test_compare}
\end{figure}

\begin{table}[t]
  \caption{Learning rates for finetuning-based (FT) test-time adaptation for results shown in Table \ref{tab:main_results}, Table \ref{tab:lora_results}, Table \ref{tab:baseline} and Table \ref{tab:lora_baseline}.}
  \label{tab:lr}
  \centering
  \begin{tabular}{@{}l|c@{}}
    \toprule
   \multicolumn{1}{c}{Training Methods} & Learning Rate (LR) \\ 
   \midrule
    MAPD & 1.0 \\
    Multi-Task\textsuperscript{PD} & 0.8 \\
    In-Context\textsuperscript{PD} & 0.8 \\
    ModelAvg\textsuperscript{PD} & 0.6 \\
    NoMeta-task\textsuperscript{PD} & 1.0 \\
    LoRA & 0.2 \\
    \bottomrule
    \end{tabular}
\end{table}
\clearpage
\newpage
\subsubsection{Detailed Results}
\label{sec_app:det_result}
\begin{table}[ht]
  \caption{Comparison of different prompt distillation approaches on single-image tasks from  VL-ICL Bench \citep{zong2025vlicl}. We report accuracy for  different numbers of shots (--S). "Avg" is only calculated for $\geq 1$ shot(s). FT = Finetuning with $\leq$ 30 gradient steps, ICL = In-Context Learning, TTA= Test-Time Adaptation. More details are mentioned in Appendix \ref{sec_app:test_time_adapt}. We do not compare on 0-shot results. The model used for this evaluation is LLaVA-ATT-Qwen2.5 7B which is described in Section \ref{sec:arch}. Meta-Tasks used (\cmark) or not (\xmark) during training. We also provide results for higher number of shots in Appendix \ref{sec_app:more_shots} and qualitative results in Appendix \ref{sec_app:quality} and \ref{sec_app:op_abls}.}
  \label{tab:baseline}
  \centering
  \resizebox{\textwidth}{!}{
  \begin{tabular}{@{}l|c|crlll|c||rrrrr|c@{}}
    \toprule
   \multicolumn{1}{c}{\multirow{2}{*}{Methods}} & \multicolumn{1}{c}{\multirow{2}{*}{\shortstack{Meta\\Task}}} & \multicolumn{6}{c}{Open-MI (2-way)} & \multicolumn{6}{c}{Operator Induction} \\
    \cmidrule(r){3-8} \cmidrule(r){9-14}
   \multicolumn{1}{c}{} & \multicolumn{1}{c}{} & \textcolor{gray!90}{0-S} & 1--S & 2--S & 4--S & \multicolumn{1}{c}{5-S} & \multicolumn{1}{c}{Avg} & \textcolor{gray!90}{0-S} & 1--S & 2--S & 4--S & \multicolumn{1}{c}{8--S} & \multicolumn{1}{c@{}}{Avg}\\
    \midrule
    \rowcolor{gray!20}
    \textbf{TTA with ICL} & & & & & & & 
    & & & & & & \\
    NoMeta-task$^{\text{PD}}$ & \xmark & \textcolor{gray!90}{0.0}  & {35.0} & {47.0} & {48.0} & {45.0} & {43.8} & \textcolor{gray!90}{11.7} & {13.3} & {13.3} & {10.0} & {11.7} & {12.1} \\
    Model-Avg\textsuperscript{PD} & \xmark & \textcolor{gray!90}{0.0} & 20.0 & 22.0 & 30.0 & 34.5 & {26.6} & \textcolor{gray!90}{8.3} & 11.7 & 6.7 & 8.3 & 10.0 & {9.2} \\
    In-Context\textsuperscript{PD} & \cmark & \textcolor{gray!90}{0.0} & 30.0  & 56.0 & 55.0  & 63.5 & {51.1} & \textcolor{gray!90}{10.0} & 20.0 & 18.5 & 18.0 & 26.0 & {20.6} \\
    Multi-Task\textsuperscript{PD} & \cmark & \textcolor{gray!90}{0.0} & 43.0 & 50.0 & 51.0 & 50.5 & {48.6} & \textcolor{gray!90}{8.3} & 13.3 & 11.7 & 3.3 & 11.7 & {10.0} \\
    MAPD & \cmark & \textcolor{gray!90}{0.0} & 42.5 & 53.0 & 57.0 & 60.5 & {53.3} & \textcolor{gray!90}{15.0} & 13.3 & 13.3 & 1.7 & 10.0 & {9.6} \\
    \midrule
    \rowcolor{gray!20}
    \textbf{TTA with FT $\leq$30} & & & & & & & 
    & & & & & & \\ 
    NoMeta-task$^{\text{PD}}$ & \xmark & \textcolor{gray!90}{0.0}  & 21.5 & 67.5 & 89.0 & 94.0 & {68.0} & \textcolor{gray!90}{11.7} & 26.7 & 23.3 & 46.7 & 58.3 & {38.8}\\
    Model-Avg\textsuperscript{PD} & \xmark & \textcolor{gray!90}{0.0}  & {28.5} & {53.5} & {83.0} & {87.5} & {63.1} & \textcolor{gray!90}{8.3} & {31.5} & {28.0} & {45.0}& {55.5} & {40.0}\\
    In-Context\textsuperscript{PD} & \cmark & \textcolor{gray!90}{0.0} & 35.5 & 54.5 & 79.5 & 88.5 & {64.5} & \textcolor{gray!90}{10.0} & 21.7 & 18.3 & 41.7 & 41.7 & {30.9} \\
    Multi-Task\textsuperscript{PD} & \cmark & \textcolor{gray!90}{0.0} & 37.0 & 73.5 & 93.5 & 94.5 & {74.6} & \textcolor{gray!90}{8.3} & 31.0 & 28.3 & \textbf{61.0} & 60.0 & {45.1} \\
    \textbf{MAPD} & \cmark & \textcolor{gray!90}{0.0}  & \textbf{43.5} & \textbf{78.0} & \textbf{94.5} & \textbf{95.5} & \textbf{{77.9}}  & \textcolor{gray!90}{15.0} & \textbf{32.0} & \textbf{38.3} & 58.3 & \textbf{62.0} & \textbf{{47.7}} \\
    \midrule
    \addlinespace[1ex]
    \multicolumn{1}{c}{\multirow{2}{*}{Methods}} & \multicolumn{1}{c}{\multirow{2}{*}{\shortstack{Meta\\Task}}} &
   \multicolumn{6}{c}{CLEVR Count Induction} & \multicolumn{6}{c}{TextOCR} \\
    \cmidrule(r){3-8} \cmidrule(r){9-14}
   \multicolumn{1}{c}{} & \multicolumn{1}{c}{} & \textcolor{gray!90}{0--S} & 1--S & 2--S & 4--S & \multicolumn{1}{c}{8-S} & \multicolumn{1}{c}{Avg} & \textcolor{gray!90}{0-S} & 1--S & 2--S & 4--S & \multicolumn{1}{c}{8--S} & \multicolumn{1}{c}{Avg}\\
    \midrule
    \rowcolor{gray!20}
    \textbf{TTA with ICL} & & & & & & & 
    & & & & & & \\
    NoMeta-task$^{\text{PD}}$ & \xmark & \textcolor{gray!90}{0.0}  & {8.0} & {10.5} & {23.0} & {30.5} & {18.0} & \textcolor{gray!90}{20.0} & 4.5 & 9.5 & 8.5 & 4.5 & {6.8} \\
    Model-Avg\textsuperscript{PD} & \xmark & \textcolor{gray!90}{1.5} & 17.0 & 8.5 & 4.0 & 1.0 & {7.6} & \textcolor{gray!90}{12.0} & 3.0 & 2.5 & 3.0 & 1.0 & {2.8} \\
    In-Context\textsuperscript{PD} & \cmark & \textcolor{gray!90}{0.0} & 13.5  & 23.0 & 28.5 & 31.5 & {24.1} & \textcolor{gray!90}{16.0} & 22.5 & 21.0 & 23.5 & 28.0 & {23.8}\\
    Multi-Task\textsuperscript{PD} & \cmark & \textcolor{gray!90}{1.0} & 5.0 & 9.0 & 16.5 & 19.5 & {12.5} & \textcolor{gray!90}{18.0} & 4.0 & 4.5 & 8.5 & 10.5 & {6.9} \\
    MAPD & \cmark & \textcolor{gray!90}{2.0} & 11.0  & 7.0 & 15.5 & 15.5 & {12.3} & \textcolor{gray!90}{21.5} & 5.5 & 7.0 & 8.0 & 8.5 & {7.3} \\
    \midrule
    \rowcolor{gray!20}
    \textbf{TTA with FT $\leq$30} & & & & & & & 
    & & & & & & \\ 
    NoMeta-task$^{\text{PD}}$ & \xmark & \textcolor{gray!90}{0.0}  & 18.5 & 21.5 & 26.0 & 37.0 & {25.8} & \textcolor{gray!90}{20.0} & 20.5 & 23.0 & 24.0 & 22.5 & {22.5} \\
    Model-Avg\textsuperscript{PD} & \xmark & \textcolor{gray!90}{1.5}  & \textbf{26.5} & {25.0} & {29.5} & {35.5} & {29.1} & \textcolor{gray!90}{12.0} & {17.5} & {20.0} & {23.0}& {25.5} & {21.5} \\
    In-Context\textsuperscript{PD} & \cmark & \textcolor{gray!90}{0.5} & 24.5 & \textbf{30} & \textbf{34.5} & 34.5 & {30.9} & \textcolor{gray!90}{16.0} & 16.0 & 18.0 & 19.5 & 22.0 & {18.9} \\
    Multi-Task\textsuperscript{PD} & \cmark & \textcolor{gray!90}{0.0} & 25.0 & 25.5 & 31.0 & 38.0 & {29.9} & \textcolor{gray!90}{18.0} & 21.0 & 20.5 & 24.5 & 25.5 & {22.9} \\
    \textbf{MAPD} & \cmark & \textcolor{gray!90}{0.0}  & \textbf{26.5} & 27.5 & 31.0 & \textbf{40.5} & \textbf{{31.4}} & \textcolor{gray!90}{21.5} & \textbf{23.5} & \textbf{26.5} & \textbf{27.0} & \textbf{28.5} & \textbf{{26.4}} \\
    \bottomrule
  \end{tabular}
}
\end{table}

\begin{table}[ht]
  \caption{Comparison of the LoRA baselines on VL-ICL Bench \citep{zong2025vlicl}. "Avg" is only calculated for $\geq 1$ shot(s) (-S). TTA= Test-Time Adaptation. FT=Finetuning with $\leq$ 30 gradient steps. ATT=Attention-Mapper. The model used for this evaluation is LLaVA-ATT-Qwen2.5 7B.}
  \label{tab:lora_baseline}
  \centering
  \resizebox{\textwidth}{!}{
  \begin{tabular}{@{}l|crlll|c||rrrrr|c@{}}
    \toprule
   \multicolumn{1}{l}{\multirow{2}{*}{LoRA}} & \multicolumn{6}{c}{Open-MI (2-way)} & \multicolumn{6}{c}{Operator Induction} \\
    \cmidrule(r){2-7} \cmidrule(r){8-13}
   \multicolumn{1}{c}{} & \textcolor{gray!90}{0-S} & 1--S & 2--S & 4--S & \multicolumn{1}{c}{5-S} & \multicolumn{1}{c}{Avg} & \textcolor{gray!90}{0-S} & 1--S & 2--S & 4--S & \multicolumn{1}{c}{8--S} & \multicolumn{1}{c@{}}{Avg}\\
    \midrule
    \rowcolor{gray!20}
    \textbf{TTA with FT $\leq$ 30} & & & & & & 
    & & & & & & \\
    All LLM layers & \textcolor{gray!90}{0.0}  & {24.5} & {45.7} & {68.3} & {81.9} & {55.1} & \textcolor{gray!90}{8.1} & {11.7} & {10.0} & {13.3} & {18.2} & {13.3} \\
    \text{[0-15]} LLM layers & \textcolor{gray!90}{0.0} & 30.9 & \textbf{65.3} & 81.1 & \textbf{91.9} & {67.3} & \textcolor{gray!90}{8.3} & 18.3 & 26.3 & 23.1 & 34.3 & {25.5} \\
    \text{[0-15]} LLM layers + ATT & \textcolor{gray!90}{0.0} & \textbf{37.3} & 64.1 & \textbf{83.5} & 91.5 & \textbf{69.1} & \textcolor{gray!90}{10.0} & \textbf{21.5} & \textbf{28.3} & \textbf{35.5} & \textbf{36.7} & \textbf{30.5} \\
    \midrule
    \multicolumn{1}{l}{\multirow{2}{*}{LoRA}} & \multicolumn{6}{c}{CLEVR Count Induction} & \multicolumn{6}{c}{TextOCR} \\
    \cmidrule(r){2-7} \cmidrule(r){8-13}
   \multicolumn{1}{c}{} & \textcolor{gray!90}{0-S} & 1--S & 2--S & 4--S & \multicolumn{1}{c}{5-S} & \multicolumn{1}{c}{Avg} & \textcolor{gray!90}{0-S} & 1--S & 2--S & 4--S & \multicolumn{1}{c}{8--S} & \multicolumn{1}{c@{}}{Avg}\\
    \midrule
    \rowcolor{gray!20}
    \textbf{TTA with FT $\leq$ 30} & & & & & & 
    & & & & & & \\
    All LLM layers & \textcolor{gray!90}{0.0}  & {9.3} & {11.7} & {15.5} & {23.9} & {15.1} & \textcolor{gray!90}{15.0} & {6.7} & {9.1} & {13.3} & {12.5} & {10.4} \\
    \text{[0-15]} LLM layers & \textcolor{gray!90}{0.0} & 21.5 & \textbf{28.3} & \textbf{32.5} & \textbf{37.7} & \textbf{30.0} & \textcolor{gray!90}{18.3} & 20.3 & \textbf{24.5} & 25.5 & 24.9 & {23.8} \\
    \text{[0-15]} LLM layers + ATT & \textcolor{gray!90}{0.0} & \textbf{26.0} & 23.1 & 30.0 & 35.7 & {28.7} & \textcolor{gray!90}{18.3} & \textbf{20.6} & 23.4 & \textbf{26.5} & \textbf{27.5} & \textbf{24.5} \\
    \bottomrule
    
  \end{tabular}
}
\end{table}

\newpage
\subsubsection{Performance of publicly available LMMs on VL-ICL Bench}
\label{sec_app:llava_compare}

\begin{table}[ht]
  \caption{Performance of different LMMs on single-image tasks from  VL-ICL Bench. We report the "Avg" accuracy for different numbers of shots - $\{1,2,4,5,8\}$ {with 95\% binomial confidence intervals}. FT = Finetuning with $\leq$ 30 gradient steps, ICL = In-Context Learning, TTA= Test-Time Adaptation, VL-Data=Vision-Language Data, LAQ-7B=LLaVA-ATT-Qwen2.5-7B, CLIP=CLIP-ViT-L/14-336px, MLP=2-layer MLP,  ATT=Attention-Mapper. \textbf{Bold} shows best performance and \underline {Underline} is MAPD's performance with LAQ-7B LMM.}
  \label{tab:llava_models}
  \centering
  \begin{adjustbox}{max width=\columnwidth,center}
  \begin{tabular}{l|c|c|c|c|c|c|c}
    \toprule
   \multirow{2}{*}{Methods} & \multirow{2}{*}{VL-Data} & Params & \multirow{2}{*}{TTA} & \multirow{2}{*}{Open-MI} & \multirow{2}{*}{OP\_IND} & \multirow{2}{*}{CLEVR} & \multirow{2}{*}{TextOCR} \\
   {} & {} & trained & {} & {} & {} & {} & {} \\ 
    \midrule
    LLaVA v1.5-7B & 1.2M & 7B & ICL & $12.4 \pm 0.4$ & $5.4 \pm 0.5$ & $10.9 \pm 0.1$ & $4.4 \pm 0.3$ \\
    LLaVA v1.5-7B & 1.2M & 7B & FT$\leq$30 & $38.4 \pm 0.7$ & $11.4 \pm 0.6$ & $16.9 \pm 0.2$ & $15.6 \pm 0.6$ \\
    LLaVA-Next-7B & 1.3M & 7.06B & ICL & $34.4 \pm 0.7$ & $5.4 \pm 0.5$ & $21.1 \pm 0.2$ & $0.4 \pm 0.0$ \\
    LLaVA-Next-7B & 1.3M & 7.06B & FT$\leq$30 & $55.1 \pm 0.9$ & $13.4 \pm 0.6$ & $28.6 \pm 0.2$ & $7.8 \pm 0.4$ \\
    LLaVA-OneVision-7B & 10.4M & 8B & ICL & $42.1 \pm 0.9$ & $41.7 \pm 0.5$ & $34.9 \pm 0.2$ & $42.3 \pm 0.5$ \\
    LLaVA-OneVision-7B & 10.4M & 8B & FT$\leq$30 & $83.4 \pm 0.7$ & $46.1 \pm 0.5$ & $\textbf{38.9} \pm \textbf{0.2}$ & $45.5 \pm 0.5$ \\
    LLaVA-OneVision-72B & 10.4M & 73.2B & ICL & $75.1 \pm 0.6$ & $69.1 \pm 0.9$ & $37.2 \pm 0.2$ & $\textbf{52.2} \pm \textbf{1.1}$ \\
    Qwen2-VL-7B-Instruct & -NA- & 8B & ICL & $73.5 \pm 0.6$ & $69.6 \pm 0.9$ & $27.9 \pm 0.2$ & $50.5 \pm 0.9$ \\
    Qwen2.5-VL-7B-Instruct & -NA- & 8B & ICL & $44.0 \pm 0.9$ & $84.2 \pm 1.2$ & $22.0 \pm 0.2$ & $36.9 \pm 0.7$ \\
    Qwen2.5-VL-7B-Instruct & -NA- & 8B & FT$\leq$30 & $\textbf{85.6} \pm \textbf{0.7}$ & $\textbf{89.4} \pm \textbf{1.2}$ & $29.1 \pm 0.2$ & $41.1 \pm 0.5$ \\
    LAQ-7B + In-Context\textsuperscript{PD} & 1.3M & 24M & ICL & $51.1 \pm 0.9$ & $20.6 \pm 0.8$ & $24.1 \pm 0.2$ & $23.8 \pm 0.8$ \\
    LAQ-7B + In-Context\textsuperscript{PD} & 1.3M & 24M & FT$\leq$30 & $64.5 \pm 0.8$ & $30.9 \pm 0.5$ & $30.9 \pm 0.2$ & $18.9 \pm 0.7$ \\
    LAQ-7B + MAPD & 1.3M & 24M & ICL & $53.3 \pm 0.9$ & $9.6 \pm 0.5$ & $12.3 \pm 0.1$ & $7.3 \pm 0.4$ \\
    \textbf{LAQ-7B + MAPD} & 1.3M & 24M & FT$\leq$30 &  \underline{$77.9 \pm 0.7$} & \underline{$47.7 \pm 0.5$} & \underline{$31.4 \pm 0.2$} & \underline{$26.4 \pm 0.8$} \\
    \bottomrule
  \end{tabular}
  \end{adjustbox}
\end{table}

We report the performance of publicly available LMMs alongside our best-performing architecture (LLaVA-ATT-Qwen2.5-7B) on the single-image tasks from VL-ICL Bench in Table~\ref{tab:llava_models}. \textbf{We provide this as a reference and note that direct comparison across LMMs is not straightforward, given their fundamental differences in architecture, scale, and training data.}
\begin{enumerate}
\item Test-time fine-tuning of the MLP connector consistently improves over ICL for all public LMMs, supporting our hypothesis that these models are overwhelmed by image embeddings during ICL. Fine-tuning enables the connector to distil task-specific information into image embeddings before prompting the LLM, thereby improving few-shot performance.
\item Our model with MAPD-based meta-learning and fine-tuning adaptation performs comparably to other publicly available LMMs and, notably, surpasses LLaVA-OneVision-72B ICL on the Fast Open-Ended MiniImageNet (Open-MI) task, as well as ICL with its 7B counterpart (trained on substantially more data) and the stronger Qwen-VL models on other tasks.
\item Unlike other LMMs, LLaVA-ATT-Qwen2.5-7B (LAQ-7B) does not fine-tune the LLM during training and uses significantly less vision-language data (1.3M examples) and fewer trainable parameters (24M), compared to LLaVA-OneVision which trains the full model on 10.4M examples. This highlights the data and parameter efficiency of MAPD, which achieves state-of-the-art performance on Open-MI by fine-tuning only the attention-mapper for up to 30 gradient steps on the few-shot examples.
\item For LMMs such as LLaVA-OneVision, fine-tuning the attention-mapper requires substantial compute ($\geq$12 H200 GPUs) due to their large-scale training mixture (10.4M vision-language examples) and high-dimensional vision encoder embeddings, exceeding our available resources. Similarly, fine-tuning data for the Qwen-VL models is not publicly available. Given these constraints, we are unable to conduct attention-mapper fine-tuning experiments on these architectures.
\end{enumerate}

\newpage
\subsubsection{Qualitative Results}
\label{sec_app:quality}
\begin{figure}[ht]
\centering
 \makebox[\textwidth][c]{%
\includegraphics[width=\textwidth,keepaspectratio]{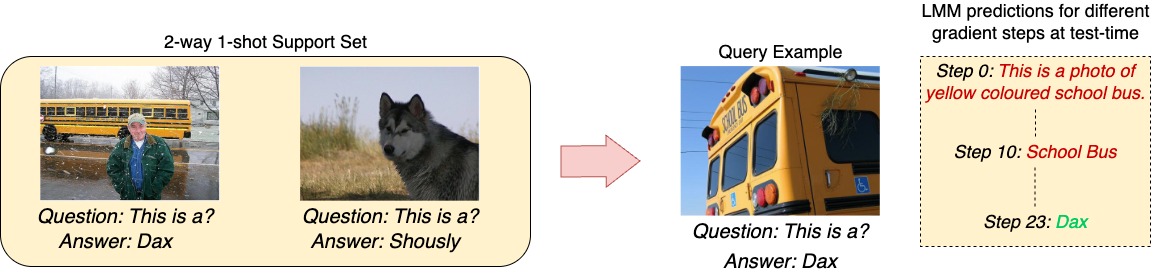}
}
\caption{OPEN\_MI predictions at test-time}
\label{img:open_mi_img_grad_a}
\end{figure}

\begin{figure}[ht]
\centering
 \makebox[\textwidth][c]{%
\includegraphics[width=\textwidth,keepaspectratio]{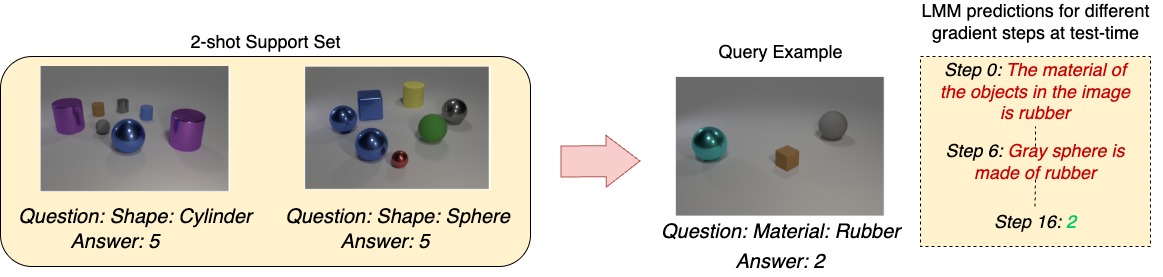}
}
\caption{CLEVR predictions at test-time}
\label{img:clevr_img_grad_a}
\end{figure}

\begin{figure}[ht]
\centering
 \makebox[\textwidth][c]{%
\includegraphics[width=\textwidth,keepaspectratio]{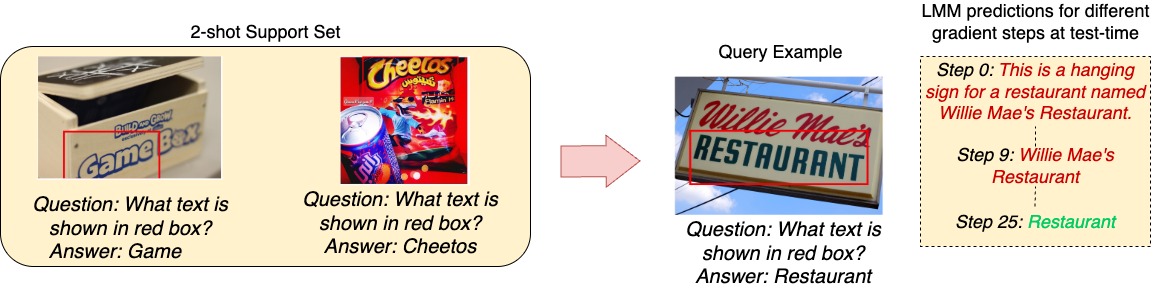}
}
\caption{TEXTOCR predictions at test-time}
\label{img:textocr_img_grad_a}
\end{figure}

\begin{figure}[!htbp]
\centering
 \makebox[\textwidth][c]{%
\includegraphics[width=\textwidth,keepaspectratio]{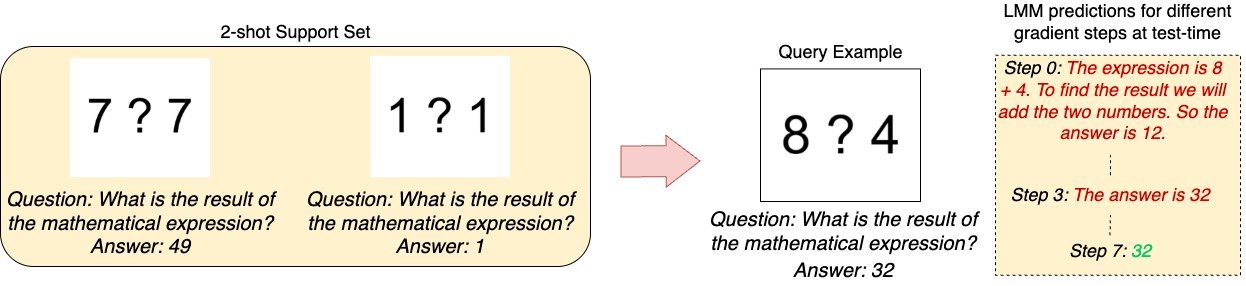}
}
\caption{Operator Induction predictions at test-time}
\label{img:opind_img_grad_a}
\end{figure}

\clearpage
\newpage

\subsubsection{Robustness against image perturbations}
\label{sec_app:img_perturb}
\begin{table}[t]
  \caption{Robustness of prompt distillation methods against image perturbations on the  Fast Open-Ended MiniImageNet dataset (2-way classification) for LLaVA-ATT-Qwen2.5 7B LMM. We  report  accuracy scores as defined in VL-ICL Bench \citep{zong2025vlicl} across 2, and 5 shots. Test-Time Adaptation = Finetuning with $\leq$ 30 gradient steps.}
  \label{tab:image_aug}
  \centering
  \resizebox{\textwidth}{!}{
  \begin{tabular}{@{}l|cc|cc|cc|cc|cc@{}}
    \toprule
   \multicolumn{1}{c}{\multirow{2}{*}{}} & \multicolumn{2}{c}{NoMeta-task\textsuperscript{PD}} & \multicolumn{2}{c}{Model-Avg\textsuperscript{PD}} & \multicolumn{2}{c}{In-Context\textsuperscript{PD}} & \multicolumn{2}{c}{ Multi-task\textsuperscript{PD}} & \multicolumn{2}{c}{MAPD} \\
   \cmidrule(r){2-3} \cmidrule(r){4-5} \cmidrule(r){6-7} \cmidrule(r){8-9} \cmidrule(r){10-11}
   \multicolumn{1}{c}{} & 2--S & \multicolumn{1}{c}{5--S} & 2--S & \multicolumn{1}{c}{5--S} & 2--S & \multicolumn{1}{c}{5-S} & 2--S & \multicolumn{1}{c}{5--S} & 2--S & 5--S \\
    \midrule
    Original & 67.5 & 94.0 & 53.5 & 87.5 & 54.5 & 88.5 & 73.5 & 94.5 & 78.0 & 95.5 \\
    \midrule
    Cropping & 65.0 & 94.0 & 51.5 & 87.5 & 51.5 & 83.0 & 72.0 & 91.5 & 76.5 & 95.0 \\
    Rotation & 67.0  & 91.0 & 50.5 & 81.5 & 50.5 & 83.5 & 72.5 & 93.5 & 78.0 & 95.5 \\
    Gaussian Blur & 67.5 & 92.5 & 51.5 & 84.5 & 49.5 & 78.0 & 71.5 & 92.5 & 77.5 & 96.0 \\
    Color Jitter & 66.5 & 92.5 & 50.5 & 89.0 & 49.5 & 81.5 & 71.5 & 94.0 & 77.0 & 94.0 \\
    CutMix & 58.5 & 86.0 & 45.5 & 70.5 & 49.0 & 75.0 & 72.0 & 92.0  & 75.5 & 92.5 \\
    MixUp & 58.0 & 84.0 & 46.0 & 70.5 & 48.0 & 75.5 & 69.0 & 89.0 & 76.5 & 91.0 \\
    \midrule
    Mean Drop in Accuracy  & $-$3.8 & $-$4.0 & $-$4.3 & $-$6.9 & $-$4.8 & $-$9.1 & $-$2.1 & $-$2.4 & $-$1.2 & $-$1.4 \\ 
    \midrule
    Net Mean Drop across Shots & \multicolumn{2}{c}{$-$3.9} & \multicolumn{2}{c}{$-$5.6} & \multicolumn{2}{c}{$-$7.0} & \multicolumn{2}{c}{$-$2.3} & \multicolumn{2}{c}{\textbf{$-$1.3}} \\
    \bottomrule
  \end{tabular}
}
\end{table}

We assess if our prompt distillation methods are robust enough to handle  perturbations applied to the images in the support set as shown in Table \ref{tab:image_aug}. We see that our method, MAPD, is most robust even in the presence of noise in the support examples as compared to other distillation methods that suffer a huge drop in performance. Advanced techniques like CutMix \citep{9008296} and MixUp \citep{zhang2018mixup} change the original image distribution substantially, affecting all methods  to a greater degree but MAPD is still close to its original performance for both 2 and 5 shots. This robustness likely stems from MAPD’s meta-learned initialization, which learns the underlying task structure from meta-tasks without over-fitting to any other spurious visual patterns and this allows it to adapt quickly to newer tasks without being influenced by noisy visual artifacts in the examples.

\subsubsection{How to select few-shot examples for better performance?}
\label{sec_app:sel_criteria}
\begin{figure}[t]
\begin{subfigure}[t]{0.48\textwidth}
    \centering
 \begin{tikzpicture}[scale=.6]
\begin{axis}[
    ybar,
    bar width=12pt,
    width=11cm,
    height=7cm,
    enlarge x limits=0.1,
    ymin=0,
    ymax=40,
    ylabel={Mean Accuracy (\%)},
    ylabel style={font=\normalsize,yshift=-.2cm},
    xtick=data,
    xticklabel style={align=center, font=\small},
    yticklabel style={font=\small},
    xticklabels={
        {NoMetaTask\textsuperscript{PD}},
        {Model-Avg\textsuperscript{PD}},
        {In-Context\textsuperscript{PD}},
        {Multi-Task\textsuperscript{PD}},
        {MAPD}
    },
    legend style={
        font=\small,
        at={(0.5,1.15)},
        anchor=north,
        legend columns=3
    },
    grid=major,
    grid style={dashed,gray!20},
    nodes near coords,
    every node near coord/.append style={font=\fontsize{6}{5}\selectfont,
      yshift=0pt,text=black},,
]

\addplot+[fill=Bittersweet!60, draw=black] plot coordinates {(0,25.8) (1,29.1) (2,30.9) (3, 29.9) (4, 31.4)};
\addplot+[fill=Yellow!60, draw=black] plot coordinates {(0,34.0) (1,32.9) (2,34.1) (3, 34.9) (4, 35.4)};
\addplot+[fill=RubineRed!60, draw=black] plot coordinates {(0,33.2) (1,34.2) (2,34.6) (3, 35.8) (4, 35.6)};

\legend{Original (Random), Same Attribute, Same Pair}
\end{axis}
\end{tikzpicture}
\end{subfigure}
\begin{subfigure}[t]{0.48\textwidth}
    \centering
 \begin{tikzpicture}[scale=.6]
\begin{axis}[
    ybar,
    bar width=12pt,
    width=11cm,
    height=7cm,
    enlarge x limits=0.1,
    ymin=0,
    ymax=40,
    ylabel={Mean Accuracy (\%)},
    ylabel style={font=\normalsize,yshift=-.2cm},
    xtick=data,
    xticklabel style={align=center, font=\small},
    yticklabel style={font=\small},
    xticklabels={
        {NoMetaTask\textsuperscript{PD}},
        {Model-Avg\textsuperscript{PD}},
        {In-Context\textsuperscript{PD}},
        {Multi-Task\textsuperscript{PD}},
        {MAPD}
    },
    legend style={
        font=\small,
        at={(0.5,1.15)},
        anchor=north,
        legend columns=3
    },
    grid=major,
    grid style={dashed,gray!20},
    nodes near coords,
    every node near coord/.append style={font=\fontsize{6}{5}\selectfont,
      yshift=0pt,text=black},
]

\addplot+[Bittersweet!60, draw=black] plot coordinates {(0,18.0) (1,7.6) (2,24.1) (3, 12.5) (4, 12.3)};
\addplot+[fill=Yellow!60, draw=black] plot coordinates {(0,26.5) (1,11.8) (2,28.8) (3, 23.8) (4, 21.9)};
\addplot+[fill=RubineRed!60, draw=black] plot coordinates {(0,28.4) (1,13.4) (2,30.5) (3, 24.3) (4, 21)};

\legend{Original (Random), Same Attribute, Same Pair}
\end{axis}
\end{tikzpicture}
\end{subfigure}
\caption{(a) Performance comparison of different prompt distillation approaches on the CLEVR Count Induction (details in Appendix \ref{sec_app:eval_data}). Few-shot examples for \emph{Same Attribute} and \emph{Same Pair} are selected based on their \emph{attribute-value} similarity with the query (test) example. Mean Accuracy is computed for 1,2,4 and 8 shots. \textbf{Left}: Finetuning (FT) based Test-time Adaptation. (b) \textbf{Right}: In-Context Learning (ICL) based Test-time Adaptation.} 
\label{fig:abl_clevr}
\end{figure}

We further assess how performance varies for different prompt distillation approaches based on the selection of few-shot examples on the CLEVR Count Induction task (details in Appendix \ref{sec_app:eval_data}) as an example. We propose two selection methods based on similarity of attributes and their corresponding values for every query (test) example. If the query has attribute and value as \emph{shape: sphere}, we select the few-shot examples based on - a) Same Attribute - \emph{shape}, (b) Same Pair - \emph{shape: sphere} and compare both of them with the original setup as proposed in the VL-ICL benchmark \citep{zong2025vlicl} which retrieves the few-shot examples randomly. In Figure \ref{fig:abl_clevr}(a), we first see that for finetuning-based (FT) adaptation, the performance of all the baselines increases by 4.8\% for Same Attribute and 5.3\% for Same Pair on an average. MAPD performs best in the Same Attribute setting (Mean Acc = 35.4\%) and Multi-Task\textsuperscript{PD} performing best in the Same Pair setting (Mean Acc = 35.8\%). In Figure \ref{fig:abl_clevr}(b), we see that for In-Context Learning (ICL) adaptation, the similarity-based few-shot selection methods have a greater impact in performance and improve the mean accuracy of all the baselines by 7.7\% for Same Attribute and 8.6\% for Same Pair on an average. In-Context\textsuperscript{PD} performs the best in both Same Attribute and Same Pair settings with mean accuracies of 28.8\% and 30.5\% respectively for ICL adaptation. We also notice that the Same Pair setup is generally the best few-shot selection method giving best performance for all the approaches. This indicates that choosing few-shot examples that are similar to query example induces better task understanding during test-time adaptation. We also see that the selection of few-shot examples shows less variance with FT adaptation compared to ICL adaptation, thereby showing higher robustness of FT adaptation.

\subsubsection {Details on Ablation Study for Operator Induction}
\label{sec_app:op_abls}
We break down the ablation study on operator induction tasks (Section \ref{sec:abls}; Figure \ref{fig:other_analysis}(b)) into 3 components: 1) Task Induction, 2) Perception, and 3) Mathematical Reasoning. We test these components separately with the help of suitable prompts for our LMM to answer questions in specific formats. Figure \ref{fig:op_prompts} shows our prompts used for different components.

\begin{figure}[h]
  \centering
  \begin{PlainBox}

    \begin{itemize}[leftmargin=.2cm]
        \item \textbf{Task Induction} - \textit{"What mathematical operation should be used in this example? Strictly answer in one word."}
        \item \textbf{Perception} - \textit{"What are the numbers in this example? Do not calculate the answer after applying mathematical operation. Only give the numbers shown in the example. Stricly give numbers in numeric digits and your result should be in the format $>$ Number A: xxx $||$ Number B: xxx."}
        \item \textbf{Mathematical Reasoning} - \textit{"Think step-by-step and give proper reasoning steps first and then given your final answer. The format should be $>$ Reasoning: xxx $||$ Answer: xxx . The Reasoning part should contain reasons to derive the answer and the Answer part should only contain the answer. Your response should strictly follow this format and not just give the answer of the mathematical operation. It's important that you give reasoning before you answer."}
    \end{itemize}
  \end{PlainBox}
\caption{(Operator Induction Task) Prompts to the LMM for generating answers in specific formats suited for evaluation.}
  \label{fig:op_prompts}
\end{figure}

We list out a few examples which we curate for the Operator Induction task to enhance mathematical reasoning. Each image in the dataset contains a set of 2 numbers or operands and a hidden mathematical operation. The result of the correct mathematical operation is also provided for the support set examples. The task is to induce the mathematical operation used in the support set to calculate the answer of the query image containing two new operands. As finetuning on a single answer token limits the token generation capacity of the LMM, we further modify the support set examples to list out detailed mathematical steps before calculating the answer. Finetuning on this reasoning data improves both the generation capacity and reasoning ability of the LMM. We further provide a few examples of this hand-curated data in Figure \ref{fig:reason_data}. 

\begin{figure}[h]
  \centering
  \begin{SideBySideBox}
    \includegraphics[width=\linewidth]{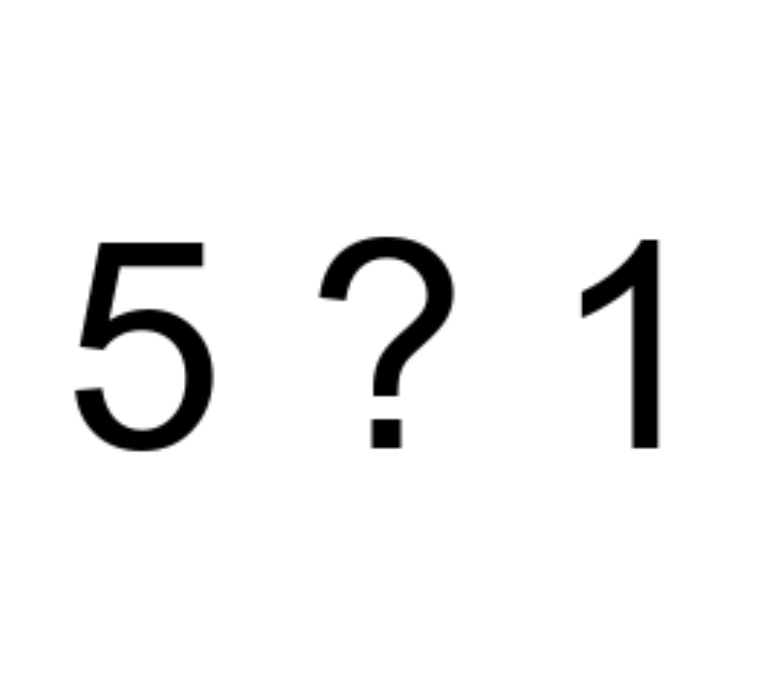}
    %
    \tcblower
    %
    \textbf{Original Answer}: \textit{4}\par
    \textbf{Detailed Answer}: 
    \textit{There are two numbers, \(5\) and \(1\). Performing some mathematical operation gives the answer \(4\). 
    So if we think about adding the numbers, \(5 + 1 = 6\), subtracting them, \(5 - 1 = 4\), multiplying them, \(5 \times 1 = 5\). 
    This implies that the hidden operation must be subtraction (–) and the result is \(4\).}
  \end{SideBySideBox}
    \begin{SideBySideBox}
    \includegraphics[width=\linewidth]{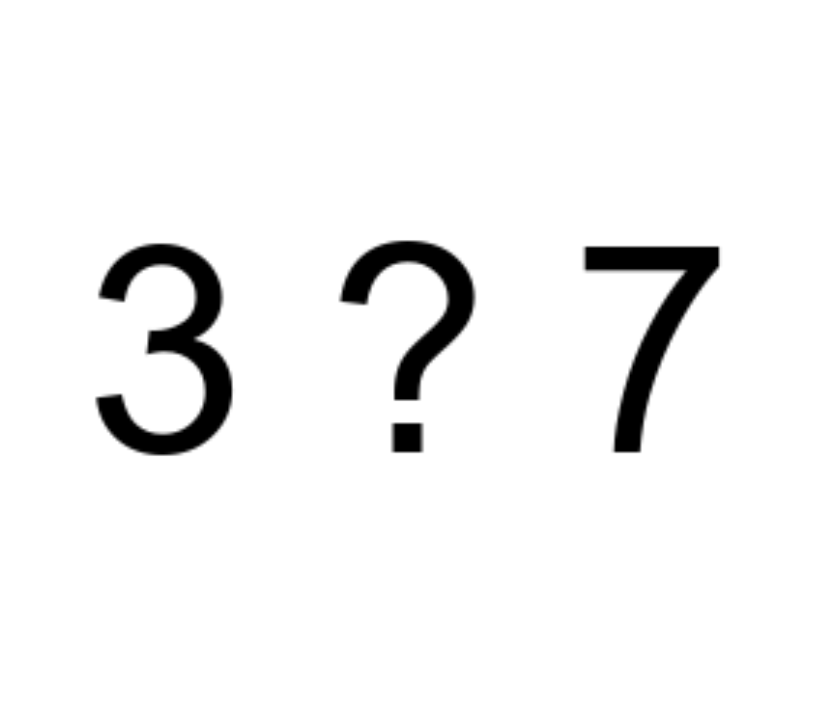}
    %
    \tcblower
    %
    \textbf{Original Answer}: \textit{21}\par
    \textbf{Detailed Answer}: 
    \textit{There are two numbers, $3$ and $7$. Performing some mathematical operation gives the answer $21$. So if we think about multiplying the numbers, $3 \times 7 = 21$, adding the numbers, $3 + 7 = 10$, subtracting the numbers, $3 - 7 = -4$. This implies that the hidden operation must be multiplication or $\times$ and the result is $21$.}
  \end{SideBySideBox}
  
  \caption{(Operator Induction Math Reasoning) Few examples of our hand-curated data with mathematical reasoning steps.}
  \label{fig:reason_data}
\end{figure}

We used Qwen2.5VL-32B-Instruct \citep{qwen2025qwen25technicalreport} as a judge for evaluating the Mathematical Reasoning component of the problem where LMMs responded with detailed reasoning steps before the answer. Evaluation of responses was done by prompting the judge to score a response between 0--3 based on if it thinks the reasoning and answer are correct. We then calculated mean score as the percentage of total score assigned by the Qwen-2.5-VL (Judge) to the responses relative to the maximum possible score.
\begin{align}
    \text{Mean Percent Score} = \dfrac{\sum_{i=1}^NS_i}{3*N} \times 100
\end{align}
where $S_i$ is the score assigned by Qwen2.5-VL for the ith response and $N$ is the total number of responses. We provide the prompt to the judge for this evaluation in Figure \ref{fig:judge_prompts}.

\begin{figure}[h]
  \centering
  \begin{PlainBox}
    \textbf{Judge Prompt} - \textit{"You are given a few in-context examples of a mathematical induction problem. The in-context examples each have an image with two numbers and a '?' which is supposed to be some mathematical operation. You are given a solution that gives the answer and the reasoning on how to calculate that answer using some mathematical operation applied on those two numbers in the image. The task is to induce the correct mathematical operation from the given examples, and use that operation to calculate the result of a query image with different numbers.\\\\
    After this, you are then given a reference answer written by experts and a candidate response. The candidate response is in format Reasoning: xxx $||$ Answer: xxx . 
      The reasoning part contains reasoning about how the candidate arrived at the solution, and the Answer part contains their final answer. Your task is to judge if the reasoning and the answer of the candidate response are correct or not after considering the in-context examples, query image, question, reference answer, and your own reasoning of the mathematical induction problem.  \\\\
      The rating should be done on a scale of 0--3, where 0 indicates when the response is ambiguous or does not follow the format, 1 is for when both the reasoning and answer are wrong, 2 is for when either only reasoning or answer is correct, 3 is for when both the reasoning and answer are correct.\\\\ Be strict in your judgement and do not give a higher rating unless the candidate response contains accurate reasoning and exact answer. Thorougly check each and every part of the candidate response and make sure it does not contain extra irrelevant operations or answers. If it does then give a lower rating accordingly. The candidate response should follow the format and conclude with the correct answer. If it does not, that means their answer is wrong. Also give your rationale before rating. Give the final rating as $>$ Rating: xxx"}
  \end{PlainBox}
\caption{(Operator Induction Math Reasoning) Prompts for the Qwen2.5VL-32B-Instruct to evaluate LMM responses on a scale of 0--3. It is given 1 to 4 in-context examples for understanding the mathematical induction task before the LMM (candidate) response for better evaluation.}
  \label{fig:judge_prompts}
\end{figure}

We also provide a few examples of LMM predictions for task induction (Figure \ref{fig:pred_task}) and perception (Figure \ref{fig:pred_percept}) and mathematical reasoning (Example 1: Figure \ref{fig:prediction1},  \ref{fig:judge_pred1} and Example 2: Figure \ref{fig:prediction2}, \ref{fig:judge_pred2})

\begin{figure}[!htbp]
  \centering
  {Example (Task Induction): Support shot 1}
  \begin{SideBySideBox}
    
    \includegraphics[width=\linewidth]{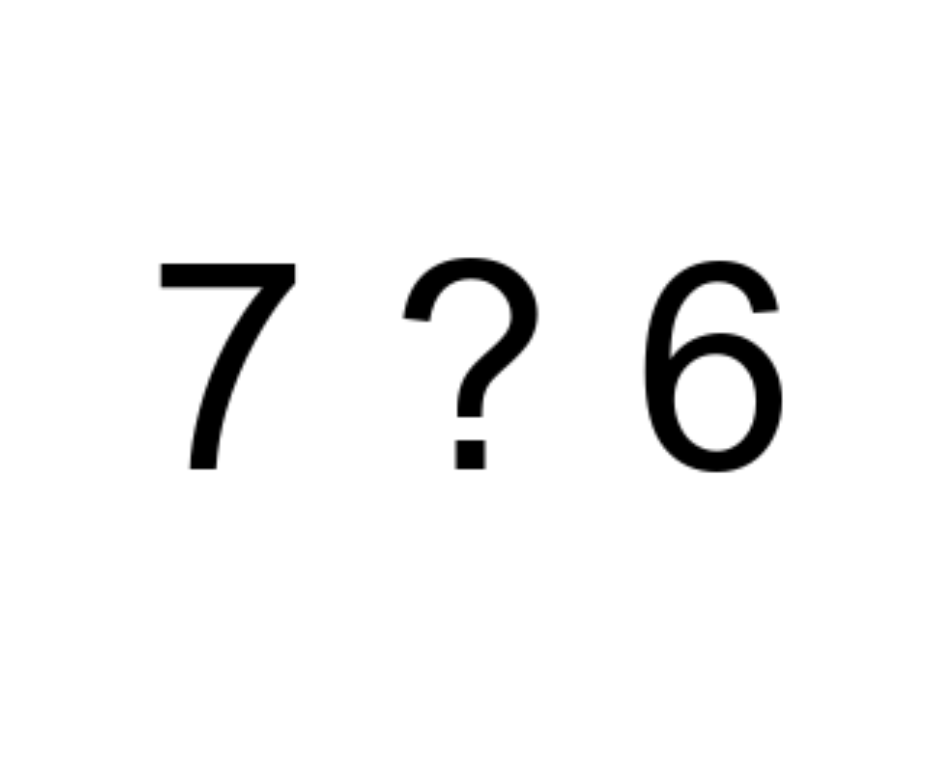}
    
    %
    \tcblower
    %
    \vspace{20pt}
    \textbf{Question}: \textit{What is the result of the following mathematical expression?}\par
    \textbf{Answer}: \textit{42} 
  \end{SideBySideBox}
  
{Example (Task Induction): Support shot 2}
  \begin{SideBySideBox}
    
    \includegraphics[width=\linewidth]{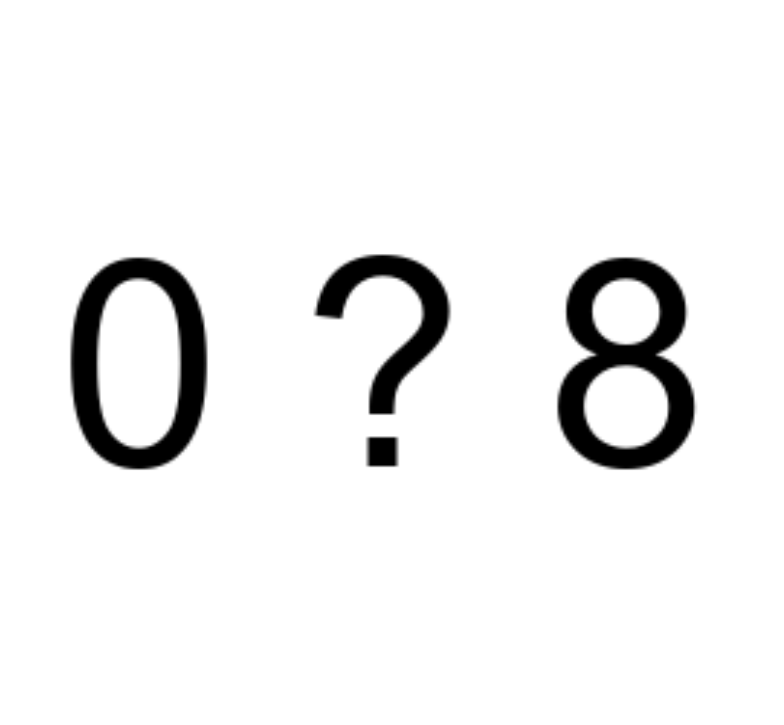}
    
    %
    \tcblower
    %
    \vspace{20pt}
    \textbf{Question}: \textit{What is the result of the following mathematical expression?}\par
    \textbf{Answer}: \textit{0} 
  \end{SideBySideBox}
  {Example (Task Induction): Query}
  \begin{SideBySideBox}
    \includegraphics[width=\linewidth]{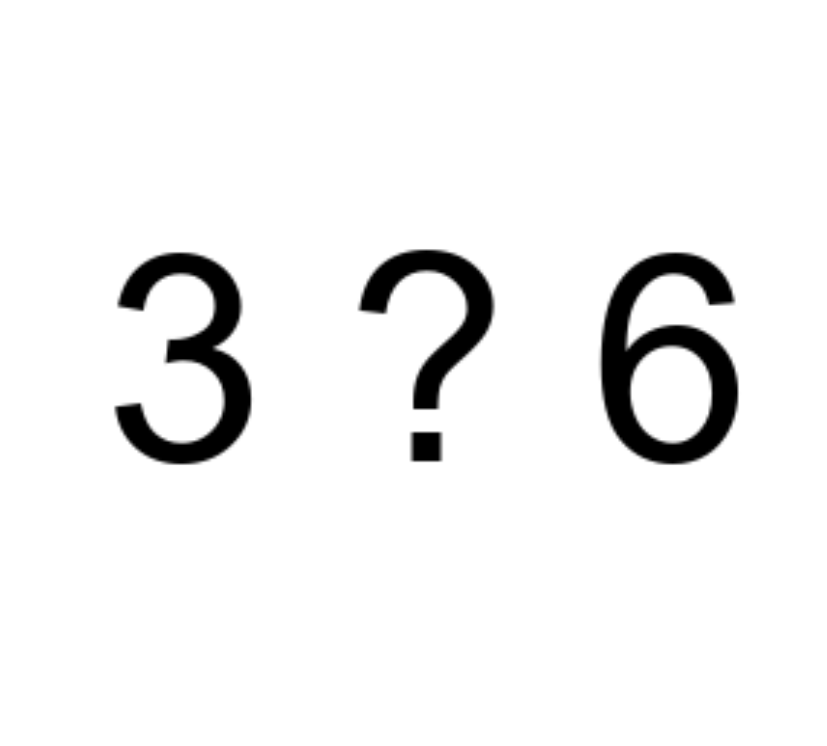}
    %
    \tcblower
    %
    \vspace{20pt}
   \textbf{Question}: \textit{What mathematical operation should be used in this example? Strictly answer in one word.}\par
   \vspace{2pt}
    \textbf{LMM prediction}: 
    \textit{Multiplication}
  \end{SideBySideBox}
  
  \caption{(Operator Induction Task Induction) An example of a 2-shot task induction for multiplication operation}
  \label{fig:pred_task}
\end{figure}

\begin{figure}[!htbp]
  \centering
  {Example (Perception): Support shot 1}
  \begin{SideBySideBox}
    
    \includegraphics[width=\linewidth]{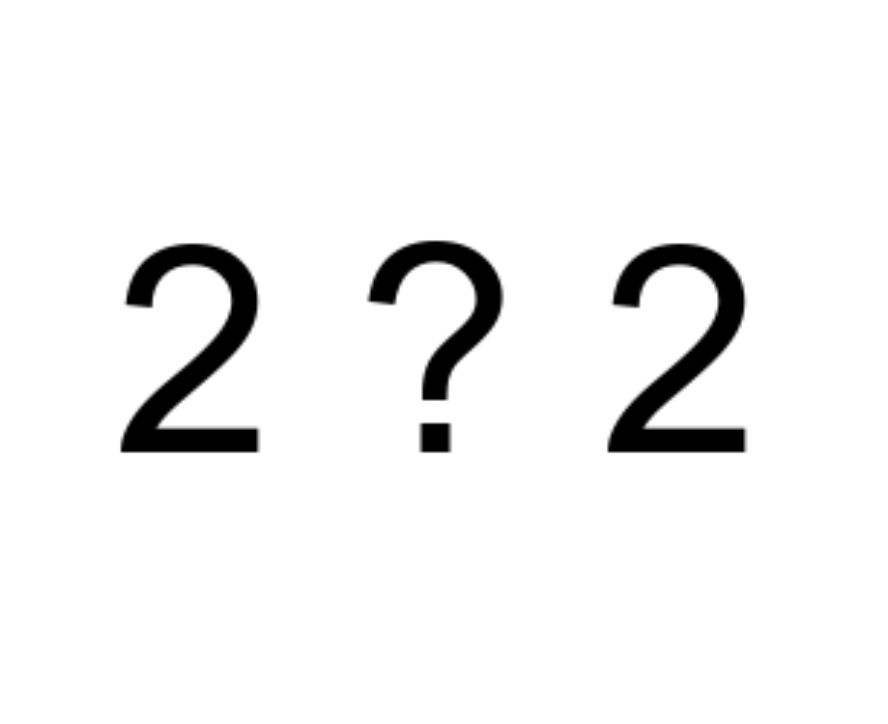}
    
    %
    \tcblower
    %
    \vspace{20pt}
    \textbf{Question}: \textit{What is the result of the following mathematical expression?}\par
    \textbf{Answer}: \textit{0} 
  \end{SideBySideBox}
  
{Example (Perception): Support shot 2}
  \begin{SideBySideBox}
    
    \includegraphics[width=\linewidth]{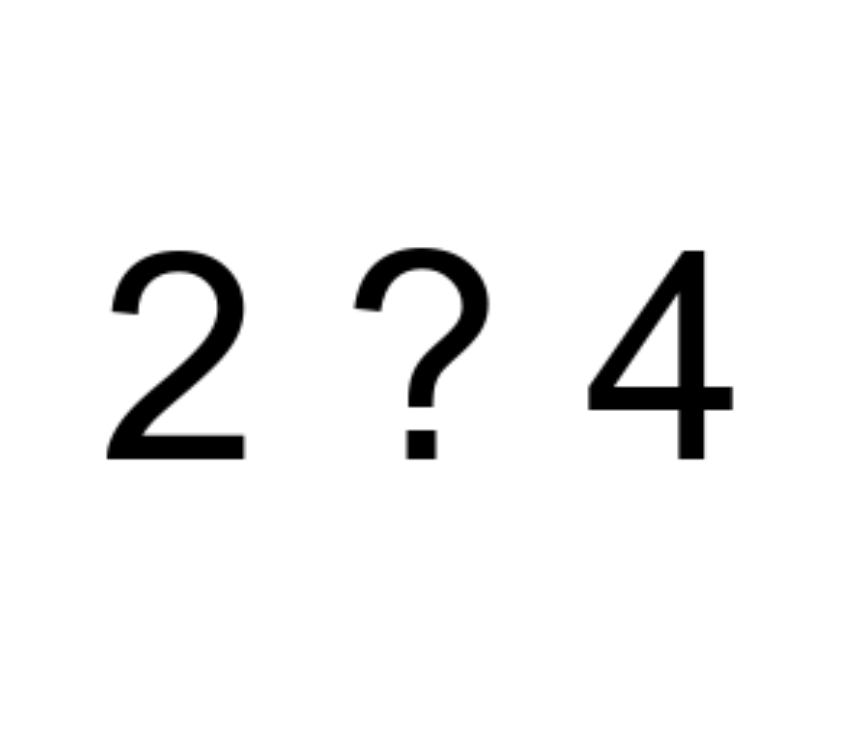}
    
    %
    \tcblower
    %
    \vspace{20pt}
    \textbf{Question}: \textit{What is the result of the following mathematical expression?}\par
    \textbf{Answer}: \textit{-2} 
  \end{SideBySideBox}
  {Example (Perception): Query}
  \begin{SideBySideBox}
    \includegraphics[width=\linewidth]{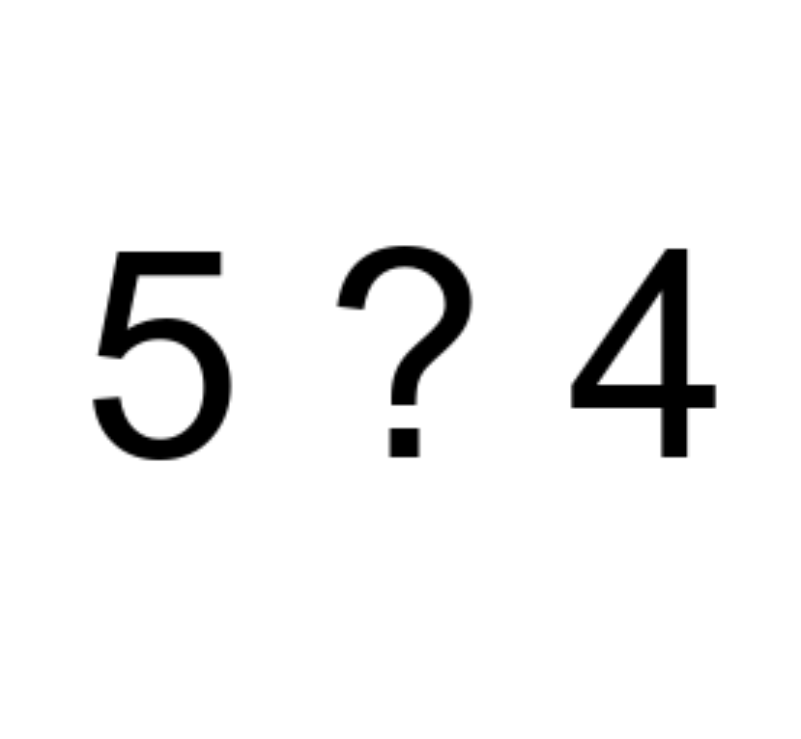}
    %
    \tcblower
    %
    \vspace{10pt}
   \textbf{Question}: \textit{What are the numbers in this example? Do not calculate the answer after applying mathematical operation. Only give the numbers shown in the example. Stricly give numbers in numeric digits and your result should be in the format $>$ Number A: xxx $||$ Number B: xxx.}\par
   \vspace{2pt}
    \textbf{LMM prediction}: \textit{Number A: 5 $||$ Number B: 4}
    \textit{}
  \end{SideBySideBox}
  
  \caption{(Operator Induction Perception) An example of a 2-shot perception task to detect operands}
  \label{fig:pred_percept}
\end{figure}

\begin{figure}[!htbp]
  \centering
  {Example 1: Support}
  \begin{SideBySideBox}
    
    \includegraphics[width=\linewidth]{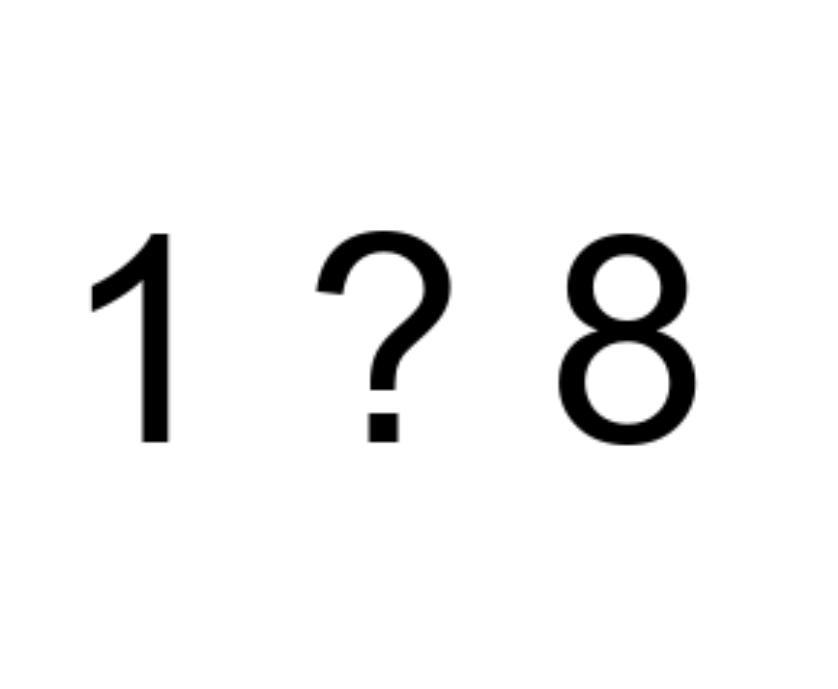}
    
    %
    \tcblower
    %
    \textbf{Question}: \textit{What is the result of the following mathematical expression?}\par
    \textbf{Answer}: 
    \textit{There are two numbers, $1$ and $8$. Performing some mathematical operation gives the answer $8$. So if we think about subtracting the numbers, $1 - 8 = -7$, multiplying the numbers, $1 \times 8 = 8$, adding the numbers, $1 + 8 = 9$. This implies that the hidden operation must be multiplication or x and the result is $8$.}
    
  \end{SideBySideBox}
  {Example1: Query}
  \begin{SideBySideBox}
    \vspace{30pt}
    \includegraphics[width=\linewidth]{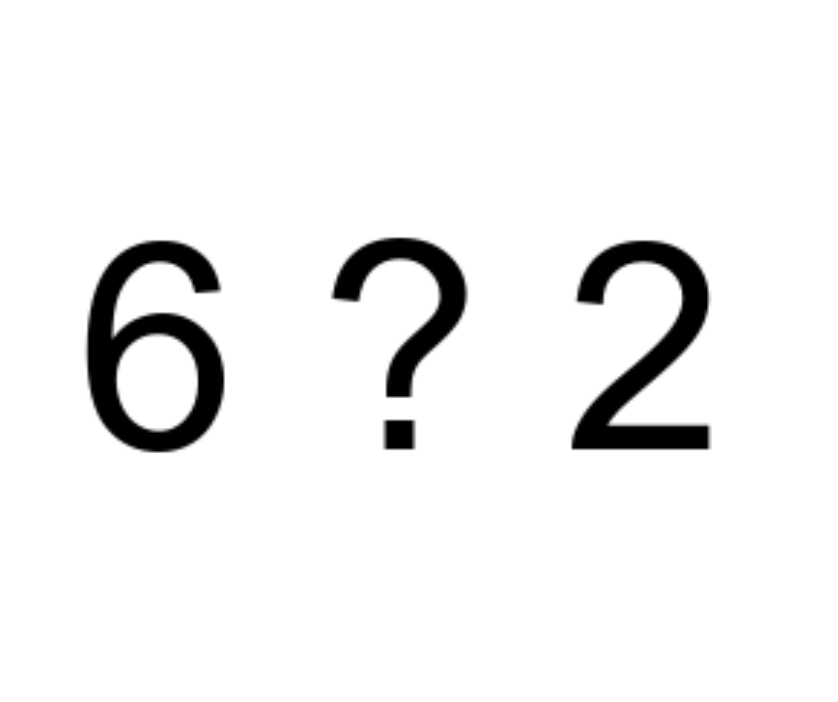}
    %
    \tcblower
    %
   \textbf{Question}: \textit{What is the result of the following mathematical expression? Think step-by-step and give proper reasoning steps first and then given your final answer. The format should be $>$ Reasoning: xxx $||$ Answer: xxx .The Reasoning part should contain reasons to derive the answer and the Answer part should only contain the answer. Your response should stricly follow this format and not just give the answer of the mathematical operation. It's important that you give reasoning before you answer.}\par
    \textbf{LMM prediction}: 
    \textit{There are two numbers, $6$ and $2$. We need to find the result of an operation between them. The question is asking for the result of the operation, which is $12$. So the answer is Reasoning: The two numbers are $6$ and $2$. The operation between them is multiplication. $6 \times 2 = 12$ $||$ Answer: $12$}
  \end{SideBySideBox}
  
  \caption{(Operator Induction Math Reasoning) An example of a 1-shot mathematical reasoning task with Judge Rating: 3 (shown in Figure \ref{fig:judge_pred1})}
  \label{fig:prediction1}
\end{figure}

\begin{figure}[h]
  \centering
{Example 1: Judge Response}
\begin{PlainBox}
{\bfseries Judge Response -- }{\itshape{\#\#\# Analysis of the Problem\\

\#\#\#\# In-Context Example:\\
The in-context example provides the expression `1 ? 8` and states that the result is **8**. The solution explains that among the possible operations (subtraction, multiplication, addition), the operation that yields the result **8** is **multiplication** (`1 × 8 = 8`). This establishes the pattern that the hidden operation is **multiplication**.\\

\#\#\#\# Query Image:\\
The query image presents the expression `6 ? 2`. Based on the in-context example, we infer that the hidden operation is **multiplication**.\\

\#\#\#\# Candidate Response:\\
The candidate response states:

\begin{tabular}{lp{11cm}}
- & **Reasoning**: The two numbers are 6 and 2. The operation between them is multiplication. \(6 \times 2 = 12\).\\
- & **Answer**: 12
\end{tabular}
\smallskip 

\#\#\#\# Reference Answer:\\
The reference answer is also **12**.\\

\#\#\# Evaluation of the Candidate Response
\begin{enumerate}
\item **Correct Identification of the Operation**:
   - The candidate correctly identifies the operation as **multiplication** based on the in-context example. This is consistent with the pattern established in the in-context example.

\item **Correct Application of the Operation**:
   - The candidate applies the multiplication operation correctly: \(6 \times 2 = 12\).

\item **Final Answer**:
   - The candidate's final answer is **12**, which matches the reference answer.

\item **Clarity of Reasoning**:
   - The reasoning provided is clear and logically follows the pattern established in the in-context example. The candidate explicitly states the operation and shows the calculation step.

\item **Format Compliance**:
   - The candidate response follows the required format: "Reasoning: xxx $||$ Answer: xxx".
\end{enumerate}
\smallskip

\#\#\# Final Verdict\\
The candidate's response is both logically sound and mathematically correct. Both the reasoning and the final answer are accurate, and the response adheres to the required format.\\

$>$ **Rating: 3**
}}
\end{PlainBox}
\caption{(Operator Induction Math Reasoning) The Judge (Qwen2.5-VL-32B) evaluates the response of the LMM in Figure \ref{fig:prediction1} to provide correct rating.}
\label{fig:judge_pred1}
\end{figure}

\begin{figure}[!htbp]
  \centering
  {Example 2: Support}
  \begin{SideBySideBox}
    
    \includegraphics[width=\linewidth]{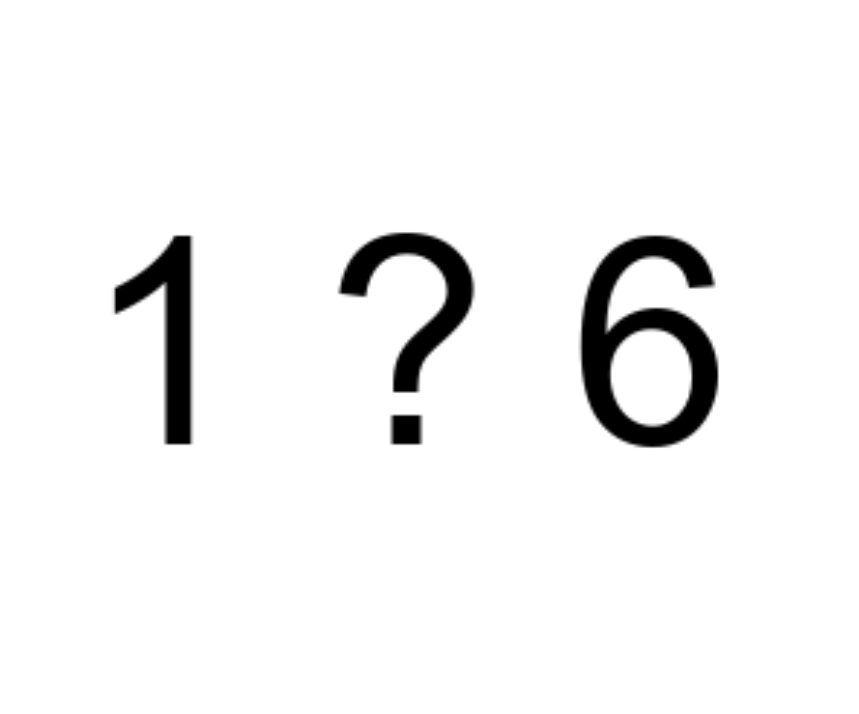}
    
    %
    \tcblower
    %
    \textbf{Question}: \textit{What is the result of the following mathematical expression?}\par
    \textbf{Answer}: 
    \textit{There are two numbers, $1$ and $6$. Performing some mathematical operation gives the answer $6$. So if we think about subtracting the numbers, $1 - 6 = -5$, multiplying the numbers, $1 \times 6 = 6$, adding the numbers, $1 + 6 = 7$. This implies that the hidden operation must be multiplication or $\times$ and the result is $6$}
    
  \end{SideBySideBox}
  {Example 2: Query}
  \begin{SideBySideBox}
    \vspace{30pt}
    \includegraphics[width=\linewidth]{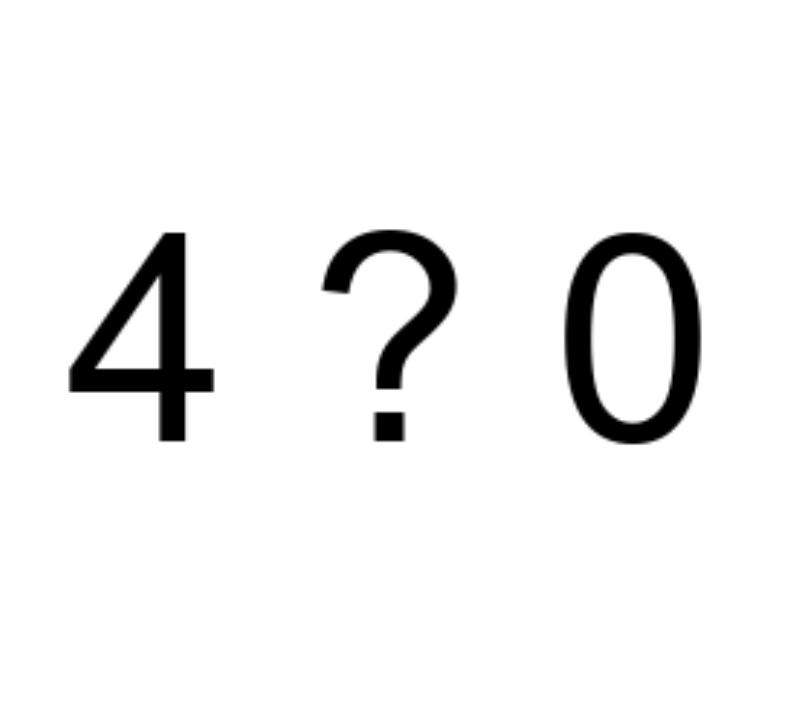}
    %
    \tcblower
    %
   \textbf{Question}: \textit{What is the result of the following mathematical expression? Think step-by-step and give proper reasoning steps first and then given your final answer. The format should be $>$ Reasoning: xxx $||$ Answer: xxx .The Reasoning part should contain reasons to derive the answer and the Answer part should only contain the answer. Your response should stricly follow this format and not just give the answer of the mathematical operation. Its important that you give reasoning before you answer.}\par
    \textbf{LMM prediction}: 
    \textit{Reasoning: There are two numbers, $4$ and $1$. Performing some mathematical operation gives the answer $5$. So if we think about adding the numbers, $4 + 1 = 5$, subtracting the numbers, $4 - 1 = 3$, multiplying the numbers, $4 \times 1 = 4$. This implies that the hidden operation must be addition or $+$ and the result is $5$. $||$ Answer: $5$}
  \end{SideBySideBox}
  
  \caption{(Operator Induction Math Reasoning) An example of a 1-shot mathematical reasoning task with Judge Rating: 1 (shown in Figure \ref{fig:judge_pred2})}
  \label{fig:prediction2}
\end{figure}

\begin{figure}[!htbp]
  \centering
\begin{PlainBox}
{\bfseries Judge Response -- }{\itshape{\#\#\# Analysis of the Problem\\
\#\#\#\# In-Context Example:\\
The in-context example provides the expression `1 ? 6` and the solution states that the result is **6**. The reasoning provided suggests testing different operations:
\begin{itemize}[noitemsep]
\item  Subtraction: \(1 - 6 = -5\);
\item Multiplication: \(1 \times 6 = 6\);
\item Addition: \(1 + 6 = 7\)\\
\end{itemize}
\vspace*{-.5cm}
From these tests, the solution concludes that the hidden operation is **multiplication**, as \(1 \times 6 = 6\).\\

\#\#\#\# Query Image:\\
The query image is `4 ? 0`. We need to determine the result of this expression based on the same hidden operation identified in the in-context example.\\

\#\#\#\# Reference Answer:\\
The reference answer provided is **0**.\\

\#\#\#\# Candidate Response:\\
\begin{tabular}{lp{12cm}}
\multicolumn{2}{l}{The candidate's reasoning and answer are as follows:}\\
- &**Reasoning**: The candidate incorrectly identifies the numbers as **4 and 1** instead of **4 and 0**. They test different operations:\\
  - & Addition: \(4 + 1 = 5\); Subtraction: \(4 - 1 = 3\); Multiplication: \(4 \times 1 = 4\)\\
  - &Conclusion: The hidden operation is **addition**, and the result is **5**.\\
- &**Answer**: The candidate concludes that the result is **5**.\\
\end{tabular}
\smallskip

\#\#\# Evaluation of the Candidate Response
\begin{enumerate}[noitemsep]
\item  **Correct Identification of the Hidden Operation**:

\begin{tabular}{lp{11cm}}
   - & The in-context example clearly indicates that the hidden operation is **multiplication** (\(1 \times 6 = 6\)).\\
   - & However, the candidate incorrectly assumes the hidden operation is **addition** based on their misinterpretation of the numbers.
\end{tabular}
\item  **Correct Interpretation of the Numbers**:

\begin{tabular}{lp{11cm}}
   - & The query image shows the numbers **4 and 0**, but the candidate mistakenly uses **4 and 1** in their reasoning.\\
   -  & This error leads to incorrect calculations and conclusions.\\ \end{tabular}

\item **Correct Application of the Operation**:

\begin{tabular}{lp{11cm}}
   - & Even if we overlook the misidentification of the numbers, the candidate's reasoning is flawed because they do not apply the correct hidden operation (multiplication) as indicated by the in-context example.\\\end{tabular}

\item 4. **Final Answer**:

\begin{tabular}{lp{11cm}}
   - & The candidate's final answer is **5**, which is incorrect. The correct answer, based on the reference and the hidden operation of multiplication, should be **0** (\(4 \times 0 = 0\)). \\ \end{tabular}
   \end{enumerate}

\#\#\# Rating Justification\\
- The candidate's reasoning is **incorrect** because:
\begin{enumerate}
  \item  They misidentify the numbers in the query image.
  \item They incorrectly deduce the hidden operation as addition instead of multiplication.
  \end{enumerate}
- The candidate's final answer (**5**) is also **wrong**.\\

Given these errors, the candidate's response does not align with the in-context example or the reference answer. Both the reasoning and the answer are incorrect.\\

\#\#\# Final Rating:\\
$>$ Rating: 1
}}
\end{PlainBox}
\caption{(Operator Induction Math Reasoning) The Judge (Qwen2.5-VL-32B) evaluates the response of the LMM in Figure \ref{fig:prediction2} to provide correct rating.}
\label{fig:judge_pred2}
\end{figure}

\clearpage
\newpage



\subsubsection{Attention Entropy Analysis}
\label{sec_app:att_entropy}

We analyze attention patterns by extracting post-softmax attention scores of the underlying LLM for the LLaVA-ATT-Qwen2.5 7B model on the VL-ICL bench for different shot scenarios. For each generation step, we first re-normalize the extracted attention scores over soft prompt tokens (containing image information) $q_i = \displaystyle \dfrac{a_i}{\sum_{j=1}^{|P|} a_j}$, where $P$ refers to soft prompts. We then calculate the normalized Shannon entropy as, $$H(Q) = \displaystyle \dfrac{-\sum_{i=1}^{n} q_i \log q_i}{log(n)}$$ to account for the difference in the number of soft prompts. We average this across all generation steps, LLM attention heads, layers, and examples across different VL-ICL datasets for MAPD with FT $\leq 30$ and our best In-Context\textsuperscript{PD} baseline that uses ICL and report results in Table \ref{tab:att_entropy}. We see a more uniform distribution of attention scores for MAPD over the fixed set of soft prompts, i.e. 256 and hence the entropy remains consistent even on increasing the number of shots. On the other hand, for In-Context\textsuperscript{PD}, as we increase the number of shots, the number of soft prompts increase. For example, 8-shot + 1 query, ICL requires $256 \times (8+1) =  2304$ soft prompts. Given this relatively larger context, we see many soft prompt tokens containing very low attention scores leading to a lower value of normalized entropy. This highlights an inherent limitation of the LMM that it is unable to attend to all the soft prompts over longer context lengths.

\begin{table}[h]
  \caption{Normalized Attention Entropy on VL-ICL Bench. TTA = Test-Time Adaptation. FT $\leq 30$ = Finetuning with $\leq 30$ gradient steps. ICL = In-Context Learning.}
  \label{tab:att_entropy}
  \centering
  \begin{tabular}{@{}c|c|c|c|c|c|c@{}}
\toprule
   Training Methods & TTA & 0-S & 1-S & 2-S & 4-S & 8-S \\ 
   \midrule
    In-Context\textsuperscript{PD} & ICL & \textbf{0.84} & 0.75 & 0.63 & 0.51 & 0.45 \\
    MAPD & FT$\leq 30$ & \textbf{0.84} & \textbf{0.81} & \textbf{0.84} & \textbf{0.80} & \textbf{0.81}  \\
    \bottomrule
    \end{tabular}
\end{table}

\subsubsection{Scaling to More Shots}
\label{sec_app:more_shots}

Here, we look into the performance of different prompt distillation methods with finetuning-based test time adaptation for larger number of shots and for 3 tasks from the VL-ICL datasets for LLaVA-ATT-Qwen2.5 7B LMM. We see similar performance gains with the introduction of more shots as shown in Table \ref{tab:baseline}. Both the meta-task learning methods, Multi-Task\textsuperscript{PD} and MAPD perform quite well with MAPD showing outstanding performance for Operator Induction.

\begin{table}[h]

  \caption{Operator Induction Results.}
  \label{tab:op_large_shots}
  \centering
  \begin{tabular}{@{}c|c|c|c|c@{}}
\toprule
   Training Methods & Meta-task & 16-S & 32-S & 64-S \\ 
   \midrule
    NoMeta-task\textsuperscript{PD} & \xmark & 73.3 & 73.3 & 80.0 \\    
    Model-Avg\textsuperscript{PD} & \xmark & 71.7 & 78.3 & 80.5 \\
    In-Context\textsuperscript{PD} & \cmark & 58.3 & 53.3 & 76.7 \\
    Multi-Task\textsuperscript{PD} & \cmark & 73.3 & 67.7 & 80.0 \\
    MAPD & \cmark & \textbf{80.0} & \textbf{81.0} & \textbf{83.3} \\
    \bottomrule
    \end{tabular}
\end{table}

\begin{table}[h]

  \caption{CLEVR Count Induction Results.}
  \label{tab:clevr_large_shots}
  \centering
  \begin{tabular}{@{}c|c|c|c|c@{}}
\toprule
   Training Methods & Meta-task & 16-S & 32-S & 64-S \\ 
   \midrule
    NoMeta-task\textsuperscript{PD} & \xmark & 35.5 & 30.0 & 36.5 \\    
    Model-Avg\textsuperscript{PD} & \xmark & 30.0 & 34.5 & 37.0 \\
    In-Context\textsuperscript{PD} & \cmark & 25.5 & 34.5 & 32.5 \\
    Multi-Task\textsuperscript{PD} & \cmark & 38.0 & \textbf{41.5} & 38.5 \\
    MAPD & \cmark & \textbf{40.0} & 40.5 & \textbf{41.0} \\
    \bottomrule
    \end{tabular}
\end{table}

\begin{table}[h]
  \centering

  \caption{TextOCR Results.}
  \label{tab:textocr_large_shots}
  
  \begin{tabular}{@{}c|c|c|c|c@{}}
\toprule
   Training Methods & Meta-task & 16-S & 32-S & 64-S \\ 
   \midrule
    NoMeta-task\textsuperscript{PD} & \xmark & 29.0 & 26.5 & 30.5 \\    
    Model-Avg\textsuperscript{PD} & \xmark & 29.0 & 29.5 & 31.5 \\
    In-Context\textsuperscript{PD} & \cmark & 26.5 & 26.0 & 28.5\\
    Multi-Task\textsuperscript{PD} & \cmark & 27.0 & \textbf{32.5} & \textbf{33.5} \\
    MAPD & \cmark & \textbf{30.5} & 31.5 & 31.5 \\
    \bottomrule
    \end{tabular}
\end{table}

\newpage

\subsection{Test-Time Compute Analysis for ICL vs FT}
\label{sec_app:tta_compute}
\begin{figure}[h]
\begin{subfigure}[t]{0.48\textwidth}
\vspace{0pt}\centering
\begin{tikzpicture}[scale=.6]
\begin{axis}[
    ybar,
    bar width=12pt,
    width=12cm,
    height=7cm,
    enlarge x limits=0.15,
    ymin=0,
    ymax=16.0,
    ylabel={Time taken (in seconds)},
    ylabel style={font=\normalsize,yshift=-.2cm},
    xtick=data,
    xticklabel style={align=center, font=\small},
    yticklabel style={font=\small},
    xticklabels={
        {1-Shot},
        {2-Shot},
        {4-Shot},
        {5-Shot}
    },
    legend style={
        font=\small,
        at={(0.5,1.14)},
        anchor=north,
        legend columns=4
    },
    grid=major,
    grid style={dashed,gray!20},
    nodes near coords,
    every node near coord/.append style={font=\fontsize{7}{5}\selectfont,
      yshift=0pt,text=black}
]
\addplot+[style={fill=Blue!70, draw=black}] plot coordinates {(0,3.0) (1,4.0) (2,6.5) (3,7.8)};
\addplot+[style={fill=Orange!70, draw=black}] plot coordinates {(0,6.0) (1,7.7) (2,12.3) (3,14.3)};
\legend{ICL, FT}
\end{axis}
\end{tikzpicture}
\end{subfigure}\hfill
\begin{subfigure}[t]{0.48\textwidth}
\vspace{0pt}\centering
\begin{tikzpicture}[scale=.6]
\begin{axis}[
    width=11cm,
    height=7cm,
    xlabel={TFLOPs},
    ylabel={Mean Accuracy},
    ylabel style={font=\normalsize,yshift=-.2cm},
    ymin=0.0, ymax=45,
    xmin=0.0, xmax=800,
    xtick={10, 100, 200, 400, 600, 800},
    ytick={0, 10, 20, 25, 30, 35, 40, 45},
    grid=both,
    axis lines=box,
    legend style={
        font=\small,
        at={(0.5,1.03)},
        anchor=south,
        legend columns=2
        minimum width=7cm
    },
    tick label style={font=\small},
    title style={font=\small},
    label style={font=\small},
    every node near coord/.append style={font=\footnotesize, xshift=3pt, yshift=5pt},
    enlargelimits=0.05,
    line width=1.5pt,
    mark size=2pt,
    mark options={solid},
    axis lines=left,
]
\addplot[
  color=blue,
    mark=*,
    thick
] coordinates {
    (0, 0.0)
    (100, 25.5)
    (200, 30.0)
    (400, 32.5)
    (600, 30.5)
    (800, 31.5)
};

\addplot[
  color=orange,
    mark=*,
    thick
] coordinates {
    (0, 0.0)
    (100, 20.5)
    (200, 26.5)
    (400, 34.5)
    (600, 35.5)
    (800, 37.5)
    
};

\legend{ICL, FT}
\end{axis}
\end{tikzpicture}
\end{subfigure}
\caption{(a) \textbf{Left}: Computational time taken per test example (query) by ICL (blue) and FT (orange) for different number of shots. (b) \textbf{Right}: FLOPs matched evaluation across all the VL-ICL test sets with mean accuracy for ICL (blue) and  FT (orange).}
\label{fig:test_time_compute}
\end{figure}

{Test-time finetuning (FT) for 30 gradient steps takes about twice as much inference time per test example (query) compared to in-context learning (ICL) under different few-shot scenarios as shown in Figure \ref{fig:test_time_compute}(a). This is not surprising as fine-tuning involves gradient computation, which is more expensive to run than a single forward pass in ICL.}

For a more fair comparison, we examine the amount of computation required between these different test-time adaptation methods. Figure~\ref{fig:test_time_compute}(b), shows  FLOPs-matched evaluation curves for ICL and FT, using In-Context\textsuperscript{PD} and MAPD as representative training methods, respectively.  We report mean accuracy across all (single-image) VL-ICL datasets. Test-time computation (TFLOPs) scales with the number of shots for ICL, while for FT, it can be scaled by increasing either number of shots or gradient steps.  We note that given a low test-time computational budget, ICL performs better than FT, but as the amount of computation is increased FT outperforms ICL. This indicates that FT adaptation is resource-intensive but scales better than ICL as the amount of computation is increased at test time. 

After 400 TFLOPs, In-Context\textsuperscript{PD} performance begins to decline because the large number of shots used ($\geq$32) exceeds its trained context length of 8,192 tokens. Training In-Context\textsuperscript{PD} with longer context would require $>$4 H200 GPUs, which exceeds our available compute resources. On the other hand, MAPD by design does not require training on long context lengths due to the use of a fixed set of distilled soft prompts for all shots. Additionally, we find that MAPD  is much more data-efficient:  at 400 TFLOPs, it achieves comparable performance with only 8 shots and 20 gradient steps, indicating better few-shot test-time adaptation.

\end{document}